\documentclass[manuscript,acmsmall,screen,nonacm]{acmart} %anonymous
\AtBeginDocument{%
  }

\setcopyright{acmlicensed}
\copyrightyear{2026}
\acmYear{2026}
\acmDOI{XXXXXXX.XXXXXXX}
\acmISBN{978-1-4503-XXXX-X/2018/06}
\acmJournal{TIST}
\acmVolume{0}
\acmNumber{0}

\usepackage{latexsym}
\usepackage{longtable, tabu, afterpage}
\usepackage{booktabs}
\usepackage{multirow}
\usepackage{makecell}
\usepackage{multicol}
\usepackage{pifont}% http://ctan.org/pkg/pifont
\usepackage[capitalize]{cleveref}
\usepackage{adjustbox}
\usepackage{balance}
\usepackage{tabularx}

\usepackage{tikz}
\usepackage[edges]{forest}
\usepackage{xcolor}
\definecolor{hidden-draw}{RGB}{106,142,189} 
\definecolor{hidden-blue}{RGB}{194,232,247} 
\definecolor{hidden-orange}{RGB}{217, 232, 252} 
\newcommand{\cmark}{\ding{51}}%
\newcommand{\xmark}{\ding{55}}%

\tikzstyle{my-box}=[
 rectangle,
 draw=hidden-draw,
 rounded corners,
 text opacity=1,
 minimum height=1.5em,
 minimum width=5em,
 inner sep=2pt,
 align=center,
 fill opacity=.5,
 ]
 \tikzstyle{leaf}=[my-box, minimum height=1.5em,
 fill=hidden-orange!60, text=black, align=left, font=\scriptsize,
 inner xsep=2pt,
 inner ysep=4pt,
 ]
 \tikzstyle{leaf-center}=[my-box, minimum height=1.5em,
 fill=hidden-orange!60, text=black, align=center, font=\scriptsize,
 inner xsep=2pt,
 inner ysep=4pt,
 ]

\newlength\oriarrayrulewidth  
\newcount\orilowpenalty
\newcommand\nobreakmidrule{%
 \noalign{\global\oriarrayrulewidth\arrayrulewidth\relax
          \global\orilowpenalty\@lowpenalty\relax  
          \global\@lowpenalty=\numexpr-10000\relax%
          \global\arrayrulewidth\lightrulewidth\relax}
 \hline
 \noalign{\global\@lowpenalty=\orilowpenalty\relax%
          \global\arrayrulewidth\oriarrayrulewidth\relax}}

\newcolumntype{R}[2]{%
    >{\adjustbox{angle=#1,lap=\width-(#2)}\bgroup}%
    l%
    <{\egroup}%
}
\begin{document}

%%
%% The "title" command has an optional parameter,
%% allowing the author to define a "short title" to be used in page headers.
\title{Generative Large Language Models in Automated Fact-Checking: A Survey}

%%
%% The "author" command and its associated commands are used to define
%% the authors and their affiliations.
%% Of note is the shared affiliation of the first two authors, and the
%% "authornote" and "authornotemark" commands
%% used to denote shared contribution to the research.
\author{Ivan Vykopal}
\orcid{0000-0002-0431-0044}
\affiliation{%
  \institution{Faculty of Information Technology, Brno University of Technology}
  \city{Brno}
  \country{Czech Republic}
}
\affiliation{%
  \institution{Kempelen Institute of Intelligent Technologies}
  \city{Bratislava}
  \country{Slovakia}
}
\email{ivan.vykopal@kinit.sk}

\author{Matúš Pikuliak}
\orcid{0000-0003-1353-9462}
\affiliation{%
  \institution{Kempelen Institute of Intelligent Technologies}
  \city{Bratislava}
  \country{Slovakia}}
\email{matus.pikuliak@kinit.sk}

\author{Simon Ostermann}
\orcid{0000-0002-0899-0657}
\affiliation{%
  \institution{German Research Center for Artificial Intelligence (DFKI)}
  \city{Saarbrücken}
  \country{Germany}
}
\affiliation{%
  \institution{Centre for European Research in Trusted AI (CERTAIN)}
  \city{Saarbrücken}
  \country{Germany}
}
\email{simon.ostermann@dfki.de}

\author{Marián Šimko}
\orcid{0000-0002-2306-4408}
\affiliation{%
  \institution{Kempelen Institute of Intelligent Technologies}
  \city{Bratislava}
  \country{Slovakia}}
\email{marian.simko@kinit.sk}

%%
%% By default, the full list of authors will be used in the page
%% headers. Often, this list is too long, and will overlap
%% other information printed in the page headers. This command allows
%% the author to define a more concise list
%% of authors' names for this purpose.
\renewcommand{\shortauthors}{Vykopal et al.}

\begin{abstract}
  The rapid spread of false and misleading information on online platforms poses a growing societal challenge, overwhelming the capacity of manual fact-checking and increasing the demand for scalable, reliable automation. Recent advances in generative large language models (LLMs) have broadened the scope of automated fact-checking beyond accuracy-driven prediction. LLMs are now integral components of fact-checking pipelines, supporting tasks such as generating new data, performing and assisting with fact verification, and shaping how fact-checking systems are evaluated. This survey provides a comprehensive overview of the role of generative LLMs in automated fact-checking, based on a systematic review of 199 research papers. We introduce a unifying taxonomy that captures how generative LLMs are integrated into fact-checking workflows and analyze their use across core fact-checking tasks, dataset construction and augmentation strategies, task formulations, and evaluation practices. Additionally, we investigate the impact of generative LLMs in multilingual and low-resource settings in fact-checking, highlighting trends, limitations, and gaps in current research. By consolidating fragmented research efforts and identifying methodological patterns, limitations, and open challenges, this survey maps the current state of generative LLMs in automated fact-checking. It aims to support researchers in developing more reliable, interpretable, and inclusive fact-checking systems, while outlining promising directions for future research in this rapidly evolving field.
\end{abstract}



%%
%% Keywords. The author(s) should pick words that accurately describe
%% the work being presented. Separate the keywords with commas.

% \received{20 February 2007}
% \received[revised]{12 March 2009}
% \received[accepted]{5 June 2009}

%%
%% This command processes the author and affiliation and title
%% information and builds the first part of the formatted document.
\maketitle

\section{Introduction}

The digital transformation of information ecosystems has significantly changed how claims and narratives are consumed. While online platforms and social media have expanded access to information, they have also amplified the spread of false information (e.g., misinformation or disinformation), often at a scale and speed that exceeds the capacity of traditional verification mechanisms~\cite{doi:10.1126/science.aao2998, doi:10.1126/science.aap9559}. The consequences of unchecked false information are well-documented, affecting public health~\cite{poland2010fear, info:doi/10.2196/23805}, political decision-making~\cite{doi:10.1177/1529100620946707}, and trust in institutions~\cite{doi:10.1177/1940161219900126, doi:10.1177/20563051211069048, bernardin2025assessing}. Addressing misinformation has therefore become a central challenge for both researchers and practitioners in the social sciences and computer science.

\textit{Fact-checking} represents a principled response to this challenge. Fact-checking methods systematically assess the veracity of claims using external evidence and transparent reasoning~\cite{guo-etal-2022-survey}. However, manual fact-checking remains costly and difficult to scale. Professional fact-checkers must operate under severe resource constraints, while the volume, diversity, and multilingual nature of online content continue to grow, outspeeding manual efforts. These limitations have motivated substantial research efforts toward automated fact-checking pipelines, which aim to assist human experts across various stages, such as claim detection, evidence retrieval, claim verification, and explanation generation~\cite{thorne-etal-2018-fever, ijcai2021p0619}. Importantly, these systems are generally conceived as decision-support tools rather than fully autonomous judges of truth~\cite{10.1145/3764592}.

Recent advances in LLMs have reshaped research on automated fact-checking. Beyond improvements in general-purpose language understanding and generation tasks~\cite{NEURIPS2020_1457c0d6, yang2025qwen3technicalreport}, LLMs provide a flexible mechanism for representing claims, interacting with external knowledge sources, and generating natural language explanations~\cite{10.5555/3495724.3496517}. In this survey, we focus on \textit{generative LLMs}, i.e.~models capable of producing free-form textual outputs. Such models enable new approaches to textual fact-checking, including claim normalization, sub-claim decomposition, evidence synthesis, and explanation generation. However, their use introduces challenges, including hallucination, limited faithfulness to evidence, difficulties in reliable evaluation, among others~\cite{augenstein2024factuality, wang-etal-2024-factuality}. Understanding these trade-offs is therefore essential for the responsible integration of LLMs into fact-checking workflows.

% An increasing number of studies have explored the application of generative LLMs to individual fact-checking tasks, often demonstrating gains in flexibility or scalability, particularly in multilingual or low-resource settings~\cite{pikuliak-etal-2023-multilingual, schlichtkrull2023averitec}. In parallel, several surveys have explored automated fact-checking and factuality from different perspectives. Earlier surveys provide task-centric overviews of fact-checking pipelines, but they largely predate the widespread adoption of generative LLMs~\cite{guo-etal-2022-survey}. More recent surveys focus on hallucination, factual consistency, or the evaluation of language generation~\cite{augenstein2024factuality, wang-etal-2024-factuality}, treating fact-checking primarily as a diagnostic lens rather than a standalone application. As a result, existing surveys tend to cover individual components or perspectives and do not provide a comprehensive analysis of how generative LLMs are employed across the entire textual fact-checking pipeline.

The purpose of this survey is to provide a structured and comprehensive synthesis of research on generative LLMs for automated textual fact-checking. We seek to clarify how generative LLMs are currently used, what methodological patterns have emerged, and where their limitations remain most pronounced. In this process, we emphasize design choices, evaluation practices, and assumptions that influence the reliability and interpretability of LLM-assisted fact-checking systems.

To this end, we conducted a systematic literature review and analyzed \textbf{199 research papers}. Based on this analysis, we introduce a unifying taxonomy that categorizes the roles of generative LLMs across core fact-checking tasks, including claim detection, previously fact-checked claim retrieval, evidence retrieval, and fact verification. Beyond these core pipeline tasks, we examine the use of LLMs for synthetic data generation and for evaluating fact-checking outputs, two areas that have received comparatively limited attention in prior surveys. We further analyze multilingual and low-resource fact-checking, highlighting both promising strategies and persistent gaps in linguistic coverage.

% Our contribution, after mentioning what we did, we need to focus on the highlights of the paper and our main contributions

This survey makes the following contributions:
\begin{itemize}
    \item We provide a systematic \textit{taxonomy} of how generative LLMs are integrated into automated textual fact-checking pipelines, covering data creation, prediction, and evaluation.
    \item We synthesize datasets, task formulations, and evaluations used in LLM-based fact-checking, \textit{identifying methodological trends and limitations}.
    \item We offer a focused analysis of \textit{multilingual and low-resource fact-checking} with generative LLMs, emphasizing challenges related to cross-lingual transfer, script diversity, and data scarcity.
    \item We \textit{discuss open challenges} and future research directions, with particular focus on faithfulness, interpretability, and the responsible deployment of LLM-assisted fact-checking systems.
\end{itemize}

% Research questions, together with the contributions, we need to mention the research questions we are tackling

Guided by these contributions, this survey analyzes the literature along the following key dimensions:
\begin{enumerate}
    \item \textbf{Tasks and Challenges.} How are generative LLMs applied across different fact-checking tasks, and what task-specific challenges arise from their use?
    \item \textbf{Datasets and Limitations.} Which datasets are used for LLM-based fact-checking, and how is synthetic data generation used to address limitations such as data scarcity and class imbalance?
    \item \textbf{Methodological Paradigms and Techniques.} What methodological paradigms are employed when integrating generative LLMs into fact-checking pipelines, and how are these paradigms approached by using LLMs?
    \item \textbf{Evaluation.} How are generative LLMs used for evaluating fact-checking outputs and systems, and what are the limitations of LLM-based evaluation?
    \item \textbf{Multilingual Fact-Checking.} How are generative LLMs applied in multilingual and low-resource fact-checking scenarios, and what strategies are used to mitigate linguistic and cross-lingual challenges?
    \item \textbf{Challenges and Future Directions.} What are the challenges and possible future directions of using generative LLMs for automated fact-checking?
\end{enumerate}

In this survey, we bring together a wide and rapidly growing body of research to provide a clear and coherent perspective on how generative language models are shaping automated fact-checking. Our goal is to guide future work toward building fact-checking systems that are reliable, scalable, and effective across languages, while highlighting the challenges and responsibilities involved in using these models in real-world verification workflows.

% RQ1: How do generative LLMs address different automated fact-checking tasks, and what task-specific challenges and limitations emerge from their application?

% RQ2: Which datasets are used for LLM-based fact-checking, and how do synthetic data generation techniques help overcome limitations such as dataset scarcity, class imbalance, and gaps in multilingual coverage?

% RQ3: Which methodological paradigms are employed with generative LLMs to perform automated fact-checking, and how do these approaches influence system performance, reasoning capabilities, and interpretability?

% RQ4: How can generative LLMs be leveraged to evaluate fact-checking outputs, and how effectively do LLM-based evaluations capture factuality, consistency, and alignment with human judgment?

% RQ5: How are generative LLMs applied in multilingual and low-resource fact-checking scenarios, and what strategies are used to address challenges such as linguistic diversity, low-resource languages, and cross-lingual adaptation?

% RQ6: What are the challenges and possible future directions of using generative LLMs for automated fact-checking?

\section{Background}

\subsection{Existing Surveys}

Several surveys have examined automated fact-checking and factuality from different perspectives. However, these studies differ substantially in scope and depth, and often provide only a partial coverage of generative LLMs within fact-checking pipelines. In this section, we summarize the most relevant surveys and clarify how their focus diverges from the objectives of our work. Table~\ref{tab:survey-compasion} provides a detailed comparison of these surveys.

Before generative LLMs became widely used, surveys on automated fact-checking primarily focused on traditional fact-checking approaches. For example, \citet{guo-etal-2022-survey} organize fact-checking as a multi-stage pipeline comprising claim detection, evidence retrieval, verdict prediction, and justification generation. Their survey provides a comprehensive overview of the tasks, corresponding datasets, modeling strategies, and research challenges. However, it primarily frames the problem in terms of traditional approaches, including early neural networks and transformer-based architectures. Consequently, generative LLMs remain largely unexplored. Furthermore, the survey does not address LLM-specific techniques such as in-context learning, retrieval-augmented generation, synthetic data generation with LLMs, or LLM-based evaluation, which are central to modern fact-checking systems.

More recent surveys have shifted attention toward factuality and hallucinations in LLMs. \citet{augenstein2024factuality} focus on factuality challenges in recent LLMs rather than on automated fact-checking as a process. Their survey analyzes hallucination phenomena, limitations of existing evaluation benchmarks, and broader societal and regulatory implications of factual errors in generated text. Although the authors acknowledge that LLMs can support fact-checking workflows (e.g., through claim extraction or cross-lingual verification), they do not discuss fact-checking tasks, datasets, or commonly used techniques. As a result, the use of LLMs for downstream fact-checking or for evaluation is not systematically addressed.

\begin{table}
\centering
\caption{Comparison of existing surveys across key analytical dimensions covered in this work.}
\resizebox{\textwidth}{!}{
\begin{tabular}{lcccccl}
\toprule
 \textbf{Paper} & \textbf{\makecell[c]{Tasks\\\& Challenges}} & \textbf{\makecell[c]{Datasets\\\& Limitations}} & \textbf{\makecell[c]{Methodological\\Paradigms\\\& Impact}} & \textbf{\makecell[c]{Evaluation}} & \textbf{\makecell[c]{Multilingual\\Fact-Checking}} & \textbf{Notes} \\
 \midrule
 
\citet{guo-etal-2022-survey} & \cmark & \cmark & \xmark & \xmark & \xmark & \makecell[l]{Focuses on traditional FC tasks and datasets.\\Does not analyze LLM paradigms, LLM-based techniques, evaluation\\approaches, or multilingual fact-checking.} \\
 
\hline

\citet{augenstein2024factuality} & \xmark & \cmark & \xmark & \xmark & \xmark & \makecell[l]{Focuses on factuality and hallucination.\\Datasets are discussed without a fact-checking task framing.\\Does not cover fact-checking tasks, LLM paradigms, evaluation approaches,\\or multilingual fact-checking.} \\
 
 \hline
 
\citet{wang-etal-2024-factuality} & \xmark & \cmark & \xmark & \cmark & \xmark & \makecell[l]{Emphasizes factuality evaluation and mitigation.\\Techniques are discussed outside fact-checking pipelines.\\Does not address fact-checking tasks, LLM paradigms, or multilingual settings.} \\
 
 \hline
 
\citet{dmonte2025claimverificationagelarge} & \cmark & \cmark & \xmark & \cmark & \xmark & \makecell[l]{Covers LLM as predictors in claim verification.\\Synthetic data generation receives limited attention.\\LLM-based evaluation not systematically analyzed.\\Multilingual aspects are only briefly mentioned.} \\
 
 \hline
 
\citet{rahman2025hallucinationtruthreviewfactchecking} & \xmark & \cmark & \xmark & \cmark & \cmark & \makecell[l]{Primarily focuses on hallucination analysis.\\Datasets and metrics are discussed across tasks.\\Does not provide a structured treatment of fact-checking tasks or paradigms.\\Multilinguality discussed but not systematically.} \\
 
 \hline
 
\citet{https://doi.org/10.1002/aaai.12188} & \xmark & \xmark & \xmark & \cmark & \xmark & \makecell[l]{High-level survey on LLMs and misinformation.\\ Focuses on detection and risks of LLM-generated misinformation.\\No discussion about fact-checking tasks, datasets, or LLM paradigms.\\Limited analysis of techniques or evaluation approaches.\\Multilingual and multimodal aspects only briefly acknowledge.} \\ 
 
 \hline
 
\textbf{Ours} & \cmark & \cmark & \cmark & \cmark & \cmark & \makecell[l]{Provides systematic coverage of fact-checking tasks, datasets,\\ methodological paradigms and techniques, evaluation approaches,\\and multilingual fact-checking.}\\
 \bottomrule
\end{tabular}
}
\label{tab:survey-compasion}
\end{table}

\citet{wang-etal-2024-factuality} present a detailed survey of factuality in LLMs, including taxonomies of factual errors, evaluation datasets and metrics, and mitigation strategies spanning pre-training, fine-tuning, retrieval augmentation, and inference-time techniques. The survey offers valuable insights into improving and measuring factuality in generated content, but its scope extends well beyond fact-checking and is not organized around fact-checking pipelines or tasks. Automated fact-checking is primarily considered a tool for evaluating or improving LLM outputs, rather than as an end-to-end application in which LLMs play multiple roles.

A more direct focus on LLM-based fact-checking is provided by \citet{dmonte2025claimverificationagelarge}, who survey claim verification approaches built on LLMs. This work reviews LLM-based pipelines, retrieval strategies, prompting methods, transfer-learning techniques, and generation strategies for veracity prediction and explanation generation. It also summarizes commonly used datasets, evaluation metrics, and shared tasks. However, the survey primarily emphasizes LLMs as predictors within verification pipelines, with limited coverage of synthetic data generation and only partial discussion of LLMs as evaluators. Multilingualism is mentioned mainly as an open challenge rather than being systematically analyzed.

In addition, \citet{rahman2025hallucinationtruthreviewfactchecking} examine the intersection of hallucination research and fact-checking. Their survey provides a detailed taxonomy of evaluation metrics, discusses datasets used for both hallucination analysis and fact-checking, and reviews standard prompting, fine-tuning, and retrieval-augmented techniques. Nevertheless, the paper is predominantly concerned with hallucination detection and mitigation, and the effects of hallucination on automated fact-checking. Therefore, fact-checking tasks, paradigms, and multilingual considerations receive relatively limited emphasis.

Finally, a complementary perspective is offered by \citet{https://doi.org/10.1002/aaai.12188}, who survey opportunities and challenges of combating misinformation in the age of LLMs. Their study provides a high-level overview of how LLMs can both support misinformation detection and amplify misinformation through large-scale generation. While the survey discusses detection strategies and emerging risks of LLM-generated misinformation, it does not frame the problem in terms of automated fact-checking pipelines. Core fact-checking tasks, benchmark datasets, and LLM paradigms are not systematically analyzed, and techniques are treated at a conceptual level. Multilingual and multimodal aspects are acknowledged as an important direction, but without a structured analysis.

In summary, existing surveys either precede the generative LLM era or focus primarily on factuality and hallucination rather than on automated fact-checking as a holistic process. While individual papers examine specific components such as datasets, evaluation metrics, or LLM-based prediction strategies, none offer a unified analysis of how LLMs are employed across synthetic data generation, downstream fact-checking tasks, and evaluation (see Table~\ref{tab:survey-compasion}). Moreover, multilingual fact-checking and corresponding approaches remain underexplored. These gaps highlight the need for a comprehensive survey that systematically categorizes and compares the roles of generative LLMs throughout the automated fact-checking pipeline, which is the primary objective of our survey.

\subsection{Scope of the Survey}

% What is the focus of this survey paper: 
% (1) Methods and Approaches that are used for fact-checking when applying generative LLMs
% (2) Fact-checking tasks that can be automated by LLMs - there are already some survey papers on that, one or two of them are from KInIT
% (3) Datasets commonly used within the fact-checking and especially when employing LLMs
% (4) Languages covered - how researchers handle the multilingual aspect of fact-checking, are there any papers on other languages than English? Are the low-resource languages also covered? How is the multilinguality or low-resource aspect handled?

% The goal of this survey is to provide a comprehensive overview of how generative LLMs are employed in the automation of fact-checking. Generative LLMs offer new capabilities for understanding, reasoning, and text generation, presenting both opportunities and challenges distinct from those of traditional discriminative models. Unlike general surveys on fact-checking, our objective is to systematically analyze how LLMs are leveraged to support fact-checking, highlighting methodological innovations, practical challenges, and emerging trends.

Our analysis is structured around the key dimensions introduced above and examines how generative LLMs are applied across different stages of automated fact-checking. In particular, we organize the literature around three main use cases that capture the primary roles of generative LLMs identified in the reviewed studies.

The first use case concerns \textit{downstream prediction tasks}, including claim detection, identifying previously fact-checked claims, gathering evidence to verify or refute claims, and directly assessing the veracity of statements. These tasks correspond to the core fact-checking pipeline and relate to the first observed dimension (\textit{Tasks and Challenges}). The second use case involves \textit{synthetic data generation}, where LLMs create high-quality artificial examples to mitigate the scarcity of annotated samples and support model training in low-resource or specialized domains. This use case is closely related to the second dimension (\textit{Datasets and Limitations}). The third use case focuses on \textit{evaluation and quality assurance}, where LLMs are used to assess outputs of fact-checking systems using measures of factuality, consistency, or explainability, corresponding to the dimension of \textit{Evaluation}. Together, these use cases represent the primary roles of generative LLMs observed in the literature.

Beyond these use cases, we review the datasets employed across these downstream tasks, highlighting differences in structure, domain coverage, and linguistic diversity. Dataset characteristics strongly influence model performance, particularly in low-resource or multilingual settings, and synthetic data has emerged as a crucial strategy for addressing data scarcity. The dataset descriptions and their limitations closely align with the \textit{Tasks and Challenges} dimension. In addition, we examine methodological approaches for integrating LLMs into fact-checking pipelines and commonly used techniques, such as prompting strategies, fine-tuning, and adapter-based approaches. These techniques correspond to the dimension of \textit{Methodological Paradigms and techniques}. We further analyze multilingual coverage, focusing on approaches applied to languages beyond English, including low-resource and non-Latin script languages. This analysis addresses the dimension of \textit{Multilingual Fact-Checking}. Finally, we review research trends and discuss open challenges and future directions for LLM-based fact-checking, corresponding to the dimension of \textit{Challenges and Future Directions.}

To maintain a clear focus, we restrict our analysis to studies that primarily consider \textbf{textual or tabular data}, employ \textbf{encoder-decoder and decoder-only models}, and target \textbf{tasks aimed at verifying the truthfulness of claims}. Studies on related problems, such as hate speech detection, propaganda detection, stance detection, or multimodal misinformation, are excluded because of their distinct objectives and methodologies. This scope allows us to provide a focused overview of how generative LLMs contribute to automating textual fact verification.

\section{Methodology}

\subsection{Collection Process}

\paragraph{Publications Sources.}

\begin{figure}
    \centering
    \includegraphics[width=1\linewidth]{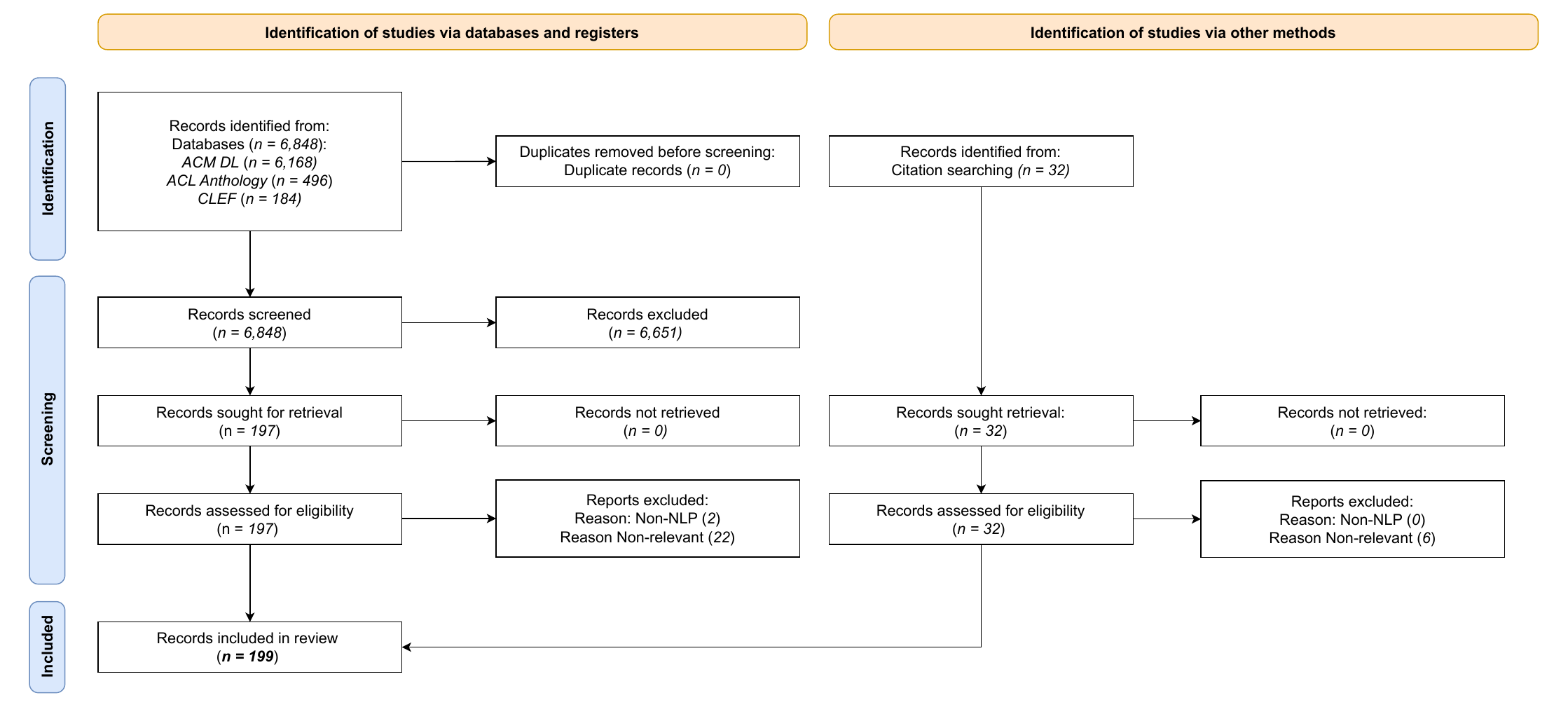}
    \caption{PRISMA 2020 flow diagram illustrating the standardized methodology for collecting relevant papers, together with the number of papers processed in each step.}
    \label{fig:prisma-diagram}
\end{figure}

To identify and collect relevant papers, we followed a standardized methodology for literature reviews based on the \textit{Preferred Reporting Items for Systematic Reviews and Meta-Analyses} methodology \textit{(PRISMA)}~\cite{Pagen71}. We conducted a systematic search across well-known research repositories that publish work in the NLP domain and artificial intelligence. Our primary sources included the \textit{ACL Anthology}\footnote{\url{https://aclanthology.org/}} and \textit{ACM Digital Library}\footnote{\url{https://dl.acm.org/}}, which are among the most used archives for peer-reviewed publications in NLP and computational linguistics. Furthermore, we incorporated publications from the \textit{CLEF CheckThat!}\footnote{\url{https://checkthat.gitlab.io/}} shared tasks, as this initiative is well-known in the fact-checking community, particularly for multilingual and cross-lingual misinformation detection.

In addition, we expanded our collection by incorporating the papers cited in the related work. Specifically, we examined the reference lists of retrieved papers and identified subsequent papers that cited them, allowing us to include both seminal and recent contributions. Figure~\ref{fig:prisma-diagram} shows the standardized PRISMA 2020 flow diagram~\cite{Pagen71} detailing each step of the review process along with the number of research papers processed. 

Overall, we collected 199 papers that employ generative LLMs for fact-checking or for generating synthetic data to support fact-checking tasks. These works cover a wide range of applications, including claim detection, evidence retrieval, previously fact-checked claim retrieval, fact verification, and fake news detection. A subset of the papers focuses on improving evaluation pipelines or developing novel metrics to assess the factuality and reliability of LLM outputs. The collected studies also vary in scope, with some addressing domain-specific datasets, while others investigate multilingual or cross-lingual fact-checking scenarios. Together, this collection provides a comprehensive overview of the current research landscape at the intersection of generative LLMs and automated fact-checking.

\paragraph{Keywords and Search Strategy.}

Based on prior surveys and the commonly defined subtasks of the fact-checking pipeline~\cite{thorne-vlachos-2018-automated, ijcai2021p0619, guo-etal-2022-survey}, we designed a keyword-based search strategy. The selected keywords, listed in Table~\ref{tab:search-keywords}, were iteratively refined during data collection to maximize coverage of the retrieved results. We searched for each keyword within the title, abstract, and the keywords of the databases. The retrieved papers were reviewed using a two-step approach. First, an automatic filter using LLMs was applied to remove irrelevant papers. Second, the remaining papers were manually reviewed to verify relevance and quality.

\begin{table}[t]
    \centering
    \caption{A curated list of search keywords used to retrieve research papers on fact-checking. Keywords are grouped by task to cover generic topics and individual fact-checking tasks.}
    \label{tab:search-keywords}
    \small
    \begin{tabularx}{\textwidth}{l|X}
    \toprule
    \textbf{Tasks} & \multicolumn{1}{c}{\textbf{Search keywords}} \\
    \midrule
    \multirow{1}{*}{Generic}  & \textit{fact-checking}; \textit{fact-checking dataset}; \textit{disinformation}; \textit{misinformation}\\
    \hline
    \multirow{1}{*}{Claim detection} & \textit{check-worthy claim detection}; \textit{claim detection}; \textit{claim normalization}\\
    \hline
    \multirow{1}{*}{\makecell[l]{Previously fact-checked\\claim retrieval}} & \textit{previously fact-checked claim retrieval}; \textit{previously fact-checked claim detection}; \textit{fact-checked claim detection}; \textit{claim-matching}; \textit{claim matching}; \textit{verified claim retrieval}\\
    \hline
    \multirow{1}{*}{Evidence retrieval} & \textit{evidence retrieval}; \textit{rationale selection}\\
    \hline
    \multirow{1}{*}{\makecell[l]{Fact verification \&\\fake news detection}} & \textit{factuality prediction}; \textit{claim verification}; \textit{fact verification}; \textit{fake-news detection}; \textit{justification prediction}; \textit{explanation generation}; \textit{disinformation detection}; \textit{misinformation detection}; \textit{rumour verification}; \textit{rumour detection}; \textit{rumor verification}; \textit{rumor detection}; \textit{veracity prediction}; \textit{truth assessment}\\
    \bottomrule
    \end{tabularx}
\end{table}

\paragraph{Exclusion Criteria.}

To maintain a clear focus on our objectives, we applied exclusion criteria to filter papers that did not meet the topical or methodological constraints of our survey. Each retrieved paper was reviewed through a two-step process: an initial screening based on the title and abstract, followed by a full-text inspection when relevance was uncertain. Duplicate entries across multiple databases were identified and filtered to avoid redundancy.

We excluded studies that:

\begin{itemize}
    \item Did not involve generative LLMs in synthetic data generation, prediction, or LLM-based evaluation (e.g., including only encoder-only models).
    \item Addressed tasks outside the fact-checking pipeline (such as hate speech detection, propaganda, or toxicity detection).
    \item Focused on multimodal or only non-textual inputs (e.g., image, audio, or video-based misinformation detection).
    \item Used relevant search keywords in unrelated contexts (e.g., fact-checking in the context of hallucination generation).
    \item Were classified as posters, abstracts, or works in progress in ACM publications, as these provide limited methodological details or preliminary results.
\end{itemize}

These criteria were meant to ensure that our final set of selected papers is directly relevant to the survey's objectives, namely the application of generative LLM to the automation of textual fact-checking tasks. %By enforcing these criteria, we aimed to provide a coherent, methodologically rigorous, and reliable overview of current research in automated fact-checking using generative models.

\subsection{Manual Inspection Method}

To systematically analyze the collected papers, we conducted a manual inspection guided by the survey dimensions introduced earlier. First, we examined the specific \textit{fact-checking tasks} addressed in each paper, including how these tasks were defined, the datasets used, evaluation metrics applied, and languages covered. These findings are summarized in Section~\ref{sec:fact-checking-tasks}, where we outline four core fact-checking tasks.

Next, we analyzed the \textit{methods and approaches} employed across studies, focusing on how generative LLMs were utilized, either as synthetic data generators, predictors, evaluators, or as components of multi-step fact-checking pipelines. This analysis is presented in Section~\ref{sec:methods-taxonomy}, where we introduce our taxonomy of use cases in fact-checking.

During this manual review, we also noted recurring patterns and qualitative insights, including common challenges in low-resource languages, the growing use of synthetic data to address data scarcity, and emerging trends in evaluation practices. These observations informed the structure of our taxonomy and guided our discussion of methodological and linguistic gaps.

\section{Tasks in Automated Fact-Checking}
\label{sec:fact-checking-tasks}

\begin{figure}
    \centering
    \includegraphics[width=1\linewidth]{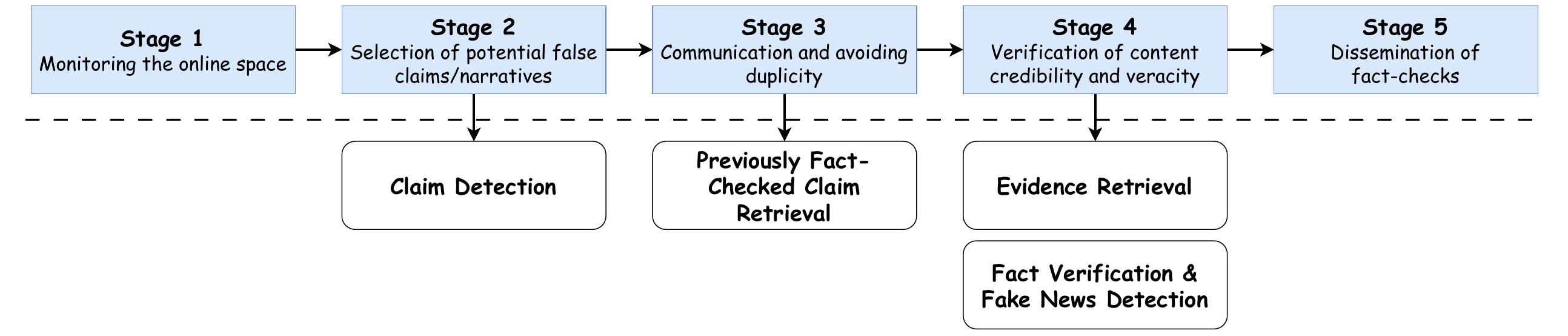}
    \caption{Overview of the fact-checking pipeline proposed by \citet{10.1145/3764592} and the mapping of specific fact-checking tasks to each stage.}
    \label{fig:fc-pipeline}
\end{figure}

Automated fact-checking is a complex multi-stage process designed to assess the veracity of information, thereby helping mitigate the spread of misinformation~\cite{guo-etal-2022-survey}. At its core, the process involves (1) detecting potentially misleading claims, (2) retrieving supporting or refuting evidence, (3) verifying statements against trustworthy sources, and (4) generating explanations or full fact-checking articles~\cite{sahnan2025llmsautomatefactcheckingarticle}. Each stage presents unique challenges, including addressing diverse domains, multiple languages, and noisy or incomplete sources, while also maintaining interpretability and minimizing bias in automated decisions.

Figure~\ref{fig:fc-pipeline} illustrates a fact-checking pipeline, based on interviews with the fact-checkers done by~\citet{10.1145/3764592}, and highlights how different tasks in automated fact-checking map to specific stages. The figure shows the division between tasks traditionally performed by human fact-checkers and those that can be supported or fully automated using AI systems. This mapping provides a structured view that motivates our organization of this survey. We categorize tasks according to their role in the pipeline and the extent to which generative LLMs have been applied to automate them.

Following this structure, we focus on four tasks that encompass the majority of automated fact-checking research: \textit{claim detection}, \textit{previously fact-checked claim retrieval}, \textit{evidence retrieval}, and \textit{fact-verification and fake news detection}. These tasks are not only central to automated workflows but also reflect the increasing advancement of modern models, which can move beyond simple classification or retrieval to generate explanations and comprehensive, human-readable fact-checking articles.

The following subsections provide a detailed description of each task, including definitions, recent advances in generative LLMs, the datasets used, and inherent limitations. With this structure, we aim to provide a clear overview of current capabilities, open challenges, and opportunities for future research in automated fact-checking.

\subsection{Claim Detection}

\begin{figure*}
\centering
\resizebox{\textwidth}{!}{
\begin{forest}
			forked edges,
			for tree={
				grow=south,
				% reversed=true,
				% anchor=base west,
				% parent anchor=east,
				% child anchor=west,
                node options={align=center},
                align=center,
				base=left,
				font=\small,
				rectangle,
				draw=hidden-draw,
				rounded corners,
				minimum width=4em,
				edge+={darkgray, line width=1pt},
				s sep=3pt,
				inner xsep=2pt,
				inner ysep=3pt,
				ver/.style={rotate=90, child anchor=north, parent anchor=south, anchor=center},
			},
			where level=1{text width=5.0em,font=\scriptsize}{},
			where level=2{text width=5.6em,font=\scriptsize}{},
			where level=3{text width=6.8em,font=\scriptsize}{},
			[
			Claim Detection, text width=10em
			[
			Classification, text width=5em
			[
			Binary\\Classification, text width=5em
                [
                    \textit{\# of Papers: 17}\\
                    \textit{English-only: 7}\\
                    \textit{Non-English: 1}\\
                    \textit{Multilingual: 9}
                    ,leaf-center, text width=5em
                ]
       %          [
       %          {\citet{gangi-reddy-etal-2022-newsclaims},}
       %          {\citet{agrestia2022polimi},}
       %          {\citet{du2022nus},}\\
       %          {\citet{sawinski2023openfact},}
       %          {\citet{hyben2023bigger},}
       %          {\citet{cao2023large},}
       %          {\citet{setty2024endtoend},}\\
       %          {\citet{ni-etal-2024-afacta},}
       %          {\citet{vykopal-etal-2025-soft},}
       %          {\citet{aarnes2024iaigroupcheckthat2024},}
       %          {\citet{eren2024turquaz},}\\
       %          {\citet{golik2024dshacker},}
       %          {\citet{gruman2024clac},}
       %          {\citet{yufeng2024factfinders},}\\
       %          {\citet{weering2024fc_rug},}
       %          {\citet{10.1145/3543507.3583870},}
       %          {\citet{10.1145/3654777.3676359}}
    			% , leaf, text width=20em, align = left
       %          ]
			]
            [
			Multiclass\\Classification, text width=5em
       %          [
       %          {\citet{wang-etal-2024-factcheck},}
       %          {\citet{10.1145/3689212}}
    			% , leaf, text width=20em, align = left
       %          ]
                [
                    \textit{\# of Papers: 2}\\
                    \textit{English-only: 2}\\
                    \textit{Non-English: 0}\\
                    \textit{Multilingual: 0}
                    ,leaf-center, text width=5em
                ]
			]
			[
			Multilabel\\Classification, text width=5em
                [
                    \textit{\# of Papers: 1}\\
                    \textit{English-only: 0}\\
                    \textit{Non-English: 0}\\
                    \textit{Multilingual: 1}
                    ,leaf-center, text width=5em
                ]
       %          [
       %          {\citet{agrestia2022polimi}}
    			% , leaf, text width=20em, align = left
       %          ]
			]
			]
            [
            Span\\Detection, text width=5em
                [
                    \textit{\# of Papers: 3}\\
                    \textit{English-only: 2}\\
                    \textit{Non-English: 0}\\
                    \textit{Multilingual: 1}
                    ,leaf-center, text width=5em
                ]
       %          [
       %          {\citet{gangi-reddy-etal-2022-zero},}
       %          {\citet{mittal-etal-2023-lost}}
    			% , leaf, text width=29.1em, align = left
       %          ]
            ]
            [
    			Claim\\Normalization, text width=5em
                [
                    \textit{\# of Papers: 11}\\
                    \textit{English-only: 2}\\
                    \textit{Non-English: 0}\\
                    \textit{Multilingual: 9}
                    ,leaf-center, text width=5em
                ]
           %          [
           %          {\citet{sundriyal2023chaos},}
           %          {\citet{li-etal-2024-self},} % newly added
           %          {\citet{amatya2025factiverse},}
           %          {\citet{anikina2025dfkinit2b},}
           %          {\citet{beltran2025umuteam},}\\
           %          {\citet{hashmi2025investigators},}
           %          {\citet{almada2025akcit},}
           %          {\citet{pramov2025ds},}
           %          {\citet{saeed2025mma},}
           %          {\citet{sawinski2025openfact},}\\
           %          {\citet{vineetha2025saivineetha},}
           %          {\citet{wilder2025unh}}
        			% , leaf, text width=29.1em, align = left
           %          ]
    			]
                [
    			Claim\\Decomposition, text width=5em
                [
                    \textit{\# of Papers: 9}\\
                    \textit{English-only: 8}\\
                    \textit{Non-English: 0}\\
                    \textit{Multilingual: 1}
                    ,leaf-center, text width=5em
                ]
           %          [
           %          {\citet{kamoi-etal-2023-wice},}
           %          {\citet{li-etal-2024-self},}
           %          {\citet{ni-etal-2024-afacta},}
           %          {\citet{heil2025ds},}
           %          {\citet{strong-etal-2024-zero},}\\
           %          {\citet{hu-etal-2025-decomposition},}
           %          {\citet{10.1145/3654777.3676359}}
        			% , leaf, text width=29.1em, align = left
           %          ]
    			] 
			[
			  Question\\Generation, text width=5em
              [
                    \textit{\# of Papers: 7}\\
                    \textit{English-only: 7}\\
                    \textit{Non-English: 0}\\
                    \textit{Multilingual: 0}
                    ,leaf-center, text width=5em
                ]
       %        [
       %          {\citet{chen-etal-2022-generating},}
       %          {\citet{10.1145/3539618.3591907},}
       %          {\citet{zhang-gao-2024-reinforcement},}
       %          {\citet{chen-etal-2024-complex},}
       %          {\citet{10.1145/3696410.3714748},}\\
       %          {\citet{10.1145/3626772.3657874},}
       %          {\citet{10.1145/3627673.3679985}}
    			% , leaf, text width=29.1em, align = left
       %          ]
			]
            [
                Others, text width=5em
                [
                    \textit{\# of Papers: 5}\\
                    \textit{English-only: 5}\\
                    \textit{Non-English: 0}\\
                    \textit{Multilingual: 0}
                    ,leaf-center, text width=5em
                ]
                % [
                %     {\citet{deng-etal-2024-document},}
                %     {\citet{kotitsas-etal-2024-leveraging},}
                %     {\citet{eren2024turquaz}}
                %     , leaf, text width=29.1em, align = left
                % ]
            ]
			]
		\end{forest}
}
\caption{Taxonomy of approaches for claim detection. The diagram organizes studies by task type and shows the number of papers for each approach, further distinguished by language coverage. Approaches applied only to English are labeled as English-only, those applied to a single non-English language are labeled non-English, and approaches evaluated on two or more languages are labeled multilingual.}
\label{fig:taxonomy-claim-detection}
\end{figure*}

\paragraph{Definition.} 

\textit{Claim detection} is a fundamental task in automated fact-checking that focuses on identifying statements within text that contain verifiable information and may require verification~\cite{thorne-vlachos-2018-automated, ijcai2021p0619}. This task is also known as \textit{check-worthy claim detection}~\cite{10.1145/3539618.3591907}, \textit{factual claim detection}~\cite{ni-etal-2024-afacta} or \textit{propositional claim detection}~\cite{nenno2024propositional}. The statements are often referred to as \textit{check-worthy} or \textit{verifiable} claims, and they can encompass a wide range of content, from factual assertions in news articles~\cite{gangi-reddy-etal-2022-newsclaims} to statements made on social media platforms~\cite{sundriyal2023chaos, anikina2025dfkinit2b, beltran2025umuteam}. The importance of claim detection lies in its role as the first step in the fact-checking pipeline, where, after accurately identifying claims, subsequent verification, evidence retrieval, and explanation tasks can be applied efficiently. Moreover, with the growing volume of information across media platforms such as social media and online news portals, it has become essential to carefully select which statements to verify based on their potential societal impact and the possible harm they could cause if spread.

\paragraph{LLM Applications.}

Recent approaches leverage generative LLMs not only for \textbf{classification} but also for more advanced operations such as \textbf{span detection}~\cite{gangi-reddy-etal-2022-zero}, \textbf{claim normalization}~\cite{sundriyal2023chaos}, \textbf{decomposition}~\cite{kamoi-etal-2023-wice}, or \textbf{question generation}~\cite{chen-etal-2022-generating} as shown in Figure~\ref{fig:taxonomy-claim-detection}. While many studies have approached claim detection as a binary classification problem, recent generative approaches have demonstrated the capability of LLMs to perform higher-level reasoning. LLMs can generate alternative phrasings of complex claims~\cite{metropolitansky-larson-2025-towards, zhao-etal-2024-pacar}, highlight embedded sub-claims~\cite{kamoi-etal-2023-wice, ni-etal-2024-afacta, hu-etal-2025-decomposition}, normalize noisy social media posts into concise verifiable claims~\cite{sundriyal2023chaos, amatya2025factiverse, anikina2025dfkinit2b, hashmi2025investigators}, or transform statements into structured questions suitable for automated verification~\cite{chen-etal-2022-generating, 10.1145/3539618.3591907, chen-etal-2024-complex}. These generative capabilities enhance a system's ability to capture subtle or nuanced assertions that traditional methods may overlook, particularly in low-resource or multilingual settings.

\paragraph{Datasets.}

\begin{table}[]
    \caption{Overview of datasets for claim detection. The table lists each dataset along with its output type, number of languages, and dataset size (in thousands of instances).}
    \label{tab:checkworthy-datasets}
    \small
    \centering
    \begin{tabularx}{\textwidth}{Xlrrl}
        \toprule
        \textbf{Name} & \textbf{Output Type} & \textbf{\# Langs} & \textbf{Size [K]} & \textbf{Citation} \\
        \midrule
        BioClaims & 2 classes & 1 & 1.2 & \citet{wuhrl-klinger-2021-claim} \\
        ClaimBuster & 3 classes & 1 & 24 & \citet{arslan2020benchmark} \\
        ClaimDecomp & Questions & 1 & 1.2 & \citet{chen-etal-2022-generating} \\
        CLAN & Normalized claim & 1 & 6 & \citet{sundriyal2023chaos} \\
        CLEF-2020 & Ranked list & 2 & 13 & \citet{barron2020checkthat} \\
        CLEF-2021 & Ranked list & 5 & 22 & \citet{nakov2021overview} \\
        CLEF-2022 & 2 classes & 6 & 30.3 / 22.9 /22.8 & \citet{nakov2022overview} \\
        CLEF-2023 & 2 classes & 4 & 70.8 & \citet{alam2023overview} \\
        CLEF-2024 & 2 classes & 4 & 66.6 & \citet{barron2024clef} \\
        CLEF-2025 & Normalized claim & 20 & 77.8 & \citet{alam2025overview} \\
        CW-CURE & 2 classes & 1 & 3.3 & \citet{10.1145/3543507.3583870}\\
        EnvClaims & 2 classes & 1 & 2.7 & \citet{stammbach-etal-2023-environmental} \\
        HCN Dataset & Major Claim & 1 & 0.9 & \citet{kotitsas-etal-2024-leveraging}\\
        LESA-2021 & 2 classes & 1 & 379 & \citet{gupta-etal-2021-lesa} \\
        % Monant & 2 classes & 1 & 51 & \citet{10.1145/3477495.3531726} \\
        MultiClaim & 2 classes / Normalized claim & 27/39 & 28 & \citet{pikuliak-etal-2023-multilingual} \\
        NewsClaims & Spans & 1 & 0.9 & \citet{gangi-reddy-etal-2022-newsclaims} \\
        PoliClaim & 2 classes & 1 & 12.4 & \citet{ni-etal-2024-afacta} \\
        X-CLAIM & Spans & 6 & 7 & \citet{mittal-etal-2023-lost} \\
        \bottomrule
    \end{tabularx}
\end{table}

Claim detection datasets, summarized in Table~\ref{tab:checkworthy-datasets}, vary in domain, language, annotation scheme, and granularity of supervision. Early English-only datasets such as \textit{ClaimBuster}~\cite{arslan2020benchmark}, \textit{BioClaims}~\cite{wuhrl-klinger-2021-claim}, or \textit{CW-CURE}~\cite{10.1145/3543507.3583870} framed claim detection as binary or ternary sentence-level classification. While these tasks favored traditional classifiers, generative LLMs can naturally replace them by reframing detection as instruction-following, where models assess check-worthiness based on contextual and linguistic cues.

Much of the field is driven by the \textit{CLEF CheckThat! Lab} datasets~\cite{barron2020checkthat, nakov2021overview, nakov2022overview, alam2023overview, barron2024clef, alam2025overview}, evolving from ranked lists and binary labels ~\cite{nakov2022overview} to generative formulations like \textit{claim normalization}~\cite{alam2025overview}, where models convert noisy social media posts into concise, verifiable statements. This task formulation inherently benefits from generative LLMs, which can rephrase, condense, or disentangle complex user-generated text into statements suitable for further verification.

Beyond classification, datasets such as \textit{ClaimDecomp}~\cite{chen-etal-2022-generating} require decomposing complex claims into atomic sub-questions, encouraging reasoning over intermediate representations. Span-level datasets such as \textit{NewsClaims}~\cite{gangi-reddy-etal-2022-newsclaims} and \textit{X-CLAIM}~\cite{mittal-etal-2023-lost} shift the focus from classifying entire sentences to identifying the precise segment of text that constitutes the check-worthy claim. These segments are typically annotated with start-end character or token indices within a larger passage. This structure aligns naturally with extractive and generative LLM frameworks, where models are prompted to locate and extract the relevant spans rather than make coarse sentence-level predictions.

Multilingual datasets, such as \textit{MultiClaim}~\cite{pikuliak-etal-2023-multilingual} and the latest \textit{CLEF CheckThat!} iterations~\cite{nakov2022overview,alam2025overview}, enable multilingual and cross-lingual research and highlight LLMs' capacity to generalize across languages and scripts. These resources reveal how models handle linguistically diverse misinformation and leverage pretraining to detect claims in low-resource languages.

% Overall, claim detection datasets (Table~\ref{tab:checkworthy-datasets}) have progressed from simple classification~\cite{wuhrl-klinger-2021-claim} toward generative and reasoning-based formulations~\cite{chen-etal-2022-generating, sundriyal2023chaos}. This evolution mirrors the broader shift in automated fact-checking, emphasizing the language-aware, context-sensitive capabilities of LLMs that can transform, decompose, and identify claims across languages and domains.

\paragraph{Limitations.}

While LLMs have advanced claim detection beyond traditional classifiers, they face task-specific limitations. LLMs can misinterpret nuanced languages, sarcasm, or rhetorical questions as verifiable claims, particularly in informal or noisy text from social media~\cite{zeng2025worsezeroshotfactcheckingdataset}. They are also sensitive to surface patterns, which can lead to missing implicit or context-dependent claims. Generative outputs may introduce hallucinations of overgeneralizations when normalizing or decomposing claims~\cite{sundriyal2023chaos, kamoi-etal-2023-wice}, reducing reliability for downstream verification. Moreover, LLMs can inherit biases from pretraining corpora, affecting the detection of claims in underrepresented languages, domains, or cultural contexts~\cite{ni-etal-2024-afacta}, and raising ethical concerns around automation bias in high-stakes fact-checking scenarios.

\subsection{Previously Fact-Checked Claim Retrieval}

\begin{figure*}
\centering
\resizebox{0.45\textwidth}{!}{
\begin{forest}
			forked edges,
			for tree={
				grow=south,
                node options={align=center},
                align = center,
				base=left,
				font=\small,
				rectangle,
				draw=hidden-draw,
				rounded corners,
				minimum width=4em,
				edge+={darkgray, line width=1pt},
				s sep=3pt,
				inner xsep=2pt,
				inner ysep=3pt,
				ver/.style={rotate=90, child anchor=north, parent anchor=south, anchor=center},
			},
			where level=1{text width=5.0em,font=\scriptsize}{},
			where level=2{text width=5.6em,font=\scriptsize}{},
			where level=3{text width=6.8em,font=\scriptsize}{},
			[
			Previously Fact-Checked\\Claim Retrieval, text width=10em
			[
			Ranking, text width=5em
                [
                    \textit{\# of Papers: 3}\\
                    \textit{English-only: 3}\\
                    \textit{Non-English: 0}\\
                    \textit{Multilingual: 0}
                    ,leaf-center, text width=5em
                ]
       %          [
       %          {\citet{shliselberg2022riet},}
       %          {\citet{singh2023utdrm},}
       %          {\citet{neumann-etal-2023-deep}}
    			% , leaf, text width=29.1em, align = left
       %          ]
			]
            [
            Classification, text width=5em
                [
    			Binary\\Classification, text width=5em
                    [
                        \textit{\# of Papers: 2}\\
                        \textit{English-only: 1}\\
                        \textit{Non-English: 0}\\
                        \textit{Multilingual: 1}
                        ,leaf-center, text width=5em%, align = left
                    ]
           %          [
           %          {\citet{pisarevskaya-zubiaga-2025-zero},}
           %          {\citet{vykopal2025largelanguagemodelsmultilingual}}
        			% , leaf, text width=20em, align = left
           %          ]
    			]
                [
    			3-class\\Classification, text width=5em
                [
                    \textit{\# of Papers: 2}\\
                    \textit{English-only: 2}\\
                    \textit{Non-English: 0}\\
                    \textit{Multilingual: 0}
                    ,leaf-center, text width=5em%, align = left
                ]
           %          [
           %          {\citet{choi2023automated},}
           %          {\citet{choi2024factgpt}}
        			% , leaf, text width=20em, align = left
           %          ]
    			]
            ]
			]
		\end{forest}
}

\caption{Taxonomy of approaches for claim detection. The diagram organizes studies by task type and shows the number of papers for each approach, further distinguished by language coverage (English-only, non-English, and multilingual).}
\label{fig:taxonomy-claim-matching}
\end{figure*}

\paragraph{Definition.}

The task of \textit{previously fact-checked claim retrieval} has been referred to in the literature under several names, including \textit{fact-checked claims detection}~\cite{shaar-etal-2020-known}, \textit{fact-checking URL recommendation}~\cite{10.1145/3209978.3210037}, \textit{verified claim retrieval}~\cite{10.1007/978-3-030-58219-7_17}, \textit{searching for fact-checked information}~\cite{vo-lee-2020-facts}, or \textit{claim matching}~\cite{kazemi-etal-2021-claim}. This task aims to reduce redundant work for fact-checkers by identifying whether a claim has already been verified, and to look for similar debunked claims~\cite{ijcai2021p0619}. It plays a crucial role in fact-checking workflows, as misleading or false claims often reappear across different languages, platforms, and time periods. Therefore, the objective is to retrieve semantically similar claims or fact-checks to quickly locate prior verifications, rather than re-evaluating the same or slightly modified claims.

\paragraph{LLM Applications.}

Recent research employing generative LLMs for this task remains scarce, with only a few studies exploring their use (see Figure~\ref{fig:taxonomy-claim-matching}). Most approaches employ LLMs to re-rank or filter fact-checks retrieved by traditional information retrieval systems~\cite{shliselberg2022riet, vykopal2025largelanguagemodelsmultilingual}, such as BM25~\cite{INR-019} or dense passage retrieval~\cite{karpukhin-etal-2020-dense}. Others reformulate the task as a form of \textbf{textual entailment}, using LLMs to determine whether the input claim and a retrieved claim express equivalent or contradictory information~\cite{choi2023automated, choi2024factgpt}. Moving beyond simple textual entailment, researchers also focused on \textbf{classifying the relevance} between claims and retrieved fact-checked claims, to identify which fact-checks can be used to verify the information within the input claim~\cite{pisarevskaya-zubiaga-2025-zero, vykopal2025largelanguagemodelsmultilingual}.

\paragraph{Datasets.}

\begin{table}[]
    \caption{Overview of datasets used for previously fact-checked claim detection. The table summarizes each dataset by output type, number of supported languages, number of input claims, fact-checked claims, and claim–fact-check pairs (in thousands).}
    \label{tab:claim-matching-datasets}
    \small
    \centering
    \resizebox{\textwidth}{!}{%
    \begin{tabular}{llrrrrl}
    \toprule
        \textbf{Name} & \textbf{Output Type} & \textbf{\# Langs} & \textbf{\# Claims [K]} & \textbf{\# FC Claims [K]} & \textbf{\# Pairs [K]} & \textbf{Citation}\\
        \midrule
        AMC-16K & 2 classes & 20 & - & - & 16 & \citet{vykopal2025largelanguagemodelsmultilingual} \\
        ClaimMatch & 2 classes & 1 & - & - & 2 & \citet{pisarevskaya-zubiaga-2025-zero} \\
        CLEF-2020 & Ranked list & 1 & 1.2 & 10.4 & 1.2 & \citet{barron2020checkthat} \\
        CLEF-2021 & Ranked list & 2 & 2.9 & 63.4 & 3.2 & \citet{nakov2021overview} \\
        CLEF-2022 & Ranked list & 2 & 3.3 & 65 & 3.6 & \citet{nakov2022overview2} \\
        Data Common & - & X & N/A & N/A & N/A & - \\
        MultiClaim & Ranked list & 27/39 & 28 & 205.8 & 31.3 & \citet{pikuliak-etal-2023-multilingual} \\
        PolitiFact & - & 1 & N/A & N/A & N/A & - \\
        Snopes & - & 1 & N/A & N/A & N/A & - \\
        \bottomrule
    \end{tabular}
    }
\end{table}

Datasets for previously fact-checked claim detection are typically constructed by aligning user-generated content, such as social media posts or news headlines, with existing verified claims from fact-checking organizations~\cite{barron2020checkthat, pikuliak-etal-2023-multilingual}. Table~\ref{tab:claim-matching-datasets} summarizes the main datasets, indicating their multilingual coverage, the number of claims and fact-checked claims, and resulting aligned pairs.

Early resources, such as those from the \textit{CheckThat!} shared tasks~\cite{barron2020checkthat, nakov2021overview, nakov2022overview2}, focused on claim retrieval, where input claims are paired with lists of verified claims. Systems are evaluated based on their ability to rank relevant fact-checks highly using standard retrieval metrics (e.g., NDCG@k, MAP@k)~\cite{pikuliak-etal-2023-multilingual}. Generative LLMs can leverage this ranking structure by framing retrieval as a \textit{natural language inference} (NLI)~\cite{choi2023automated}.

Later datasets, such as \textit{ClaimMatch}~\cite{pisarevskaya-zubiaga-2025-zero} and \textit{AMC-16K}~\cite{vykopal2025largelanguagemodelsmultilingual}, provide explicit claim-fact-check pairs for binary classification, indicating whether a candidate fact-check is relevant. This structure is particularly suitable for LLM prompting and reasoning, complementing ranking-based retrieval.

Large-scale multilingual datasets, including \textit{MultiClaim}~\cite{pikuliak-etal-2023-multilingual}, extend the evaluation to 27-39 languages and multiple domains, with over 30,000 aligned pairs. These datasets enable cross-lingual retrieval and transfer learning, where generative LLMs can exploit multilingual representations to align semantically equivalent claims across languages, paraphrase, and reason across scripts.

% Overall, as shown in Table~\ref{tab:claim-matching-datasets}, the evolution of claim-matching datasets, from ranking-based retrieval to binary relevance and multilingual alignment, reflects a shift toward tasks that benefit from LLMs' generative and language-aware capabilities.

\paragraph{Limitations.}

Generative LLMs applied to previously fact-checked claim retrieval face several challenges. They often struggle to recognize paraphrased, implicit, or context-dependent equivalences between claims and fact-checks, leading to missed matches or false positives~\cite{pikuliak-etal-2023-multilingual}. Cross-lingual and multilingual retrieval remains difficult, as LLMs must align semantically equivalent claims across languages and scripts while accounting for cultural nuances. LLMs can also be sensitive to surface-level patterns, producing inconsistent relevance judgments when input claims are noisy or contain subtle modifications~\cite{vykopal2025largelanguagemodelsmultilingual}. Finally, generative outputs may lack transparency, making it hard to justify why a claim was matched to a particular fact-check, which limits their trustworthiness in real-world fact-checking workflows.

\subsection{Evidence Retrieval}

\begin{figure*}
\centering
\resizebox{\textwidth}{!}{
\begin{forest}
			forked edges,
			for tree={
				grow=south,
                node options={align=center},
                align = center,
				base=left,
				font=\small,
				rectangle,
				draw=hidden-draw,
				rounded corners,
				minimum width=4em,
				edge+={darkgray, line width=1pt},
				s sep=3pt,
				inner xsep=2pt,
				inner ysep=3pt,
				ver/.style={rotate=90, child anchor=north, parent anchor=south, anchor=center},
			},
			where level=1{text width=5.0em,font=\scriptsize}{},
			where level=2{text width=5.6em,font=\scriptsize}{},
			where level=3{text width=6.8em,font=\scriptsize}{},
			[
			Evidence Retrieval, text width=10em
			[
			Ranking, text width=5em
			[
			Classification\\Ranking, text width=5em
                [
                    \textit{\# of Papers: 5}\\
                    \textit{English-only: 5}\\
                    \textit{Non-English: 0}\\
                    \textit{Multilingual: 0}
                    ,leaf-center, text width=5em
                ]
       %          [
       %          {\citet{10.1145/3404835.3463120},}
       %          {\citet{sarrouti-etal-2021-evidence-based},}
       %          {\citet{jiang-etal-2021-exploring-listwise},}\\
       %          {\citet{FERNANDEZPICHEL2022105211},}
       %          {\citet{pradeep2020scientific}}
    			% , leaf, text width=20em, align = left
       %          ]
			]
                [
                    \textit{\# of Papers: 1}\\
                    \textit{English-only: 1}\\
                    \textit{Non-English: 0}\\
                    \textit{Multilingual: 0}
                    ,leaf-center, text width=5em%, align = left
                ]
            % [
            %     {\citet{pasin2024seupd}}
            %     , leaf, text width=29.1em, align = left
            % ]
			]
            [
            Query\\Generation, text width=5em
                [
                    \textit{\# of Papers: 3}\\
                    \textit{English-only: 2}\\
                    \textit{Non-English: 0}\\
                    \textit{Multilingual: 1}
                    ,leaf, text width=5em%, align = left
                ]
           %      [
           %          {\citet{prietochavana2023automated},}
           %          {\citet{abdallah2025ngu_research}}
        			% , leaf, text width=29.1em, align = left
           %      ]
            ]
			% [
			%   Rationale\\Selection, text width=5em
   %              [
   %                  \textit{\# of Papers: 7}\\
   %                  \textit{English-only: 7}\\
   %                  \textit{Non-English: 0}\\
   %                  \textit{Multilingual: 0}
   %                  ,leaf-center, text width=5em%, align = left
   %              ]
   %              % [
   %              %     {\citet{pradeep2020scientific},}
   %              %     {\citet{kamoi-etal-2023-wice},}
   %              %     {\citet{wang-etal-2023-check-covid},}
   %              %     {\citet{10.1145/3565287.3617630},}
   %              %     {\citet{tan2023evidencebased},}\\
   %              %     {\citet{ko-etal-2023-claimdiff},}
   %              %     {\citet{fukuoka2025ksu}}
   %              %     , leaf, text width=29.1em, align = left
   %              % ]
			% ]
            [
            Question\\Generation, text width=5em
                [
                    \textit{\# of Papers: 4}\\
                    \textit{English-only: 2}\\
                    \textit{Non-English: 0}\\
                    \textit{Multilingual: 2}
                    ,leaf-center, text width=5em%, align = left
                ]
           %      [
           %          {\citet{schlichtkrull2023averitec},}
           %          {\citet{pan2023qacheck},}
           %          {\citet{setty2024endtoend},}
           %          {\citet{arana-catania-etal-2022-natural},}
           %          {\citet{le2025lis}}
        			% , leaf, text width=29.1em, align = left
           %      ]
            ]
            [
            Summary\\Generation, text width=5em
                [
                    \textit{\# of Papers: 3}\\
                    \textit{English-only: 2}\\
                    \textit{Non-English: 0}\\
                    \textit{Multilingual: 1}
                    ,leaf-center, text width=5em%, align = left
                ]
           %      [
           %          {\citet{zeng-etal-2024-ru22fact},}
           %          {\citet{chen-etal-2024-complex},}
           %          {\citet{li-etal-2025-imrrf},}
        			% , leaf, text width=29.1em, align = left
           %      ]
            ]
            [
                Others, text width=5em
                [
                    \textit{\# of Papers: 4}\\
                    \textit{English-only: 3}\\
                    \textit{Non-English: 0}\\
                    \textit{Multilingual: 1}
                    ,leaf, text width=5em%, align = left
                ]
                % [
                %     {\citet{gao-etal-2023-precise},}
                %     {\citet{10.1145/3731120.3744614},}
                %     {\citet{zhao-etal-2024-pacar},}
                %     {\citet{li-etal-2025-imrrf},}
                %     {\citet{acosta2025ucom_unam_pln}}
                %     , leaf, text width=29.1em, align = left
                % ]
            ]
			]
		\end{forest}
}
\caption{Taxonomy of approaches for evidence retrieval. The diagram organizes studies by task type and shows the number of papers for each approach, further distinguished by language coverage (English-only, non-English, and multilingual).}
\label{fig:taxonomy-evidence-retrieval}
\end{figure*}

\paragraph{Definition.}

\textit{Evidence retrieval} refers to the process of identifying and ranking information from trusted sources that can support or refute a claim~\cite{ijcai2021p0619, guo-etal-2022-survey}. In the literature, this task is also referred to as \textit{document retrieval}~\cite{chakrabarty-etal-2018-robust}, \textit{passage retrieval}~\cite{kobayashi-etal-2017-automated}, \textit{evidence}~\cite{tan-etal-2025-improving} or \textit{rationale selection}~\cite{wang-etal-2023-check-covid}, with terminology reflecting differences in evidence granularity or retrieval formulation, such as ranking, filtering, or generation-based approaches. Evidence retrieval is a central component of automated fact-checking pipelines, providing the factual grounding necessary for subsequent veracity assessment. Depending on the claim and domain, evidence may consist of text passages~\cite{thorne-etal-2018-fever}, tables~\cite{lu-etal-2023-scitab}, or scientific data~\cite{wang-etal-2023-check-covid},  typically drawn from large knowledge bases such as Wikipedia, news archives, scientific literature, and the open web.

\paragraph{LLM Applications.}

Generative LLMs have introduced new paradigms for evidence retrieval, as illustrated in Figure~\ref{fig:taxonomy-evidence-retrieval}. Beyond serving as rankers~\cite{10.1145/3404835.3463120, pasin2024seupd}, LLMs are increasingly used for \textbf{query generation}, where claims are reformulated into effective search queries~\cite{prietochavana2023automated, abdallah2025ngu_research}, and for \textbf{rationale selection}, identifying the most relevant sentences from retrieved passages~\cite{pradeep2020scientific, kamoi-etal-2023-wice}. They are also applied to \textbf{question generation} and \textbf{summary generation}, guiding retrieval by decomposing to sub-questions~\cite{schlichtkrull2023averitec} or condensing evidence for downstream verification~\cite{zeng-etal-2024-ru22fact}. Overall, the taxonomy highlights a shift from retrieval as a ranking problem to a generative process, in which LLMs actively shape how evidence is searched, selected, and represented.

\paragraph{Datasets.}

\begin{table}[]
    \caption{Overview of commonly used evidence retrieval datasets, summarizing each dataset by output type, number of supported languages, dataset, and corpus size (in thousands).}
    \label{tab:evidence-retrieval-datasets}
    \small
    \centering
    \resizebox{\textwidth}{!}{%
    \begin{tabular}{llrrrl}
    \toprule
         \textbf{Name} & \textbf{Output Type} & \textbf{\# Langs} & \textbf{\# Claims [K]} & \textbf{\# Corpus [K]} & \textbf{Citation} \\
         \midrule
         AVeriTec & Question Generation & 1 & 4.6 & & \citet{schlichtkrull2023averitec} \\
         Check-COVID & Sentence selection & 1 & 1.5 & 0.4 & \citet{wang-etal-2023-check-covid} \\
         COVID-Fact & Sentence selection & 1 & 4 & - & \citet{saakyan-etal-2021-covid} \\
         FEVER & Document / Sentence selection & 1 & 185 & 5,417 & \citet{thorne-etal-2018-fever} \\
         FEVEROUS & Document / Sentence selection & 1 & 87 & 5,417 & \citet{aly-etal-2021-fact} \\
         HoVer & Document / Sentence selection & 1 & 26.1 & - & \citet{jiang-etal-2020-hover} \\
        MS-MARCO & Sentence selection & 1 & 1,000 & 8,800 & \citet{bajaj2018msmarcohumangenerated} \\
        MSVEC & Rationale selection & 1 & 0.2 & - & \citet{10.1145/3565287.3617630} \\
        SciFact & Document / Sentence selection & 1 & 1.4 & 5.2 & \citet{wadden-etal-2020-fact} \\
        TREC-COVID & Document selection & 1 & 0.05 & 191.2 & \citet{ROBERTS2021103865} \\
        TREC 2020 Health Misinfo. & Document selection & 1 & & & \citet{clarke2020overview} \\
         \bottomrule
    \end{tabular}
    }
\end{table}

Evidence retrieval datasets vary in scale, domain, and granularity of annotation, reflecting the diversity of sources from which factual evidence can be extracted. Table~\ref{tab:evidence-retrieval-datasets} summarizes commonly used datasets, highlighting their output types, language coverage, and the size of both claims and evidence corpora.

General-purpose benchmarks such as \textit{FEVER}~\cite{thorne-etal-2018-fever} remain central to evidence retrieval and claim verification. They link claims to supporting or refuting documents and sentences from Wikipedia, with \textit{FEVEROUS}~\cite{aly-etal-2021-fact} additionally incorporating tabular evidence. Their explicit supervision at multiple levels of granularity has made them standard benchmarks for evaluating evidence retrieval and selection, also for generative LLMs.

Several datasets adapt evidence retrieval to specialized domains. \textit{SciFact}~\cite{wadden-etal-2020-fact} focuses on scientific claim verification using biomedical abstracts, while \textit{Check-COVID}~\cite{wang-etal-2023-check-covid} and \textit{COVID-Fact}~\cite{saakyan-etal-2021-covid} target COVID-19-related misinformation. Although smaller in scale, these datasets provide high-quality annotations and highlight challenges for LLMs in domains with limited evidence sources.

Other datasets emphasize reasoning and retrieval formulation. \textit{HoVER}~\cite{jiang-etal-2020-hover} requires retrieval across multiple documents, where evidence may need to be combined from various sources. On the other hand, \textit{AVerTeC}~\cite{schlichtkrull2023averitec} adopts a question-driven approach to decompose claims into intermediate queries that guide evidence extraction. In addition, resources from information retrieval, such as \textit{MS MARCO} dataset~\cite{bajaj2018msmarcohumangenerated} and \textit{TREC-COVID}~\cite{ROBERTS2021103865}, are frequently reused to benchmark passage and document-level retrieval in fact-checking pipelines.

\paragraph{Limitations.}

Research on using generative LLMs for evidence retrieval remains relatively scarce, as most pipelines still rely on traditional retrieval systems or hybrid approaches~\cite{prietochavana2023automated, pasin2024seupd}. When applied, LLMs pose several limitations for evidence retrieval. They can hallucinate evidence, generating supporting text that does not exist in the source corpus, and are highly sensitive to prompt formulation, which can lead to inconsistent or incomplete retrieval~\cite{kamoi-etal-2023-wice, zeng-etal-2024-ru22fact}. LLMs also struggle with granularity and reasoning depth, since selecting the most relevant sentences or passages from long documents is challenging, and they may miss the context needed to accurately support or refute complex claims. Cross-lingual and open-domain retrieval is particularly difficult, as models must identify semantically equivalent evidence across languages, domains, and noisy web sources. Finally, current evaluation metrics do not fully capture the credibility, completeness, or interpretability of LLM-retrieved evidence, complicating assessment and deployment in real-world fact-checking workflows~\cite{tan-etal-2025-improving}.

\subsection{Fact Verification and Fake News Detection}

\paragraph{Definition.}

\textit{Fact verification} and \textit{fake news detection} are the most widely studied tasks in automated fact-checking using LLMs, serving as the foundation for veracity assessment~\cite{ijcai2021p0619}. Fact verification focuses on evaluating the truthfulness of short, explicit claims, typically by comparing them against retrieved evidence from trusted sources~\cite{zeng2021automatedfactcheckingsurvey}. On the other hand, fake news detection aims to assess the truthfulness of longer and more contextually rich texts, such as news articles, considering contextual, linguistic, and stylistic features that may indicate misinformation~\cite{10.1145/3485127}. Both tasks are also referred to as \textit{claim verification}~\cite{dmonte2025claimverificationagelarge}, \textit{misinformation detection}~\cite{10.1145/3711896.3737437}, \textit{truth assessment}~\cite{nakashole-mitchell-2014-language}, or \textit{rumor detection}~\cite{li-etal-2019-rumor}, depending on the application and granularity of analysis. 

\begin{figure*}
\centering
\resizebox{\textwidth}{!}{
\begin{forest}
			forked edges,
			for tree={
				grow=south,
                node options={align=center},
                align = center,
				base=left,
				font=\small,
				rectangle,
				draw=hidden-draw,
				rounded corners,
				minimum width=4em,
				edge+={darkgray, line width=1pt},
				s sep=3pt,
				inner xsep=2pt,
				inner ysep=3pt,
				ver/.style={rotate=90, child anchor=north, parent anchor=south, anchor=center},
			},
			where level=1{text width=5.0em,font=\scriptsize}{},
			where level=2{text width=5.6em,font=\scriptsize}{},
			where level=3{text width=6.8em,font=\scriptsize}{},
			[
			Fact Verification \&\\Fake News Detection, text width=10em
			[
			Classification, text width=5em
			[
			Binary\\Classification, text width=5em
                [
                    \textit{\# of Papers: 49}\\
                    \textit{English-only: 39}\\
                    \textit{Non-English: 1}\\
                    \textit{Multilingual: 9}
                    ,leaf-center, text width=5em
                ]
			]
            [
			3-Class\\Classification, text width=5em
                [
                    \textit{\# of Papers: 70}\\
                    \textit{English-only: 59}\\
                    \textit{Non-English: 4}\\
                    \textit{Multilingual: 7}
                    ,leaf-center, text width=5em
                ]
			]
            [
			4+\\Classification, text width=5em
                [
                    \textit{\# of Papers: 19}\\
                    \textit{English-only: 13}\\
                    \textit{Non-English: 0}\\
                    \textit{Multilingual: 6}
                    ,leaf-center, text width=5em
                ]
			]
			]
            [
                Regression, text width=5em
                [
                    \textit{\# of Papers: 9}\\
                    \textit{English-only: 6}\\
                    \textit{Non-English: 0}\\
                    \textit{Multilingual: 3}
                    ,leaf-center, text width=5em
                ]
            ]
            [
            Explanation\\Generation, text width=5em
                [
                    \textit{\# of Papers: 46}\\
                    \textit{English-only: 40}\\
                    \textit{Non-English: 1}\\
                    \textit{Multilingual: 5}
                    ,leaf-center, text width=5em
                ]
                [
                Fact-Checking\\Article\\Generation, text width=5em
                [
                    \textit{\# of Papers: 1}\\
                    \textit{English-only: 1}\\
                    \textit{Non-English: 0}\\
                    \textit{Multilingual: 0}
                    ,leaf-center, text width=5em
                ]
            ]
            ]
            [
            Others, text width=5em
                [
                    \textit{\# of Papers: 3}\\
                    \textit{English-only: 2}\\ 
                    \textit{Non-English: 0}\\
                    \textit{Multilingual: 1}
                    ,leaf-center, text width=5em
                ]
            ]
			]
		\end{forest}
}
\caption{Taxonomy of approaches to fact verification and fake news detection. The diagram groups studies by task and indicates the number of papers for each approach, with a breakdown by language coverage.}
\label{fig:taxonomy-fact-verification}
\end{figure*}

\paragraph{LLM Applications.}

In fact verification and fake news detection, LLMs are commonly employed for classification, with the goal of classifying claims or full articles into multiple classes, with labels ranging from binary (true/false) to ternary (supports/refutes/not enough information) or more fine-grained categories depending on the dataset~\cite{guo-etal-2022-survey}. Beyond classification, generative LLMs are increasingly employed to produce \textbf{natural language explanations} that justify the veracity prediction by highlighting supporting or contradicting evidence, referencing retrieved sources, or outlining reasoning steps~\cite{tan2023evidencebased, ma2023exfever}. Incorporating explanation generation enhances interpretability, facilitates human oversight, and helps fact-checkers understand and trust automated predictions, though faithfulness remains a challenge, as LLMs may generate plausible-sounding rationales that do not accurately reflect the evidence or reasoning~\cite{10.1007/978-3-031-47896-3_1}. All commonly used approaches for fact verification and fake-news detection are summarized in Figure~\ref{fig:taxonomy-fact-verification}.

A recent extension of explanation generation is the \textbf{generation of fact-checking articles}~\cite{sahnan2025llmsautomatefactcheckingarticle}. Instead of a label or brief justification, LLMs produce structured, coherent reports that explain why the claim is being verified, include the claim context, provide evidence-based reasoning, and offer a final verdict understandable to the public. This approach enables LLMs to synthesize information across sources and generate fluent, informative articles, but also raises challenges in factual accuracy, hallucination mitigation, and consistency with retrieved evidence.

\paragraph{Datasets.}

\begin{table}[]
    \caption{Overview of fact verification and fake news detection datasets, including label types, language coverage, dataset size, availability of gold evidence, and the type of evidence.}
    \label{tab:datasets-claim-verification}
    \tiny
    \centering
    \resizebox{\textwidth}{!}{%
    \begin{tabular}{llrrcll}
    \toprule
         \makecell[c]{\textbf{Name}} & \makecell[c]{\textbf{Output Type}} & \makecell[c]{\textbf{\# Langs}} & \makecell[c]{\textbf{\# Size [K]}} & \makecell[c]{\textbf{Golden Evidence}} & \makecell[c]{\textbf{Evidence Type}} & \makecell[c]{\textbf{Citation}} \\
         \midrule
         ANTiVax & 2 classes & 1 & 15 & & Twitter & \citet{HAYAWI202223} \\
         ArCOV19-Rumors & 3 classes & 1 & 0.1 & \checkmark & FC websites + Research Articles & \citet{haouari-etal-2021-arcov19} \\
         ArFactEx & 2 classes / Explanation & 1 & 0.1 & & News & \citet{althabiti2024takeed} \\
         AVeriTec & 4 classes & 1 & 4.6 &  & - & \citet{schlichtkrull2023averitec} \\
         BingCheck & 2 classes & 1 & 0.4 & \checkmark & LLMs & \citet{li-etal-2024-self} \\
         BoolQ-FV / AmbiFC & 3 classes & 1 & 20.2 & \checkmark & Wikipedia & \citet{glockner-etal-2024-ambifc} \\
         Celebrity & 2 classes & 1 & 0.5 & & News articles &\citet{perez-rosas-etal-2018-automatic}\\
         ChatGPT-FC & 6 classes / Score (0-100) & 1 & 22.3 & & PolitiFact & \citet{li2023revisitfakenewsdataset} \\
         Check-COVID & 3 classes & 1 & 1.5 & \checkmark & Journal Articles (CORD-19) & \citet{wang-etal-2023-check-covid} \\
         CHECKWHY & 3 classes / Argument generation & 1 & & \checkmark & Wikipedia & \citet{si-etal-2024-checkwhy} \\
         CHEF & 3 classes & 1 & 10 & \checkmark & Google Search & \citet{hu-etal-2022-chef} \\
         ClaimDiff & 2 classe / Rationale selection & 1 & 2.9 & & News articles & \citet{ko-etal-2023-claimdiff} \\
         CLEF 2024 & 3 classes & 2 & 34 & \checkmark & Social media & \citet{10.1007/978-3-031-71908-0_2}\\
         CLEF 2025 (Numerical claims) & 3 classes & 3 & 23.6 & \checkmark & Web search & \citet{SettyOverviewOT} \\
         CLIMATE-FEVER & 3 classes & 1 & 1.5 & \checkmark & Wikipedia & \citet{diggelmann2021climatefeverdatasetverificationrealworld} \\
         CoAID & 2 classes & 1 & 5.2 & & Research Articles & \citet{cui2020coaidcovid19healthcaremisinformation} \\
         COCO & 2 classes & 1 & 2.6 & & Social media & \citet{langguth2023coco} \\
         Constraint & 2 classes & 1 & 11 & & FC websites + Social media & \citet{10.1007/978-3-030-73696-5_3} \\ 
         CorFEVER & 3 / 7 classes & 1 & 2 & \checkmark & Wikipedia + Web search & \citet{tan-etal-2025-improving} \\
         CorXFact & 3 classes / Explanation & 1 & & \checkmark & & \citet{tan-etal-2025-improving} \\
         CoVERT & 3 classes & 1 & 0.7 & \checkmark & Research articles & \citet{mohr-etal-2022-covert} \\
         COVID-Fact & 3 classes & 1 & 4 & \checkmark & Google Search & \citet{saakyan-etal-2021-covid} \\
         COVID-19 Scientific & 2 classes & 1 & 0.1 & \checkmark & Research articles & \citet{lee2020misinformationhighperplexity} \\
         CT-FAN-22 & 4 classes & 2 & 1.9 & & FC websites & \citet{kohler2022overview} \\
         DanFEVER & 3 classes & 1 & 6.4 & \checkmark & Wikipedia & \citet{norregaard-derczynski-2021-danfever} \\
         Data Common & - & X & N/A & N/A & N/A & - \\
         Defalsif-AI & - & 2+ & 194.3 & & News articles & \footnote{\url{https://science.apa.at/project/defalsifai/}} \\
         Demagog & - & 3 & - & - & - & \footnote{\url{https://demagog.cz/}} \\
         ECTF & 2 classes & 1 & 72.6 & & Social media & \citet{bansal2021combiningexogenousendogenoussignals} \\
         e-FEVER & 3 classes & 1 & 67.7 & \checkmark & Wikipedia & \citet{stammbach2020fever} \\
         EUDisinfo & 2 classes & 1 & 18.5 & & FC websites & \citet{modzelewski-etal-2025-pcot} \\
         ExClaim & 2 classes & 1 & 4 & \checkmark & FC websites & \citet{Gurrapu_2022} \\
         ExClaimCheck & & 1 & & & & \citet{} \\
         EX-FEVER & 3 classes / Explanation & 1 & 61.3 & \checkmark & Wikipedia & \citet{ma2023exfever} \\
         FactBench & 2 classes & 3 & 2.8 & & DBpedia + Freebase & \citet{GERBER201585} \\
         Factcheck-Bench & & 1 & 0.6 & \checkmark & LLMs & \citet{wang-etal-2024-factcheck} \\
         FactEX & Explanation & 1 & 12.2 & & FC websites & \citet{10.1007/978-3-031-47896-3_1} \\
         FactKG & 2 classes & 1 & 108.7 & \checkmark & DBpedia &  \citet{kim-etal-2023-factkg} \\
         FacTool & 2 classes & 1 & 0.2 & & Wikipedia + LLMs & \citet{chern2023factoolfactualitydetectiongenerative} \\
         Fake News Dataset German & 2 classes & 1 & 63 & & News articles & \footnote{\url{https://www.kaggle.com/datasets/astoeckl/fake-news-dataset-german}}\\
         FakeNewsAMT & 2 classes & 1 & 0.6 & & News articles & \citet{perez-rosas-etal-2018-automatic} \\
         FakeNewsNet & 2 classes & 1 & 22.6 & & FC websites & \citet{shu2019fakenewsnetdatarepositorynews} \\
         FCTR & 8 classes & 1 & 6.8 & \checkmark & FC websites & \citet{cekinel-etal-2024-cross} \\
         FELM & 2 classes & 1 & 0.2 & \checkmark & Quora + LLM & \citet{chen2023felm} \\
         FEVER & 3 classes & 1 & 185 & \checkmark & Wikipedia & \citet{thorne-etal-2018-fever} \\
         FEVER-IT & 3 classes & 1 & 246.3 & \checkmark & Wikipedia & \citet{scaiella-etal-2024-leveraging} \\
         FEVEROUS & 3 classes & 1 & 87 & \checkmark & Wikipedia & \citet{aly-etal-2021-fact} \\
         FIN-FACT & 3 classes & 1 & 3.4 & & FC websites & \citet{10.1145/3701716.3715292} \\
         FinDVer & 2 classes / Explanation & 1 & 2.1 & \checkmark & & \citet{zhao-etal-2024-findver} \\
         FM2 & 3 classes & 1 & 13 & \checkmark & Wikipedia & \citet{eisenschlos-etal-2021-fool} \\
         FullFact & - & 1 & - & - & - & - \\
         GettingReal & & 1 & 13 & & News articles & \citet{} \\
         Global-LIAR & 2 classes & 1 & 0.6 & & FC websites & \citet{mirza2024globalliar} \\
         GossipCop & - & 1 & - & - & - & - \\
         HealthFC & 3 classes & 1 & 0.3 & \checkmark & Medicine journal & \citet{vladika-etal-2024-healthfc} \\
         HealthVer & 3 classes & 1 & 14.3 & \checkmark & Journal Articles (CORD-19) & \citet{sarrouti-etal-2021-evidence-based} \\
         HoVer & 3 classes & 1 & 26.1 & \checkmark & Wikipedia & \citet{jiang-etal-2020-hover} \\
         ISOT Fake News & 2 classes & 1 & 44 & & FC websites & \citet{Ahmed2018DetectingOS} \\
         LIAR & 6 classes / Explanation & 1 & 12.8 & & PolitiFact & \citet{wang-2017-liar} \\
         LIAR++ & 3 classes / Explanation & 1 & 6.5 &  & FC websites & \citet{russo-etal-2023-benchmarking} \\
         LIAR-NEW & 6 classes & 2 & 2 & & PolitiFact & \citet{pelrine2023towards} \\
         MisinfoCorrect & 2 classes & 1 & 0.8 & \checkmark & Social media & \citet{10.1145/3543507.3583388} \\
         MMFC & 2 classes & 1 & 0.5 & \checkmark & Wikipedia & \citet{bussotti-etal-2024-unknown} \\
         MSVEC & 2 classes & 1 & 0.2 & \checkmark & FC websites + Research Articles & \citet{10.1145/3565287.3617630} \\
         MultiDis & 2 classes & 1 & 2 & & Articles & \citet{modzelewski-etal-2025-pcot} \\
         MultiSynFact & 3 classes & 3 & 2200 & & Wikipedia + LLMs & \citet{chung2025translationllmbaseddatageneration} \\
         NewsPolyML & 4 classes & 5 & 32.5 & \checkmark & FC websites & \citet{10.1145/3643491.3660290} \\
         PHEME & 2 classes & 1 & 0.3 & & News articles & \citet{8118443} \\
         PHEMEPlus & 3 classes & 1 & 2 & \checkmark & News articles + Web search & \citet{dougrez-lewis-etal-2022-phemeplus} \\
         PolitiFact & 6 classes & 1 & N/A & N/A & FC websites & - \\
         PolyNarrative & Narrative classification & 4 & 1.5 & & News articles & \citet{nikolaidis-etal-2025-polynarrative} \\
         PubHealth & 4 classes / Explanation & 1 & 11.8 & \checkmark & FC websites + News & \citet{kotonya-toni-2020-explainable-automated} \\
         PubMedQA & 3 classes / Explanation & 1 & 273.5 & & PubMed & \citet{jin-etal-2019-pubmedqa} \\
         RAWFC & 3 classes / Explanation & 1 & 2 & & PolitiFact & \citet{yang-etal-2022-coarse} \\
         RU22Fact & 3 classes / Explanation & 4 & 16 & \checkmark & FC websites + LLMs & \citet{zeng-etal-2024-ru22fact} \\
         SciFact & 3 classes & 1 & 1.4 & \checkmark & Journal Articles (CORD-19) & \citet{wadden-etal-2020-fact} \\
         SciTab & 3 classes & 1 & 1.2 & \checkmark & arXiv & \citet{lu-etal-2023-scitab} \\
         Scitance & 3 classes & 1 & 0.7 & \checkmark & Journal Articles (CORD-19) & \citet{alvarez-etal-2024-zero} \\
         SEMTAB-FACTS & 3 classes & 1 & 180 & \checkmark & Research articles & \citet{wang-etal-2021-semeval} \\
         Snopes & - & 1 & N/A & N/A & - & - \\
         Symmetric & 2 classes & 1 & 1 & \checkmark & FEVER & \citet{schuster-etal-2019-towards} \\
         TabFact & 2 classes & 1 & 18 & & Wikipedia & \citet{chen2020tabfactlargescaledatasettablebased} \\
         % ToTTo & & 1 & 120 & & Wikipedia & \citet{parikh-etal-2020-totto} \\
         Unified-FC & 2 classes & 1 & 3 & & \citet{baly-etal-2018-integrating} \\
         VitaminC & 3 classes & 1 & 488.9 & \checkmark & Wikipedia & \citet{schuster-etal-2021-get} \\
         WatClaimCheck & 3 classes & 1 & 33.7 & \checkmark & FC websites & \citet{khan-etal-2022-watclaimcheck} \\
         Weibo21 & 2 classes & 1 & 9.1 &  & Weibo News & \citet{10.1145/3459637.3482139} \\
         WiCE & 3 classes & 1 & 7.3 & \checkmark & Wikipedia & \citet{kamoi-etal-2023-wice} \\
         X-Fact & 7 classes & 25 & 31 & \checkmark & FC websites & \citet{gupta-srikumar-2021-x} \\
         XClaimCheck & 5 classes & 1 & 16.2 & & FC websites & \citet{kao-yen-2024-magic} \\
         X Community Notes & - & X & - & - & - \\
         % YAGO & 2 classes & 1 & 1.4 & & & \citet{ojha-talukdar-2017-kgeval} \\
         XSumFaith & 2 classes & 1 & 0.5 & \checkmark & XSum~\cite{narayan-etal-2018-dont} /BBC & \citet{maynez-etal-2020-faithfulness} \\
         \bottomrule
    \end{tabular}
    }
\end{table}

Datasets for fact verification and fake news detection differ substantially in label granularity, evidence availability, language coverage, and annotation structure, directly shaping how generative LLMs are applied. Table~\ref{tab:datasets-claim-verification} summarizes the most commonly used resources.

Many datasets for fact verification frame the task as multi-class classification with explicit evidence linking claims to sentences or documents. Notable examples include \textit{FEVER}~\cite{thorne-etal-2018-fever}, \textit{FEVEROUS}~\cite{aly-etal-2021-fact}, and \textit{HoVer}~\cite{jiang-etal-2020-hover}. The availability of gold evidence encourages evidence-conditioned LLM approaches, in which models jointly reason over retrieved context to predict veracity, often producing natural-language explanations. These datasets facilitate transparent reasoning but also expose limitations when LLMs generate explanations that are not fully grounded in evidence.

In contrast, many fake news detection datasets, such as \textit{FakeNewsNet}~\cite{shu2019fakenewsnetdatarepositorynews}, \textit{ISOT Fake News}~\cite{Ahmed2018DetectingOS}, and \textit{FakeNewsAMT}~\cite{perez-rosas-etal-2018-automatic}, provide only binary labels without supporting evidence. LLMs in these settings are typically applied as text classifiers or zero-shot predictors. While explanations can be generated, their faithfulness is difficult to evaluate, highlighting a key challenge for practical deployment.

Some datasets explicitly include rationales or argumentative structures to evaluate LLMs. Resources such as \textit{LIAR}~\cite{wang-2017-liar}, \textit{PubHealth}~\cite{kotonya-toni-2020-explainable-automated}, or \textit{EX-FEVER}~\cite{ma2023exfever} allow joint verification and explanation generation, promoting research on explanation quality. Domain-specific datasets, particularly in scientific and health contexts, e.g., \textit{SciFact}~\cite{wadden-etal-2020-fact}, \textit{HealthVer}~\cite{sarrouti-etal-2021-evidence-based}, and \textit{Check-COVID}~\cite{wang-etal-2023-check-covid}, require precise reasoning over exper-annotated evidence and are often paired with retrieval-augmented or prompt-based LLM approaches.

Recent trends emphasize multilingual benchmarks and synthetic data. \textit{X-Fact} dataset~\cite{gupta-srikumar-2021-x}, and \textit{NewsPolyML}~\cite{10.1145/3643491.3660290} enable cross-lingual evaluation but highlight evidence alignment challenges, while synthetic datasets such as  \textit{MultiSynFact}~\cite{chung2025translationllmbaseddatageneration} increase scale at the risk of distributional mismatch.

% Overall, the structure of datasets plays a critical role in determining whether LLMs are used as black-box classifiers, evidence-aware reasoners, or explanation generators. Datasets with explicit evidence and rationale annotations support more transparent and controllable use of generative models, which is crucial for real-world fact-checking applications. In contrast, datasets lacking evidence supervision risk encouraging overreliance on parametric knowledge and unverifiable explanations. As LLM-based fact-checking systems evolve, there is a growing need for datasets that combine multilingual coverage, high-quality evidence annotations, and evaluation protocols that explicitly assess explanation faithfulness rather than label accuracy alone.

\paragraph{Limitations.}

LLM-based fact verification and fake news detection pose several limitations despite their wide adoption. They can hallucinate or produce plausible-sounding explanations that are not grounded in retrieved evidence, reducing reliability~\cite{ma2023exfever, sahnan2025llmsautomatefactcheckingarticle}. LLMs are sensitive to surface-level cues and may fail to account for nuanced or context-dependent information, particularly in longer texts such as news articles. Explanation and article generation improve interpretability but increase the risk of inconsistency or factual errors, while automatic metrics often fail to capture reasoning validity or evidence alignment, necessitating labor-intensive human evaluation~\cite{thakur-etal-2025-judging}. Cross-lingual and domain-specific claims further challenge LLMs, as models must generalize beyond the distribution seen during pretraining.

\section{Use Cases for Generative LLMs in Fact-Checking}
\label{sec:methods-taxonomy}

\begin{figure}
    \centering
    \includegraphics[width=\linewidth]{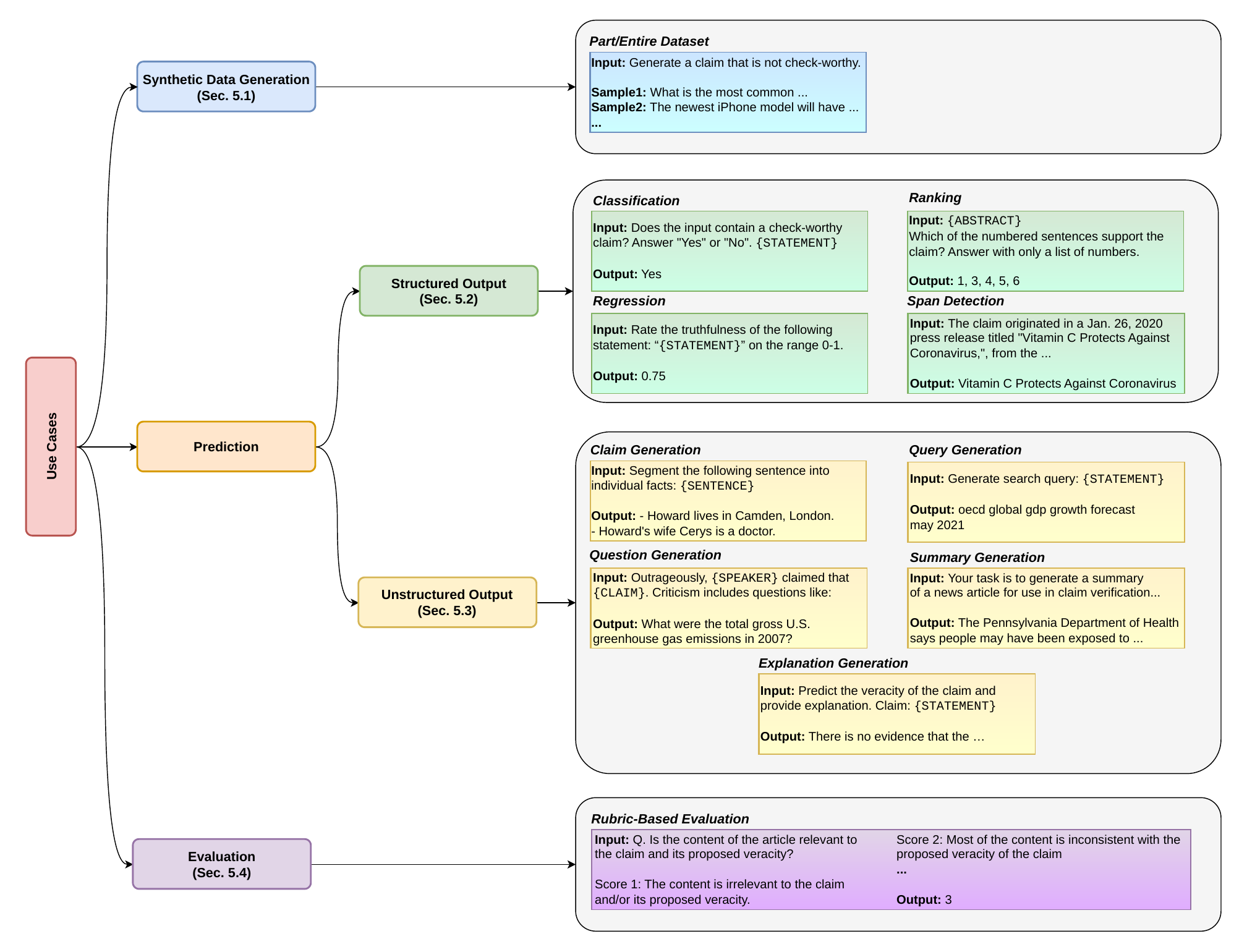}
    \caption{Taxonomy of approaches for integrating generative LLMs in fact-checking, organized by use case: synthetic data generation, prediction, and evaluation. The figure illustrates representative model inputs and outputs for each category, highlighting how LLMs support different stages of the fact-checking process.}
    \label{fig:taxonomy}
\end{figure}

In this section, we define a taxonomy of approaches that leverage generative LLMs for fact-checking tasks. With the emergence of LLMs, traditional fact-checking pipelines have been replaced by more flexible generative approaches. We organize the literature by the LLM's role and purpose in the fact-checking process. This use-case-oriented taxonomy captures how generative models are used to support different stages of automated fact-checking. The taxonomy is illustrated in Figure~\ref{fig:taxonomy} and consists of three main use cases:

\begin{enumerate}
    \item \textbf{Synthetic Data Generation}, where LLMs are employed to augment existing datasets or create new datasets to improve model robustness.
    \item \textbf{Prediction}, where LLMs directly perform or assist downstream tasks such as claim detection, previously fact-checked claim retrieval, evidence retrieval, or fact verification. Within this category, we further differentiate between \textit{structured prediction}, where the LLM produces outputs constrained to predefined labels or scores within formal answer spaces (e.g., veracity classes or detected spans), and \textit{unstructured prediction}, where the model generates free-form, natural language outputs.
    \item \textbf{Evaluation}, where LLMs serve as evaluators, assessing factuality, consistency, or overall quality of fact-checking systems and outputs, often framed as an LLM-as-a-judge paradigm.
\end{enumerate}

This taxonomy highlights the integration of LLMs in fact-checking systems, spanning from data augmentation to task-oriented inference and system-level evaluation. By focusing on the role of generative models, we aim to provide a clear understanding of how recent advances in LLMs are reshaping both the methodology and scope of automated fact-checking research.

\subsection{Synthetic Data Generation}
\label{sec:synthetic}

\begin{figure*}
\centering
\resizebox{\textwidth}{!}{
\begin{forest}
			forked edges,
			for tree={
				grow=east,
				reversed=true,
				anchor=base west,
				parent anchor=east,
				child anchor=west,
                node options={align=center},
                align = center,
				base=left,
				font=\small,
				rectangle,
				draw=hidden-draw,
				rounded corners,
				minimum width=4em,
				edge+={darkgray, line width=1pt},
				s sep=3pt,
				inner xsep=2pt,
				inner ysep=3pt,
				ver/.style={rotate=90, child anchor=north, parent anchor=south, anchor=center},
			},
			where level=1{text width=5.0em,font=\scriptsize}{},
			where level=2{text width=5.6em,font=\scriptsize}{},
			where level=3{text width=6.8em,font=\scriptsize}{},
			[
			Synthetic Data\\Generation, ver
            [
                Dataset Augmentation, text width=7.5em
                [
                    {\citet{aarnes2024iaigroupcheckthat2024},}
                    {\citet{bussotti-etal-2024-unknown},} % Need to check
                    {\citet{chen-zhang-2025-retrieverguard},}  % Need to check
                    {\citet{10.1145/3701716.3715521},} % Need to check
                    {\citet{gruman2024clac},}\\
                    {\citet{kanaani-etal-2024-triple},} % Need to check
                    {\citet{10.1145/3589335.3651521},}
                    {\citet{kim-etal-2023-covid},}
                    {\citet{lu-etal-2023-scitab},}
                    {\citet{modzelewski2023dshacker},}\\
                    {\citet{10.1145/3733567.3735571},} % Need to check
                    {\citet{10.1145/3696410.3714569},} % Need to check
                    {\citet{saeed2025mma},}
                    {\citet{10.1145/3626772.3657667},}
                    {\citet{tan-etal-2025-improving},} % Need to check
                    {\citet{wu2023fake},}\\
                    {\citet{zhang2024need},}
                    {\citet{zhu-etal-2023-explain}}
        			, leaf, align = left, text width=34.4em
                ]
			]
            [
                Synthetic Datasets, text width=7.5em
                [
                    {\citet{alvarez-etal-2024-zero},} % Need to check
                    {\citet{chen2023llmgenerated},}
                    {\citet{choi2023automated},}
                    {\citet{choi2024factgpt},}
                    {\citet{chung2025translationllmbaseddatageneration},} % Need to check
                    {\citet{10.1145/3654777.3676359},}\\
                    {\citet{10.1145/3726302.3730027},}
                    {\citet{huang2024fakegptfakenewsgeneration},}
                    {\citet{jiang2023disinformation},}
                    {\citet{li2023revisitfakenewsdataset},} % Need to check
                    {\citet{si-etal-2024-checkwhy},} % Need to check
                    {\citet{singh2023utdrm},}\\% Need to check
                    {\citet{10.1145/3648188.3675153}} % Need to check
        			,leaf, align = left, text width=34.4em
                ]
			]
			]
		\end{forest}
}
\caption{Taxonomy of synthetic data generation approaches in fact-checking. The diagram distinguishes between methods that augment existing datasets and approaches that create entirely synthetic datasets, listing representative studies for each category.}
\label{fig:synthetic-data-taxonomy}
\end{figure*}

Synthetic data generation has become increasingly used within automated fact-checking research, as manually annotated fact-checking datasets are difficult to obtain. The LLMs can be employed for both dataset augmentation~\cite{modzelewski2023dshacker, wu2023fake} and the creation of entirely synthetic corpora~\cite{10.1145/3654777.3676359,chen2023llmgenerated, huang2024fakegptfakenewsgeneration}, as illustrated in the taxonomy of synthetic data generation approaches in Figure~\ref{fig:synthetic-data-taxonomy}. The primary motivation is the scarcity, imbalance, and domain-dependence of annotated fact-checking datasets, particularly in multilingual contexts. LLMs offer a flexible mechanism for producing large volumes of additional data, as they can rewrite, translate, change style, or paraphrase existing samples, or generate new claims, posts, or misinformation narratives from scratch. Across the literature, these approaches aim to reduce annotation costs, mitigate class imbalance, diversify linguistic and stylistic distributions, and allow researchers to create controlled, task-aligned supervision signals that would be impractical to obtain manually.

A key benefit of synthetic data is the ability to systematically explore variation. Methods based on rewriting modify the wording of existing claims while preserving their meaning. This allows controlled variation in surface form and helps models become robust to paraphrases, cross-lingual noise, and stylistic differences~\cite{modzelewski2023dshacker,wu2023fake}. Style transfer and culturally contextualized rewriting make it possible to simulate low-resource languages or platforms with distinctive discourse patterns, reducing the need for manual annotation~\cite{aarnes2024iaigroupcheckthat2024}. In contrast, generative methods that create new claims or misinformation narratives allow researchers to build entire synthetic benchmarks, control the distribution of generated examples, and explicitly model different factual relationships such as entailment, contradiction, and neutrality~\cite{choi2023automated,choi2024factgpt}. Template-driven pipelines and structured prompts further support the generation of datasets with explicit metadata, relationship labels, or predefined factual constraints, enabling more systematic evaluation and controlled experimentation~\cite{10.1145/3654777.3676359,10.1145/3626772.3657667}. The flexibility of LLMs also enables the generation of sophisticated misinformation, including hallucinated articles, fabricated evidence, and stylistically tailored manipulations that resemble realistic disinformation patterns~\cite{jiang2023disinformation,chen2023llmgenerated,huang2024fakegptfakenewsgeneration}.

However, synthetic data also introduces important limitations. A frequently raised issue is distributional mismatch, where LLM-generated samples may be too clean, coherent, or stylistically homogeneous than real-world misinformation, increasing the risk of overfitting and reducing generalization. Moreover, augmenting the dataset for the imbalanced class or creating a separate class containing only generated samples can cause downstream models to learn artifacts specific to synthetic data rather than task-relevant signals. For example, in check-worthy claim detection, models trained on generated, not check-worthy claims, may learn to distinguish synthetic from human-written text rather than identifying claim check-worthiness. Highly controlled generation pipelines can inadvertently encode biases present in the LLM (e.g., ideological framing or linguistic bias). Safety guardrails often prevent LLMs from generating harmful misinformation, which may require extensive prompt engineering or other techniques (e.g., jail-breaking) to bypass content filters~\cite{chen2023llmgenerated,huang2024fakegptfakenewsgeneration}. Even when successful, these approaches raise concerns about unintentionally amplifying harmful content if such datasets are not handled responsibly. Finally, the quality of synthetic annotations, such as labels, relations, or structured metadata, can vary significantly across LLMs, requiring post-processing, clustering, and human verification in many pipelines~\cite{10.1145/3626772.3657667,lu-etal-2023-scitab}.

Despite these limitations, synthetic data generation plays a crucial role in fact-checking. The approaches described below illustrate the two major directions: \textit{augmenting existing datasets} and \textit{creating entirely new synthetic corpora}.

\paragraph{Augmenting Datasets with Synthetic Data.}

Augmentation-based synthetic generation enriches existing datasets by expanding their linguistic, stylistic, or semantic variability while retaining the underlying factual content. The most common techniques involve paraphrasing~\cite{modzelewski2023dshacker, gruman2024clac}, translation~\cite{modzelewski2023dshacker, aarnes2024iaigroupcheckthat2024} and stylistic reframing~\cite{wu2023fake, aarnes2024iaigroupcheckthat2024, 10.1145/3696410.3714569}. Paraphrasing uses concise prompts to produce lexically diverse but semantically equivalent versions of claims or social-media posts, thereby increasing robustness to reworded misinformation and reducing the impact of spurious lexical cues~\cite{modzelewski2023dshacker,gruman2024clac}. Translation-based augmentation extends this approach cross-lingually, allowing multilingual models to benefit from aligned variants in multiple languages~\cite{modzelewski2023dshacker}, often with careful prompt design to avoid hallucinated expansions. However, translation-based approaches can produce claims that are culturally inappropriate or inconsistent with the norms of the target language and region. Furthermore, style-transfer generation adapts samples to specific styles, such as those associated with social media platforms. Other examples also include transforming canonical English statements into Arabic social-media formats, which helps datasets better reflect the linguistic and stylistic patterns observed in real online posts~\cite{aarnes2024iaigroupcheckthat2024}.  

Another category of techniques employed LLMs to expand datasets in task-specific ways. Counter-factual generation creates minimally edited claims that invert meaning and veracity, providing harder negative examples and enabling more balanced training~\cite{lu-etal-2023-scitab,zhang2024need,zhu-etal-2023-explain}. Category-conditioned or key-point-guided generation produces new claims by following specific misinformation topics or the key elements of real claims. This creates diverse, structured examples that help train models when human-annotated data is limited~\cite{kim-etal-2023-covid,saeed2025mma}. Some approaches used LLMs to extract structured information from real texts, including claim components, factual flaws, or relational labels. By converting each input into multiple enriched annotations, these methods generate synthetic training signals that can guide and improve downstream fact-checking models~\cite{10.1145/3589335.3651521,10.1145/3626772.3657667}.

Across these methods, synthetic augmentation is typically combined with filtering, duplicate removal, clustering, or manual inspection to ensure that the new samples remain consistent with the original datasets and do not create out-of-distribution data~\cite{modzelewski2023dshacker, 10.1145/3733567.3735571}. The result is an expanded training set that is more balanced, more diverse, and often more aligned with the linguistic patterns encountered in real-world misinformation.

\paragraph{Creating New Synthetic Datasets.}

The second major direction involves generating entire datasets, either from scratch or using the existing data. These synthetic datasets are often designed to target specific fact-checking tasks, create controlled benchmarks, or provide training data for systems where human annotations are unavailable. In practice, such pipelines typically start from seed information, such as verified facts, existing fact-check articles, truth-value labels, or high-level templates. LLMs then generate new claims, social media posts, or news-style narratives that vary in factuality, stance, or semantic structure~\cite{choi2023automated,choi2024factgpt}.

Several approaches emphasize structured control over generation. Template-driven methods specify general fact types or relational structures and use LLMs to produce corresponding examples in natural language, often pairing each generated instance with explicit metadata such as categories or year~\cite{10.1145/3654777.3676359}. Other pipelines exploit LLMs' generative abilities to produce realistic but false or misleading content through controlled hallucinations, selective modifications, or meaning-altering rewrites~\cite{chen2023llmgenerated}. More complex multi-step strategies first extract entities, events, or claims from real sources and then selectively modify, recombine, or expand them into synthetic narratives that blend factual and fabricated elements~\cite{jiang2023disinformation}. Prompt-chaining techniques further allow models to iteratively expand simple outlines into detailed articles, deliberately modifying semantics or inventing fictional scenarios to generate high-quality synthetic misinformation~\cite{huang2024fakegptfakenewsgeneration}

Across these approaches, the generation of synthetic datasets provides substantial flexibility. The resulting data can be tailored to specific domains, linguistic phenomena, or reasoning challenges, and can be scaled far beyond what is achievable through manual annotation. At the same time, these methods require careful prompt design, validation, and ethical safeguards, as generating plausible misinformation, even for research purposes, introduces risks related to misuse, data leakage, and unintended model reinforcement. 

\subsection{Methods with Structured Output for Prediction}
\label{sec:structured}

\begin{figure*}
\centering
\resizebox{\textwidth}{!}{
\begin{forest}
			forked edges,
			for tree={
				grow=east,
				reversed=true,
				anchor=base west,
				parent anchor=east,
				child anchor=west,
                node options={align=center},
                align = center,
				base=left,
				font=\small,
				rectangle,
				draw=hidden-draw,
				rounded corners,
				minimum width=4em,
				edge+={darkgray, line width=1pt},
				s sep=3pt,
				inner xsep=2pt,
				inner ysep=3pt,
				ver/.style={rotate=90, child anchor=north, parent anchor=south, anchor=center},
			},
			where level=1{text width=5.0em,font=\scriptsize}{},
			where level=2{text width=5.6em,font=\scriptsize}{},
			where level=3{text width=6.8em,font=\scriptsize}{},
			[
			Methods with Structured Outputs, ver
			[
                Classification, text width=7.5em
                [
                    Binary\\Classification
                    [
                        {\citet{aarnes2024iaigroupcheckthat2024},}
                        {\citet{agrestia2022polimi},}
                        {\citet{bhatia2021automaticclaimreviewclimate},}
                        {\citet{bussotti-etal-2024-unknown},}
                        {\citet{cao2023large},}\\
                        {\citet{cekinel-etal-2024-cross},}
                        {\citet{chakraborty2023empirical},}
                        {\citet{chen2023llmgenerated},}
                        {\citet{dougrez-lewis-etal-2025-assessing},}\\
                        {\citet{du2022nus},}
                        {\citet{10.1145/3654777.3676359},}
                        {\citet{10.1145/3748514},}
                        {\citet{golik2024dshacker},}
                        {\citet{10.1145/3543507.3583870},}\\
                        {\citet{gruman2024clac},}
                        {\citet{10.1145/3583780.3614936},}
                        {\citet{hoes2023leveraging},}
                        {\citet{hoes2023using},}\\
                        {\citet{huang2024fakegptfakenewsgeneration},}
                        {\citet{hyben2023bigger},}
                        {\citet{jiang2023disinformation},}
                        {\citet{kamoi-etal-2023-wice},}\\
                        {\citet{kim-etal-2023-covid},}
                        {\citet{kim-etal-2023-kg},}
                        {\citet{10.1145/3701551.3703524},}
                        {\citet{kim2024llmsproducefaithfulexplanations},}
                        {\citet{ko-etal-2023-claimdiff},}\\
                        {\citet{leippold2024automated},}
                        {\citet{li-etal-2025-imrrf},}
                        {\citet{liu-etal-2024-teller},}
                        {\citet{liu2024large},}
                        {\citet{10.1145/3726302.3730092},}\\
                        {\citet{liu-etal-2025-raemollm},}
                        {\citet{lu-etal-2023-scitab},}
                        {\citet{mirza2024globalliar},}
                        {\citet{modzelewski-etal-2025-pcot},}
                        {\citet{10.1145/3733567.3735571},}\\
                        {\citet{10.1145/3627673.3679519},}
                        {\citet{ni-etal-2024-afacta},}
                        {\citet{pan2023qacheck},}
                        {\citet{pan-etal-2023-fact},}
                        {\citet{10.1145/3696410.3714569},}\\
                        {\citet{pelrine2023towards},}
                        {\citet{pham-etal-2025-claimpkg},}
                        {\citet{pisarevskaya-zubiaga-2025-zero},}
                        {\citet{10.1145/3699682.3728349},}\\
                        {\citet{russo-etal-2025-euroverdict},}
                        {\citet{sanyal2024minds},}
                        {\citet{sawinski2023openfact},}
                        {\citet{10.1145/3627673.3679985},}
                        {\citet{setty2024endtoend},}\\
                        {\citet{10.1145/3726302.3730142},}
                        {\citet{si-etal-2024-large},}
                        {\citet{si-etal-2024-checkwhy},}
                        {\citet{10.1145/3705754.3705757},}
                        {\citet{vergho2024comparing},}\\
                        {\citet{vykopal-etal-2025-soft},}
                        {\citet{vykopal2025largelanguagemodelsmultilingual},}
                        {\citet{10.1145/3627673.3679799},}
                        {\citet{weering2024fc_rug},}\\
                        {\citet{wu2023fake},}
                        {\citet{10.1145/3734520},}
                        {\citet{xie-etal-2025-fire},}
                        {\citet{yue-etal-2023-metaadapt},}
                        {\citet{yufeng2024factfinders},}\\
                        {\citet{zhang2024large},}
                        {\citet{zhao-etal-2024-pacar},}
                        {\citet{zhao-etal-2024-findver},}
                        {\citet{zhu-etal-2023-explain}}
                        , leaf, align = left, text width=27.2em
                    ]
                ]
                [
                    3-Class\\Classification
                    [
                        {\citet{abburi-etal-2025-deloitte},}
                        {\citet{abdallah2025ngu_research},}
                        {\citet{acosta2025ucom_unam_pln},}
                        {\citet{althabiti2024takeed},}
                        {\citet{alvarez-etal-2024-zero},}\\
                        {\citet{aly-etal-2023-qa},}
                        {\citet{anik2025claimiq},}
                        {\citet{10.1145/3726302.3729931},}
                        {\citet{bhatia2021automaticclaimreviewclimate},}
                        {\citet{bussotti-etal-2024-unknown},}\\
                        {\citet{chakraborty2023empirical},}
                        {\citet{choi2023automated},}
                        {\citet{choi2024factgpt},}
                        {\citet{dammu-etal-2024-claimver},}\\
                        {\citet{dmonte-etal-2025-gmu},}
                        {\citet{duesterwald2025cornellnlp},}
                        {\citet{10.1145/3565287.3617630},}
                        {\citet{gangi-reddy-etal-2022-newsclaims},}\\
                        {\citet{heil2025ds},}
                        {\citet{huang2024fakegptfakenewsgeneration},}
                        {\citet{sun2024trustllm},}
                        {\citet{JMLR:v24:23-0037},}\\
                        {\citet{kawamura-2025-team},}
                        {\citet{kolb2024authev},}
                        {\citet{la2022bum},}
                        {\citet{le2025lis},}\\
                        {\citet{li-etal-2025-minimal},}
                        {\citet{10.1145/3701716.3715599},}
                        {\citet{lu-etal-2023-scitab},}
                        {\citet{ma2023exfever},}
                        {\citet{malon-2021-team},}\\
                        {\citet{mamta-cocarascu-2025-facteval},}
                        {\citet{10.1145/3689212},}
                        {\citet{mirza2024globalliar},}
                        {\citet{petroni-etal-2021-kilt},}\\
                        {\citet{pradeep2020scientific},}
                        {\citet{10.1145/3404835.3463120},}
                        {\citet{purbey-etal-2025-1},}
                        {\citet{sarrouti-etal-2021-evidence-based},}\\
                        {\citet{scaiella-etal-2024-leveraging},}
                        {\citet{schlichtkrull2023averitec},}
                        {\citet{schuster-etal-2021-get},}
                        {\citet{si-etal-2024-checkwhy},}\\
                        {\citet{strong-etal-2024-zero},}
                        {\citet{tan2023evidencebased},}
                        {\citet{tan-etal-2025-improving},}
                        {\citet{10.1145/3705754.3705757},}
                        {\citet{10.1145/3626772.3657874},}\\
                        {\citet{vladika-etal-2025-step},}
                        {\citet{wang-etal-2023-check-covid},}
                        {\citet{wang2023explainable},}
                        {\citet{10.1145/3734520},}\\
                        {\citet{yue-etal-2024-retrieval},}
                        {\citet{zeng-gao-2023-prompt},}
                        {\citet{zeng2024justilm},}
                        {\citet{zeng2024maple},}\\
                        {\citet{zhang2024need},}
                        {\citet{zhang2023llmbased},}
                        {\citet{zhang-gao-2024-reinforcement},}
                        {\citet{zhao-etal-2024-pacar}}
                        , leaf, align = left, text width=27.2em
                    ]
                ]
                [
                    Multiclass (4+)\\Classification
                    [
                        {\citet{buchholz2023assessing},}
                        {\citet{hoes2023leveraging},}
                        {\citet{hoes2023using},}
                        {\citet{jiang-etal-2021-exploring-listwise},}
                        {\citet{kanaani-etal-2024-triple},}\\
                        {\citet{10.1145/3589335.3651521},}
                        {\citet{kao-yen-2024-magic},}
                        {\citet{leippold2024automated},}
                        {\citet{10411561},}\\
                        {\citet{liu-etal-2024-teller},}
                        {\citet{10.1145/3643491.3660290},}
                        {\citet{quelle2023perils},}
                        {\citet{schutz2022ait_fhstp},}\\
                        {\citet{shcharbakova-etal-2025-scale},}
                        {\citet{10.1145/3717867.3717896},}
                        {\citet{tan-etal-2025-improving},}
                        {\citet{tran2022ur},}\\
                        {\citet{zarharan-etal-2024-tell},}
                        {\citet{zhang2023llmbased},}
                        {\citet{zhang-gao-2024-reinforcement},}
                        {\citet{10.1145/3643562.3672613}}
                        , leaf, align = left, text width=27.2em
                    ]
                ]
                [
                    Multilabel\\Classification
                    [
                        {\citet{agrestia2022polimi}}
                        , leaf, align = left, text width=27.2em
                    ]
                ]
			]
            [
                Regression, text width=7.5em
                [
                    {\citet{10.1145/3565287.3617630},}
                    {\citet{guan2023language},}
                    {\citet{hu-etal-2025-decomposition},}
                    {\citet{jiang2023disinformation},}
                    {\citet{kolb2024authev},}
                    {\citet{li2023revisitfakenewsdataset},}\\
                    {\citet{pasin2024seupd},}
                    {\citet{pelrine2023towards},}
                    {\citet{setty2024endtoend},}
                    {\citet{vergho2024comparing}}
        			, leaf, align = left, text width=34.4em
                ]
			]
            [
                Ranking, text width=7.5em
                [
                    {\citet{10.1145/3565287.3617630},} % Move to ranking - rationale selection
                    {\citet{kamoi-etal-2023-wice},} % Move to ranking - rationale selection
                    {\citet{neumann-etal-2023-deep},}
                    {\citet{pasin2024seupd},}
                    {\citet{shliselberg2022riet},}\\
                    {\citet{singh2023utdrm},}
                    {\citet{tan2023evidencebased},} % Move to ranking - rationale selection
                    {\citet{wang-etal-2023-check-covid}} % Move to ranking - rationale selection
        			, leaf, align = left, text width=34.4em
                ]
                [
                    Classification\\Ranking
                    [
                        {\citet{FERNANDEZPICHEL2022105211},}
                        {\citet{jiang-etal-2021-exploring-listwise},}
                        {\citet{pradeep2020scientific},}
                        {\citet{10.1145/3404835.3463120},}\\
                        {\citet{sarrouti-etal-2021-evidence-based}}
                        , leaf, align = left, text width=27.2em
                    ]
                ]
            ]
            [
                Span Detection, text width=7.5em
                [
                    % {\citet{acosta2025ucom_unam_pln},(?)}
                    {\citet{gangi-reddy-etal-2022-zero},}
                    {\citet{gangi-reddy-etal-2022-newsclaims},}
                    {\citet{mittal-etal-2023-lost}}
        			, leaf, align = left, text width=34.4em
                ]
			]
			]
		\end{forest}
}
\caption{Taxonomy of structured output methods for fact-checking, including classification, regression, ranking, and span detection with representative studies.}
\label{fig:structured-taxonomy}
\end{figure*}

In automated fact-checking, prediction involves generating outputs by LLMs that can support decision-making or guide downstream fact-checking tasks. These predictions can be produced in a structured format, which allows for controlled, interpretable, and easily evaluable outputs. Structured outputs follow predefined schemas, such as categorical labels for claim verification, numerical scores representing the credibility or truthfulness of statements, ranked lists used for evidence retrieval or previously fact-checked claim retrieval, or textual spans for claim detection. Importantly, structured prediction does not involve generating arbitrary free-form text; instead, it restricts outputs to a controlled format that can be directly used by downstream modules and systematically evaluated. This restriction ensures consistency, interpretability, and reliability across multi-stage verification pipelines, making structured outputs central to automated fact-checking.

The key advantages of structured outputs lie in their ease of evaluation and integration. Since the results follow a known format, they can be directly compared against ground truth labels using standard metrics, such as accuracy, F1 score, or mean reciprocal rank (MRR). Moreover, their predictability makes them ideal for integration into downstream modules such as evidence retrieval, explanation generation, or credibility assessment. In addition, these methods offer higher execution efficiency, as LLMs are restricted to output only labels, values, or short spans, rather than providing free-form text. The greater availability of datasets with labeled targets, whether classes or scores, makes the structured output method more common, as these datasets provide us with supervision signals that are beneficial for fine-tuning.

Nevertheless, structured methods have several limitations. Constraining LLMs to predefined outputs can oversimplify the nuanced reasoning required for complex or ambiguous claims. Restricting outputs can also limit the model's ability to convey uncertainty or provide natural-language justifications. In practice, however, structured-output pipelines frequently include additional information alongside the predictions, such as confidence scores or textual rationales, allowing models to communicate uncertainty and reasoning while still maintaining controlled, evaluable outputs. Structured outputs may also be prone to producing inconsistent or invalid results across languages or topics, especially when prompts are not carefully standardized. Post-processing steps or rule-based validation are often necessary to ensure reliability. Moreover, structured outputs often do not leverage the LLM's full potential to generate human-readable text. Despite these challenges, structured output methods remain essential for quantitative benchmarking and for ensuring reproducibility and comparability across fact-checking datasets.

Within structured prediction, we distinguish four main categories illustrated in Figure~\ref{fig:structured-taxonomy}: \textit{classification}, \textit{regression}, \textit{ranking}, and \textit{span detection}. These categories are conceptually distinct yet interconnected. For example, ranking can be framed as repeated pairwise classification, while classification can be reformulated as regression with thresholds. The following sections discuss each category, synthesizing trends across prompting-based and fine-tuning approaches.

\paragraph{Classification.}

Classification-based methods are the most widely used structured approaches for prediction in fact-checking. Its prevalence stems from methodological convenience, as the majority of fact-checking datasets provide categorical labels aligned with the veracity labels used by fact-checkers. Tasks such as \textit{claim detection}, \textit{fact verification}, and \textit{previously fact-checked claim retrieval} are often formulated as binary or multi-class classification problems. The introduction of LLMs has transformed the approach to classification: rather than training specialized classifiers, researchers now rely on prompt engineering to induce label predictions from general-purpose LLMs. Unlike traditional classifiers, LLM-based classification can exploit natural language prompts to understand task semantics without requiring a task-specific architecture. By leveraging pre-trained knowledge and reasoning ability, LLMs can adapt to new domains or languages with minimal supervision. However, their generative nature introduces challenges, such as the fact that the output may not always align with the predefined label set, and decisions may depend on the phrasing of the prompt or the wording of the label. Despite these challenges, classification remains the most used method for producing structured output in fact-checking workflows.

One prominent group of approaches focuses on constraining model behavior through explicit task specification, aiming to improve label validity and output stability without relying on intermediate reasoning. \textbf{Zero-shot prompting}~\cite{NEURIPS2020_1457c0d6} provides a convenient baseline but often suffers from inconsistencies, especially when labels are semantically similar or when the task requires reasoning beyond simple lexical or syntactic cues. To address this, role specifications~\cite{10411561, 10.1145/3726302.3729931} assign the model a clear role and characteristics (e.g., fact-checker), improving task alignment, while explicit output constraints such as JSON schema~\cite{ma2023exfever, 10.1145/3731120.3744581} (e.g., \texttt{Check the claim: \{claim\}$\backslash n$Answer:}) enforce structured and valid responses. Techniques such as \textbf{rephrase and respond}~\cite{deng2024rephraserespondletlarge, modzelewski-etal-2025-pcot} further reduce ambiguity by prompting the model to rephrase the input before classification, encouraging semantic normalization. \textbf{Few-shot prompting}~\cite{NEURIPS2020_1457c0d6} extends this paradigm by introducing labeled demonstrations, improving consistency and accuracy when examples are well-chosen. However, performance is highly sensitive to the choice, ordering, and domain coverage, and studies have shown that the number of examples can influence outcomes~\cite{cao2023large, zeng-gao-2023-prompt, zeng2024justilm}.

Another group of methods enhances classification by introducing intermediate reasoning as an explicit step before label prediction. \textbf{Chain-of-though} (CoT)~\cite{10.5555/3600270.3602070} prompting encourages LLMs to provide their reasoning before producing a label, improving consistency on complex or multi-hop claims (i.e., claims that require reasoning across multiple pieces of evidence or intermediate steps to reach a conclusion)~\cite{pan-etal-2023-fact}. CoT can be applied either through explicit instructions (e.g.,"think step by step")~\cite{cao2023large, chen2023llmgenerated} or through few-shot demonstrations that exemplify the reasoning process~\cite{sawinski2023openfact, choi2023automated, pan2023qacheck, wang2023explainable, zhang2023llmbased, liu-etal-2024-teller}. While these approaches improve interpretability and consistency, they increase computational cost and may generate reasoning chains that are internally coherent yet factually incorrect. Extensions such as \textbf{self-consistency CoT}~\cite{ni-etal-2024-afacta} mitigate stochasticity by sampling and aggregating multiple reasoning paths, trading additional computation for robustness. Related techniques, including \textbf{reason-aware prompting}~\cite{huang2024fakegptfakenewsgeneration} and \textbf{few-shot logic prompting}~\cite{liu-etal-2024-teller}, guide predictions using structured summaries or decomposed subquestions, improving consistency at the cost of increased prompt complexity and potential bias inheritance.

Building on reasoning-based approaches, several methods introduce explicit critique or verification loops that separate initial prediction from validation, aiming to reduce factual errors. \textbf{Self-refine}~\cite{10.5555/3666122.3668141, kim2024llmsproducefaithfulexplanations} implements an iterative generate-critique-revise cycle, allowing the model to improve its own rationales, though errors from early iterations may propagate. \textbf{Program-of-Thoughts (PoT)}~\cite{chen2023programthoughtspromptingdisentangling, lu-etal-2023-scitab} offloads parts of the reasoning to executable code (e.g., Python snippets), enabling precise verification in numerical or symbolic domains, but requiring an execution environment. \textbf{Chain-of-Verification}~\cite{dhuliawala-etal-2024-chain, modzelewski-etal-2025-pcot} formalizes this separation by first producing an answer, then generating targeted verification questions, answering them (often with retrieved evidence), and finally revising the prediction. These methods improve traceability and factual grounding, but introduce additional complexity and dependencies on retrieval quality.

To further improve robustness, some approaches aggregate predictions across multiple prompts, reasoning paths, or models, trading efficiency for stability. \textbf{Ensemble of methods}~\cite{duesterwald2025cornellnlp, 10.1145/3726302.3730142} combine outputs from diverse strategies, reducing variance and mitigating individual model biases. While effective, ensembles increase computational cost and are less suitable for low-latency settings.

Beyond prompting-based techniques, \textbf{fine-tuning} remains an essential strategy for achieving consistent and domain-specific classification performance. Fine-tuned models typically outperform prompted ones in specialized domains or low-resource languages~\cite{zeng-gao-2023-prompt}. This technique is especially beneficial when models lack the necessary capabilities to effectively address a given task through simple prompting, particularly in complex domains where LLMs may have limited expertise. Fine-tuning strategies can be broadly divided into \textit{single-task} and \textit{multi-task fine-tuning}, depending on whether models are optimized for a single objective or jointly trained across multiple related tasks. \textbf{Single-task fine-tuning}~\cite{petroni-etal-2021-kilt, bhatia2021automaticclaimreviewclimate, agrestia2022polimi, althabiti2024takeed} achieves high accuracy but risks catastrophic forgetting~\cite{luo2024empiricalstudycatastrophicforgetting} and requires separate models for different tasks. In contrast, \textbf{multi-task fine-tuning}~\cite{du2022nus} encourages generalization by sharing representations across related objectives, enabling knowledge transfer even when multiple tasks are jointly optimized.

Finally, because factual classification depends heavily on access to reliable evidence, many approaches augment classification with external information sources or tool use. In \textit{open-book} settings~\cite{schlichtkrull2023averitec}, retrieved evidence is incorporated directly into prompts using zero-shot~\cite{strong-etal-2024-zero, 10.1145/3565287.3617630}, few-shot~\cite{10.1145/3705754.3705757, pan2023qacheck}, or CoT-based prompting~\cite{zhang-gao-2024-reinforcement, 10.1145/3627673.3679985}. When multiple evidence items are available, methods either verify claims against each piece individually and aggregate the predictions, or jointly include all evidence in a single prompt, with the latter generally yielding better results. Beyond text, tabular evidence is typically linearized before being processed by LLMs~\cite{zhang2024large}. More advanced frameworks, such as \textbf{ReAct} (Reason + Act)~\cite{yao2022react}, enable models to interleave reasoning with actions, dynamically retrieving and verifying information~\cite{quelle2023perils, zhang2023llmbased, zhang-gao-2024-reinforcement}. Similarly, \textbf{verify-and-edit}~\cite{zhao-etal-2023-verify, zhang-gao-2024-reinforcement} explicitly identify uncertain predictions and seek additional evidence before revision. Across these approaches, the main objective is to ensure that classification outputs are both structurally consistent and grounded in verifiable facts.

\paragraph{Regression.}

In regression-based approaches, LLMs produce continuous scores over a predefined scale, such as 0-1~\cite{setty2024endtoend, hu-etal-2025-decomposition} or 0-100~\cite{pelrine2023towards, li2023revisitfakenewsdataset, vergho2024comparing}, reflecting degrees of veracity, relevance, or confidence. These scores aim to quantify the model's confidence or provide a fine-grained measure of factual consistency. In practice, regression is often treated as a classification problem, since most fact-checking datasets provide only categorical labels rather than continuous ground-truth scores~\cite{li2023revisitfakenewsdataset, vergho2024comparing, setty2024endtoend}. In such scenarios, LLM-generated scores are discretized through thresholding, commonly at 0.5 for binary labels, transforming continuous predictions into categorical outputs. This thresholding mechanism provides flexibility, enabling researchers to adjust the mapping from continuous scores to categorical labels for improved accuracy, calibration, or task-specific performance~\cite{pelrine2023towards}. Alternative designs include bounded scoring ranges such as -2 and 2~\cite{pasin2024seupd}, to indicate whether a claim contradicts, is neutral toward, or supports a specific statement under verification.

Regression is particularly well-suited for tasks requiring evaluation of relevance or similarity, such as \textit{previously fact-checked claim retrieval} and \textit{evidence retrieval}. Here, continuous scores naturally represent relevance or semantic similarity, supporting effective ranking or prioritization. In \textit{fact verification}, regression is applied less frequently but offers a finer-grained depiction of claim truthfulness compared to discrete labels~\cite{guan2023language, jiang2023disinformation, 10.1145/3565287.3617630}.

The main advantage of regression-based methods lies in their ability to capture subtle variations in confidence or relevance, which can improve ranking performance and decision calibration. However, these advantages come with trade-offs. Regression outputs can be harder to interpret, since minor differences in scores do not always reflect meaningful differences in claim truthfulness or relevance. Moreover, without annotated score-based datasets, model evaluation often relies on thresholding decisions that may be somewhat arbitrary, limiting comparability across studies. Consequently, while regression can add flexibility and nuance to fact-checking systems, it remains a secondary formulation compared to categorical classification.

In terms of implementation, zero-shot prompting dominates regression applications, reflecting the low availability of continuous-score datasets (9$\times$), followed by fine-tuning (2$\times$) and few-shot prompting (1$\times$). When using prompts, researchers typically specify the desired scoring range for the LLM, which can vary depending on the dataset or task, ensuring that outputs align with the intended scale. Fine-tuning strategies are also employed. For instance, \citet{pelrine2023towards} fine-tuned LLMs via API to predict truthfulness on a 0-100 scale. This demonstrates how instruction-based fine-tuning can combine structured outputs with continuous reasoning.

\paragraph{Ranking.}
\label{par:ranking}

Ranking approaches order claims, documents, or evidence according to their estimated relevance to a given query or target claim~\cite{shliselberg2022riet, neumann-etal-2023-deep, pasin2024seupd}, and they play a central role in retrieval-oriented tasks such as \textit{previously fact-checked claim retrieval} and \textit{evidence retrieval}. Although ranking is explored less frequently than classification and regression, it provides graded measures of relevance or factual support, effectively connecting predictive outputs with reasoning-based evidence evaluation.

One common strategy, which we term \textbf{classification-based ranking}, leverages LLM output probabilities or logits over discrete labels (e.g., \texttt{true}, \texttt{false}) as ranking signals~\cite{10.1145/3404835.3463120}. For each candidate instance,
the model produces a classification, and confidence scores are used to rank items by relevance for retrieval tasks or likelihood of being factually correct for claim verification. This approach combines the interpretability of categorical outputs with the ability to prioritize information within retrieval pipelines.

A second approach is \textbf{rationale selection}, in which LLMs are instructed to identify or rank the most relevant evidence segments from retrieved documents using structured outputs such as sentence indices or relevance scores~\cite{10.1145/3565287.3617630, tan2023evidencebased, wang-etal-2023-check-covid, kamoi-etal-2023-wice}. Instead of generating free-text rationales, the model evaluates candidate sentences based on semantic relevance. For example, \citet{wang-etal-2023-check-covid} and \citet{10.1145/3565287.3617630} instruct LLMs to produce a list of sentence IDs that support or refute a claim. Other studies assign relevance scores to claim-evidence pairs using NLI-trained models and select sentences above a threshold~\cite{kamoi-etal-2023-wice}. This ranking-based formulation exploits LLMs' comprehension and reasoning to prioritize evidence that humans would likely consider relevant to a claim.

Ranking methods offer several advantages. They provide graded relevance signals, facilitate integration with retrieval pipelines, and support more nuanced decision-making than strict classification. At the same time, ranking faces challenges. Evaluation can be complex, as metrics such as mean reciprocal rank (MRR) or nDCG require human-annotated relevance judgments. Additionally, LLM-generated confidence scores can be unstable, sensitive to prompt phrasing or contextual variations. Despite these limitations, ranking serves as an important bridge between classification and reasoning-based evidence selection in retrieval-augmented fact-checking systems.

Implementation techniques vary. Classification-based ranking typically relies on fine-tuning of encoder-decoder models to predict binary relevance for each document, using prompts such as "\texttt{Query: q Document d Relevant:}" followed by model completion. Two-step fine-tuning, first on large retrieval benchmarks such as MS MARCO~\cite{nguyen2016ms}, and then on domain-specific datasets, has proven effective, especially for evidence retrieval~\cite{10.1145/3404835.3463120}. Beyond classification-based ranking, generative re-ranking approaches estimate conditional probabilities between candidate texts~\cite{shliselberg2022riet, neumann-etal-2023-deep}, refining ordering based on semantic similarity or factual alignment. Zero-shot strategies have also been applied. For instance, multiple paraphrased queries can be generated and scored for relevance to candidate claims on a 0-5 scale, and these LLM-assigned scores are then combined with traditional retrieval methods such as BM25~\cite{pasin2024seupd}.

\paragraph{Span Detection.}

Span detection focuses on identifying and generating the specific segments of the input text that correspond to a factual claim, while restricting itself to selecting parts of the original text that encode factual statements. The output is typically expressed as text spans, natural-language extracts, or directly extracted sentences, rather than as free-standing claims. Generative LLMs are typically prompted to identify which parts of the input are check-worthy, meaning they contain verifiable factual claims. This supports fact-checking pipelines by isolating only those spans that should be processed further.

Span detection falls conceptually between classification-based claim detection and claim generation. While traditional approaches classify sentences or entire texts as factual or not, span detection methods generate textual segments directly, allowing for greater flexibility when claims span partial sentences or when the boundaries of factual content do not align with syntactic units. 

The primary benefit of span detection is improved precision in downstream tasks. By focusing generative methods on only the relevant portion of the text, systems reduce noise, avoid unnecessary rewriting of non-factual content, and improve the fidelity of the claims. However, generative span detection introduces challenges in maintaining semantic boundaries. LLMs may truncate key information, combine multiple propositions into a single span, or misidentify rhetorical or speculative passages as factual. These errors propagate to subsequent steps, making robustness and consistency crucial.

Span detection is typically addressed through LLM prompting, where the LLMs are instructed to identify and extract check-worthy factual segments from the input text. \textbf{Zero-shot} and \textbf{few-shot prompting} are the dominant strategies, enabling LLMs to identify spans without paraphrasing text into a claim. For example, \citet{gangi-reddy-etal-2022-zero} evaluated both prompting techniques for detecting factual statements in text and showed that few-shot examples help models better recognize the structure and boundaries of factual claims. Their work also explores \textbf{fine-tuning} with zero-shot instructions as supervision, demonstrating that even lightweight supervision can improve span selection when evaluated under few-shot prompting. Similarly, \citet{mittal-etal-2023-lost} employed instruction-based prompting to obtain factual spans from social media posts. Their setup allows LLMs to output spans that are not strictly extractive, sometimes including paraphrases or synonyms, illustrating how generative LLMs can approximate span boundaries even when the extracted text does not perfectly match the original input. 

\subsection{Methods with Unstructured Output for Prediction}
\label{sec:unstructured}

\begin{figure*}
\centering
\resizebox{\textwidth}{!}{
\begin{forest}
			forked edges,
			for tree={
				grow=east,
				reversed=true,
				anchor=base west,
				parent anchor=east,
				child anchor=west,
                node options={align=center},
                align = center,
				base=left,
				font=\small,
				rectangle,
				draw=hidden-draw,
				rounded corners,
				minimum width=4em,
				edge+={darkgray, line width=1pt},
				s sep=3pt,
				inner xsep=2pt,
				inner ysep=3pt,
				ver/.style={rotate=90, child anchor=north, parent anchor=south, anchor=center},
			},
			where level=1{text width=5.0em,font=\scriptsize}{},
			where level=2{text width=5.6em,font=\scriptsize}{},
			where level=3{text width=6.8em,font=\scriptsize}{},
			[
			Methods with Unstructured Outputs, ver
            [
                Claim Generation, text width=7.5em
                [
                    Normalization, text width=7.5em
                    [
                        {\citet{almada2025akcit},}
                        {\citet{amatya2025factiverse},}
                        {\citet{anikina2025dfkinit2b},}
                        {\citet{beltran2025umuteam},}\\
                        {\citet{hashmi2025investigators},}
                        {\citet{pramov2025ds},}
                        {\citet{saeed2025mma},}
                        {\citet{sawinski2025openfact},}\\
                        {\citet{sundriyal2023chaos},}
                        {\citet{vineetha2025saivineetha},}
                        {\citet{wilder2025unh}}
            			, leaf, align = left, text width=25.3em
                    ]
    			]
                [
                    Decomposition, text width=7.5em
                    [
                        {\citet{10.1145/3654777.3676359},}
                        {\citet{heil2025ds},}
                        {\citet{hu-etal-2025-decomposition},}
                        {\citet{kamoi-etal-2023-wice},}
                        {\citet{li-etal-2024-self},}\\
                        {\citet{metropolitansky-larson-2025-towards},}
                        {\citet{ni-etal-2024-afacta},}
                        {\citet{strong-etal-2024-zero},}
                        {\citet{zhao-etal-2024-pacar}}
            			, leaf, align = left, text width=25.3em
                    ]
    			]
            ]
            [
                Question Generation, text width=7.5em
                [
                    {\citet{10.1145/3539618.3591907},}
                    {\citet{chen-etal-2024-complex},}
                    {\citet{chen-etal-2022-generating},}
                    {\citet{le2025lis},}
                    {\citet{10.1145/3696410.3714748},}
                    {\citet{pan2023qacheck},}\\
                    {\citet{schlichtkrull2023averitec},}
                    {\citet{10.1145/3627673.3679985},}
                    {\citet{setty2024endtoend},}
                    {\citet{10.1145/3726302.3730142},}
                    {\citet{10.1145/3626772.3657874},}
                    {\citet{zhang-gao-2024-reinforcement}}
        			, leaf, align = left, text width=34.4em
                ]
			]
            [
                Query Generation, text width=7.5em
                [
                    {\citet{abdallah2025ngu_research},}
                    {\citet{arana-catania-etal-2022-natural},}
                    {\citet{li-etal-2025-imrrf},}
                    {\citet{malon-2021-team},}
                    {\citet{prietochavana2023automated}}
        			, leaf, align = left, text width=34.4em
                ]
            ]
           %  [
           %      Rationale Selection, text width=7.5em
           %      [
           %          {\citet{fukuoka2025ksu},}
           %          {\citet{ko-etal-2023-claimdiff},}
           %          {\citet{pradeep2020scientific}}
           %          % {\citet{kamoi-etal-2023-wice},} % Move to ranking
           %          % {\citet{wang-etal-2023-check-covid},} % Move to ranking
           %          % {\citet{10.1145/3565287.3617630},} % Move to ranking
           %          % {\citet{tan2023evidencebased},} % Move to ranking
        			% , leaf, align = left, text width=34.4em
           %      ]
           %  ]
            [
                Summary Generation, text width=7.5em
                [
                    {\citet{chen-etal-2024-complex},}
                    {\citet{10.1145/3589335.3651914},}
                    {\citet{10.1145/3731120.3744581},}
                    {\citet{li-etal-2025-imrrf},}
                    {\citet{10.1145/3726302.3730092},}
                    {\citet{10.1145/3726302.3729960},}\\
                    {\citet{zarharan-etal-2024-tell},}
                    {\citet{zeng-etal-2024-ru22fact}} 
                    , leaf, align = left, text width=34.4em   
                ]
            ]
            [
                Explanation Generation, text width=7.5em
                [
                    {\citet{abburi-etal-2025-deloitte},}
                    {\citet{10.1007/978-3-031-47896-3_1},}
                    {\citet{althabiti2024takeed},}
                    {\citet{bang2023multitask},}
                    {\citet{10.1145/3726302.3729931},}
                    {\citet{bhatia2021automaticclaimreviewclimate},}\\
                    {\citet{cekinel2024explaining},}
                    {\citet{dammu-etal-2024-claimver},}
                    {\citet{10.1145/3696410.3714934},}
                    {\citet{dmonte-etal-2025-gmu},}
                    {\citet{dougrez-lewis-etal-2025-assessing},}\\
                    {\citet{duesterwald2025cornellnlp},}
                    {\citet{eren2024turquaz},}
                    {\citet{10.1145/3654777.3676359},}
                    {\citet{hsu-etal-2023-explanation},}
                    {\citet{irnawan-etal-2025-claim},}
                    {\citet{jiang2023disinformation},}\\
                    {\citet{kanaani-etal-2024-triple},}
                    {\citet{10.1145/3589335.3651521},}
                    {\citet{10.1145/3589335.3651521},}
                    {\citet{kawamura-2025-team},}
                    {\citet{kim2024llmsproducefaithfulexplanations},}
                    {\citet{li2023revisitfakenewsdataset},}\\
                    {\citet{liu2024large},}
                    {\citet{10.1145/3701716.3715599},}
                    {\citet{ma2023exfever},}
                    {\citet{10.1145/3696410.3714748},}
                    {\citet{pan2023qacheck},}
                    {\citet{pan-etal-2023-fact},}
                    {\citet{pelrine2023towards},}\\
                    {\citet{pham-etal-2025-claimpkg},}
                    {\citet{10.1145/3699682.3728349},}
                    {\citet{purbey-etal-2025-1},}
                    {\citet{russo-etal-2025-euroverdict},}
                    {\citet{russo-etal-2023-benchmarking},}\\
                    {\citet{schlichtkrull2023averitec},}
                    {\citet{setty2024endtoend},}
                    {\citet{si-etal-2024-large},}
                    {\citet{10.1145/3717867.3717896},}
                    {\citet{tan2023evidencebased},}
                    {\citet{tan-etal-2025-improving},}\\
                    {\citet{10.1145/3626772.3657874},}
                    {\citet{vergho2024comparing},}
                    {\citet{vladika-etal-2025-step},}
                    {\citet{wang2023explainable},}
                    {\citet{10.1145/3725899.3727894},}
                    {\citet{10.1145/3722570.3726897},}\\
                    {\citet{yue-etal-2024-evidence},}
                    {\citet{yue-etal-2024-retrieval},}
                    {\citet{zarharan-etal-2024-tell},}
                    {\citet{zeng2024justilm},}
                    {\citet{zeng-etal-2024-ru22fact},}
                    {\citet{zhao-etal-2024-findver}}
        			, leaf, align = left, text width=34.4em
                ]
            ]
            [
                Other, text width=7.5em
                [   
                    {\citet{deng-etal-2024-document},}
                    {\citet{fukuoka2025ksu},} % Rationale selection
                    {\citet{gao-etal-2023-precise},}
                    {\citet{ko-etal-2023-claimdiff},} % Rationale selection
                    {\citet{kotitsas-etal-2024-leveraging},}
                    {\citet{10.1145/3627673.3679519},}\\ % comments generation
                    {\citet{pan-etal-2023-fact},}
                    {\citet{pradeep2020scientific},} % Rationale selection
                    {\citet{schuster-etal-2021-get}}
                    % {\citet{abdallah2025ngu_research}}
        			, leaf, align = left, text width=34.4em
                ]
            ]
			]
		\end{forest}
}
\caption{Taxonomy of generative LLM methods with unstructured outputs for fact-checking, grouped by their primary generation task, with representative studies shown for each category.}
\label{fig:unstructured-taxonomy}
\end{figure*}

% {\citet{dammu-etal-2024-claimver}}, {\citet{yue-etal-2024-retrieval}}, {\citet{sahnan2025llmsautomatefactcheckingarticle}}

Methods with unstructured outputs leverage the generative capabilities of LLMs to produce free-form text rather than fixed labels or numerical values. In the context of fact-checking, such outputs include claim normalization, claim decomposition, question or query generation, rationale selection, explanation generation, or even the generation of entire fact-check articles. Rather than producing a single decision signal, these methods generate intermediate or auxiliary textual artifacts that can support reasoning, information retrieval, or downstream verification steps.

The purpose of unstructured outputs is therefore not limited to end-user explanations. While some outputs are designed to improve interpretability or communicate reasoning to human users, many are intermediate representations intended to guide subsequent model components or steps, improve evidence alignment, or structure complex verification processes. Compared to structured approaches that emphasize categorical predictions, unstructured methods focus on contextualization, semantic decomposition, and argumentation, reflecting the multi-step and human-centered nature of fact-checking workflows.

These methods are typically realized through prompt-based approaches that encourage LLMs to produce natural language responses aligned with specific roles or objectives. For example, models may be instructed to provide an explanation or justification for the veracity predicted by the LLM~\cite{bhatia2021automaticclaimreviewclimate, schlichtkrull2023averitec, pan-etal-2023-fact}.

The main advantages of unstructured methods are in their expressiveness and their ability to provide model reasoning. By expressing intermediate reasoning steps, these outputs can support transparency and trust in automated fact-checking systems and assist human fact-checkers in interpreting model predictions. Moreover, unstructured outputs can bridge multiple subtasks, such as explanation generation, evidence synthesis, and claim verification, within a single generative framework~\cite{tan2023evidencebased, ma2023exfever}. This makes them particularly well-suited for complex claims that require multi-step reasoning or contextual grounding.

However, compared to the methods with structured output, these methods bring notable challenges. Free-form text is more difficult to evaluate automatically, as common metrics such as BLEU~\cite{papineni-etal-2002-bleu} or ROUGE~\cite{lin-2004-rouge} often fail to capture factual correctness or reasoning quality. As a result, many studies rely on human evaluation or fact-based verification. Moreover, LLMs may generate hallucinated,  inconsistent, or post-hoc rationales that do not faithfully reflect the model's underlying decision process~\cite{rahman2025hallucinationtruthreviewfactchecking}. These issues raise concerns about the faithfulness, reliability, and misuse, especially in high-stakes fact-checking scenarios.

In this section, we categorize the methods with unstructured outputs into six categories: \textit{claim generation}, \textit{question generation}, \textit{query generation}, \textit{summary generation}, \textit{explanation generation}, and other forms of \textit{text generation}.

\paragraph{Claim Generation.}

Claim generation refers to the process of transforming raw, user-generated content (e.g., social media posts, political statements, or long-form texts) into one or more explicit, verifiable claims that can be processed by automated fact-checking systems. This step is usually performed since real-world inputs are rarely expressed as clean, standalone factual assertions. They are often noisy, implicit, opinionated, or contain multiple interconnected propositions that cannot be verified reliably in their original form. As a result, downstream components such as evidence retrieval, claim matching, and veracity prediction operate more effectively when the input is first converted into a clear and structured claim representation. In this survey, \emph{claim} is defined as a declarative statement that asserts a factual proposition about the world and can be verified against external evidence. Verifiable claims are typically explicit, factual, minimally ambiguous, and faithful to the original input, while avoiding subjective opinions, rhetorical framing, or vague references. Generative LLMs are increasingly used to perform claim generation as an intermediate step between claim detection and downstream retrieval or verification modules.

Prior work operationalizes claim generation mainly through two closely related strategies: \textit{claim normalization} and \textit{claim decomposition}. Claim normalization focuses on rewriting an input into a single, standardized, concise, and unambiguous claim by removing conversational markers, resolving pronouns, simplifying linguistic structure, and reducing emotional or stylistic variability~\cite{sundriyal2023chaos}. This approach is particularly important for social media posts, which are often noisy, informal, and embedded in longer texts. Normalization has been shown to improve semantic alignment between user claims and fact-check databases, especially for previously fact-checked claim retrieval and multilingual settings~\cite{anikina2025dfkinit2b, amatya2025factiverse, almada2025akcit, pramov2025ds}. In contrast, claim decomposition aims to split complex, compound, or ambiguous inputs into a set of smaller, atomic subclaims that can be verified independently~\cite{kamoi-etal-2023-wice, ni-etal-2024-afacta}. Decomposition is especially valuable for political claims, scientific texts, and long-form narratives that contain multiple factual statements, implicit assumptions, or intertwined propositions. While normalization typically produces a single reformulated claim and decomposition yields multiple subclaims, both approaches pursue the same objective: generating faithful, explicit, and verifiable representations of the original content.

The benefits of claim generation are reflected in improved robustness and effectiveness of downstream fact-checking steps. Normalized claims reduce linguistic noise and phrasing variability, which can lead to higher recall and precision in retrieval-based pipelines~\cite{anikina2025dfkinit2b}. Decomposed claims improve granularity, enabling systems to retrieve targeted evidence, detect partial truths, and aggregate verdicts across multiple factual statements. These advantages make claim generation a crucial bridge between processing and verification, and they align automated pipelines more closely with human fact-checking workflows, in which claims are clarified and decomposed before verification.

Despite these benefits, claim generation with LLMs introduces several limitations. Both normalization and decomposition are prone to semantic drift, where models may omit important details, hallucinate unstated content, or unintentionally alter the meaning or strength of the original claim~\cite{sundriyal2023chaos}. Normalization may collapse multiple independent statements into a single claim, thereby losing the granularity that is required for accurate verification~\cite{anikina2025dfkinit2b}. Decomposition, on the other hand, may suffer from over-segmentation, producing redundant or overly fine-grained subclaims, or under-segmentation, where distinct propositions remain merged~\cite{kamoi-etal-2023-wice, ni-etal-2024-afacta}. Ensuring faithfulness, minimality, and appropriate granularity across generated claims remains an open challenge.

Techniques for claim generation are predominantly based on prompting strategies, reflecting the flexibility of generative LLMs to reformulate and restructure input text. The most widely used approach is \textbf{zero-shot prompting}, where the model is directly instructed to rewrite and input into a clear, verifiable claim~\cite{amatya2025factiverse, almada2025akcit, sawinski2023openfact, saeed2025mma, vineetha2025saivineetha, anikina2025dfkinit2b} or to enumerate atomic subclaims~\cite{heil2025ds, strong-etal-2024-zero}. These prompts typically include high-level instructions such as removing subjective language, resolving references, or listing factual statements~\cite{anikina2025dfkinit2b}. To improve consistency and adherence to task-specific constraints, many studies adopt \textbf{few-shot prompting}, where the LLM is provided with annotated examples of normalized claims~\cite{sundriyal2023chaos, amatya2025factiverse, anikina2025dfkinit2b} or decomposed subclaims~\cite{kamoi-etal-2023-wice, 10.1145/3654777.3676359}. Few-shot examples help constrain the desired level of explicitness, minimality, and style, and are particularly effective in multilingual and low-resource settings such as Telugu, where zero-shot performance is often unstable~\cite{anikina2025dfkinit2b}. Across both normalization and decomposition, few-shot prompting has been shown to reduce output variability and better preserve the intended meaning of the original input. Several works further enhance prompting with \textbf{CoT prompting}, encouraging models to reason explicitly about which parts of the input constitute factual claims and how they should be rewritten or segmented~\cite{pramov2025ds, wilder2025unh, anikina2025dfkinit2b}. In claim normalization, CoT is often used to guide the model toward identifying the most salient factual content and mitigating the omission of important details~\cite{sundriyal2023chaos}.

Besides the standard prompting techniques, more advanced strategies introduce iterative refinement mechanism. Approaches such as \textbf{self-reflection}~\cite{sawinski2025openfact} and \textbf{self-refinement}~\cite{wilder2025unh} first generate one or more candidate claims and then use the same LLM to evaluate and revise these outputs based on predefined criteria, such as clarity, check-worthiness, or faithfulness to the input. These methods aim to reduce semantic drift and improve claim quality by explicitly incorporating feedback loops, but they require multiple model invocations and are therefore more computationally expensive.

Several studies have also compared different decomposition techniques and frameworks to understand their impact on downstream fact verification performance~\cite{hu-etal-2025-decomposition, metropolitansky-larson-2025-towards}. Methods such as \textit{FactScore}~\cite{min-etal-2023-factscore}, \textit{VeriScore}~\cite{song-etal-2024-veriscore}, \textit{WICE decomposer}~\cite{kamoi-etal-2023-wice}, \textit{DnD}~\cite{wanner-etal-2025-dndscore}, \textit{SAFE}~\cite{wei2024longformfactualitylargelanguage}, \textit{Fact-check-GPT}~\cite{wang-etal-2024-factcheck}, or \textit{AFaCTA}~\cite{ni-etal-2024-afacta} were originally designed to assess factual consistency of generated text. Researchers have reused their decomposition modules to segment input claims, exploiting their structured prompting patterns for identifying minimal verifiable units. These methods typically guide the LLM toward more principled decomposition by emphasizing explicit reasoning steps, stepwise extraction, or the separation of implicit assumptions from explicit statements.

Finally, some work explores \textbf{fine-tuning} for claim generation, often using parameter-efficient fine-tuning methods~\cite{anikina2025dfkinit2b, saeed2025mma, wilder2025unh}. Fine-tuning is typically performed in monolingual scenarios where labeled training data is available~\cite{alam2025overview}. In addition to fine-tuning, \citet{anikina2025dfkinit2b} investigated the effectiveness of the \textbf{ensemble of methods} for claim normalization, where they combined the outputs from various strategies (e.g, fine-tuning, zero-shot, and few-shot prompting), they calculated the centroid based on the embeddings of each normalized claim, and finally, they selected that normalized claim that was the closest to the centroid as the final normalized claim. While this approach proved very promising in achieving the best scores across languages, it is ineffective in terms of the computational resources required to produce the normalized claim.

\paragraph{Question Generation.}

In question generation, LLMs produce questions intended to guide the fact-checking process by identifying missing information or directing evidence retrieval. Typical outputs include wh-questions (e.g., \texttt{Who is Bruce Lee?}~\cite{zhang-gao-2024-reinforcement}) and binary questions (e.g., \texttt{Did the murder rate in 2020 increase by 26\% from 2019?}~\cite{chen-etal-2022-generating}). This approach naturally builds on claim decomposition: once a claim is broken into subclaims, the LLM can generate targeted questions that explicitly state what must be verified. Such questions help structure evidence retrieval and support multi-hop reasoning, where each question corresponds to a distinct verification step.

Question generation primarily identifies information worthy of verification and later guides the subsequent \textit{evidence retrieval} process. Making information needs explicit encourages systematic information gathering and can improve downstream retrieval quality. In multi-hop verification settings, generated questions often serve as intermediate reasoning steps, decomposing complex claims into a sequence of simple inquiries. In some systems, these questions also contribute to interpretability by making the verification strategy explicit, even when the questions themselves are not directly user-facing.

The main advantage of question generation is its ability to impose structure on inherently ambiguous verification tasks. Well-formed questions can uncover implicit assumptions, highlighting missing context, and align retrieval with the information requirements of the claim. However, generative LLMs may also produce irrelevant, overly generic, or hallucinated questions, which can misdirect retrieval and reasoning. Poorly framed questions may introduce false premises or reflect confirmation bias, ultimately degrading downstream verification performance.

Most question generation methods rely on prompting-based strategies. \textbf{Zero-shot prompting} is commonly used to elicit yes/no, wh-, or open-ended questions directly from a given input, often paired with role descriptions or explicit guidelines to ensure that generated questions are clear, self-contained, and relevant~\cite{le2025lis}. \textbf{Few-shot prompting} further improves control and diversity by providing examples of claims paired with corresponding questions~\cite{chen-etal-2024-complex}. This helps the model cover different aspects of the input and avoid generic or repetitive outputs. Some approaches further refine outputs by including examples of good and bad questions for the same claim~\cite{le2025lis} or enforcing structured formats such as JSON~\cite{10.1145/3627673.3679985}. Similar prompting strategies are used in question-generation-based decomposers, where in-context examples demonstrate how to rewrite claims into yes/no questions~\cite{chen-etal-2022-generating, 10.1145/3626772.3657874}.

More advanced techniques incorporate explicit reasoning into question generation. \textbf{CoT prompting} encourages LLMs to reason about what information is needed before generating questions, leading to more comprehensive coverage of claim components~\cite{10.1145/3626772.3657874}. Question decomposition prompts similarly break complex claims into multiple targeted inquiries, supporting multi-hop verification and improving the alignment between questions and subclaims~\cite{setty2024endtoend}. \textbf{Iterative prompting} extends this idea by feeding previously generated questions and their answers back into the model to elicit follow-up questions, enabling multi-turn investigation~\cite{pan2023qacheck}.

A smaller subset of approaches incorporates \textbf{contextual augmentation}, in which question generation is enriched with retrieved evidence or intermediate outputs from earlier stages. This additional context allows models to refine their questions based on what is already known, bridge gaps between claims and partially relevant evidence, and support more effective reranking or evidence selection~\cite{schlichtkrull2023averitec}. 

\paragraph{Query Generation.}

Query generation refers to prompting LLMs to produce search or keyword-based queries that can be used for search engines, fact-check archives, or document retrievers~\cite{prietochavana2023automated}. Unlike questions, queries are optimized for retrieval performance rather than interpretability, and typically consist of condensed keyword combinations or brief natural language expressions.

Query generation plays a crucial role in \textit{evidence retrieval}. By translating claims into retrieval-friendly formulations, LLMs can substantially improve recall and reduce the need for manual query optimization. This is particularly useful in multilingual pipelines, where queries must adapt to language-specific behavior or content availability across languages.

The main advantage of query generation is improved retrieval quality, since well-structured queries capture the core meaning of a claim while filtering out irrelevant phrasing. Generating diversified queries, that is, multiple query variants for the same claim, can improve retrieval coverage~\cite{prietochavana2023automated}. However, generated queries may introduce false details, omit key entities, or be overly verbose. Errors at the retrieval stage can propagate through the entire pipeline, making robustness a critical concern.

% Techniques used summary
LLMs are often instructed to translate claims into queries for retrieval using carefully crafted prompts. The generated queries typically need to satisfy several requirements to support high-quality search results, such as conciseness, entity coverage, and relevance. In \textbf{zero-shot settings}, instructions such as \texttt{"Generate search query:"} or \texttt{"Search query:"} are prepended to the input to elicit concise, keyword-focused queries directly from the claim~\cite{prietochavana2023automated}. \textbf{Few-shot prompting} introduces examples of well-formed queries, helping the model to consistently capture key entities, dates, numbers, or causal relationships while avoiding irrelevant or verbose outputs~\cite{arana-catania-etal-2022-natural}.

Beyond basic prompting, \citet{prietochavana2023automated} experimented with more structured strategies that incorporate additional contextual information. For instance, LLMs can be conditioned on prior verification steps, domain-specific instructions, or previously retrieved evidence, resulting in queries that are better aligned with search behavior and retrieval objectives. In addition, the authors fine-tuned the LLM to adapt to specific prefixes, query patterns, and domain-specific characteristics.

\paragraph{Summary Generation.}

Summary generation involves producing concise, human-readable summaries of textual evidence or entire fact-checking articles. LLMs are instructed to condense long-form content into short summaries that preserve key factual information, verification verdicts, and claim-relevant context~\cite{zeng-etal-2024-ru22fact, zarharan-etal-2024-tell, 10.1145/3589335.3651914}. This is particularly useful in scenarios where fact-checkers or end users need a quick understanding of verification outcomes, such as in social media monitoring or the analysis of multilingual misinformation.

The primary advantage of summary generation is its ability to translate high-volume content into human-readable insights. Summarization can integrate evidence from multiple sources, highlight agreements and contradictions among them, and provide coherent summaries that are consistent with verification judgments or fact-checking verdicts. In addition, summaries can serve as input for downstream modules~\cite{li-etal-2025-imrrf}, such as explanation generation, by reducing information overload while preserving key facts. On the other hand, challenges include hallucination, omission of critical details, and misrepresentations of nuanced claims, which can undermine verification reliability.

Summary generation often involves prompting the LLM to distill long or complex content into a compact summary that emphasizes claim-relevant information. \textbf{Zero-shot prompts} are widely used to instruct the model to focus on essential evidence while filtering out irrelevant context, with constraints such as word limits or structured output formats like JSON~\cite{zeng-etal-2024-ru22fact, zarharan-etal-2024-tell, 10.1145/3717867.3717896, 10.1145/3726302.3729960}, to facilitate downstream processing. In particular, \citet{10.1145/3726302.3729960} explicitly instructed the model to avoid generating content that is unethical, sexist, racist, or toxic, ensuring socially unbiased summaries. Additionally, \textbf{few-shot prompting} further guides models by providing examples of summaries, including examples based on both relevant and deliberately irrelevant documents. By exposing the model to documents that are not directly useful for verification, the LLM learns to prioritize claim-relevant evidence and internally filter out irrelevant content, improving the focus and quality of generated summaries~\cite{chen-etal-2024-complex}. This approach also helps highlight evidence that is directly relevant to the claim while maintaining coherence across multiple sources~\cite{chen-etal-2024-complex}.

\textbf{Fine-tuning} has been explored in several studies to enhance consistency, accuracy, and task-specific summarization behaviors, ensuring that key facts are preserved while unnecessary details are omitted~\cite{10.1145/3589335.3651914}. By shaping the model to task-specific patterns, fine-tuning can improve reliability for domain-specific or multilingual summarization tasks.

\paragraph{Explanation Generation.}

Explanation generation is a widely studied unstructured method in automated fact-checking that focuses on producing textual justification for model outputs~\cite{ma2023exfever, 10.1007/978-3-031-47896-3_1, 10.1145/3589335.3651521}. Unlike chain-of-thought reasoning, which is typically used as an internal mechanism to improve model predictions and is not exposed to end users~\cite{wang2023explainable,pelrine2023towards}, explanation generation produces outputs intended to be user-facing. These explanations serve to communicate the veracity of a claim, summarize evidence, or highlight reasoning steps, helping users and fact-checkers interpret model decisions~\cite{zeng2024justilm, schlichtkrull2023averitec}. They can be evaluated independently of the primary classification or regression task using human annotations or automatic metrics, making explanation generation a distinct methodological component~\cite{pham-etal-2025-claimpkg,vladika-etal-2025-step}. In practice, explanation generation is often combined with structured predictions, such as labels, scores, or relevance estimations, but its primary purpose is to provide transparency, improve interpretability, and support human-AI collaboration rather than directly improve predictive performance~\cite{zhao-etal-2024-findver,10.1145/3726302.3729931}.

Explanations connect previous tasks in the pipeline by synthesizing the outputs of claim normalization, claim decomposition, evidence retrieval, and claim verification. For example, explanations grounded in normalized claims tend to be more faithful than explanations generated from unconstrained, free-form justifications~\cite{schlichtkrull2023averitec}. Some systems require explanations to cite evidence snippets or follow a predefined structure to reduce hallucinations~\cite{zhao-etal-2024-findver, 10.1145/3726302.3729931}. 

The main benefit of explanation generation is increased transparency and interpretability of model outputs~\cite{pham-etal-2025-claimpkg}. Explanations help users to interpret the model's decision and provide fact-checkers with explicit cues about which parts of the input were considered important~\cite{vladika-etal-2025-step, 10.1145/3726302.3730142}. They can also surface internal inconsistencies and highlight missing or insufficient evidence, which can support error analysis and correction. Additionally, high-quality explanations can increase user trust, guide downstream components such as retrieval, and support more effective human-AI collaboration in verification workflows.

Despite the benefits, explanation generation has several limitations. LLM-produced explanations are not guaranteed to be faithful and may rationalize incorrect outputs or introduce hallucinated evidence. Maintaining consistency across languages, inputs, and model variants remains challenging, and producing explanations increases computational cost. A major challenge is evaluation: reliable assessment typically requires human-written explanations, which are costly and difficult to collect at scale. Automatic evaluation metrics primarily operate at the lexical level and often fail to capture semantic accuracy, completeness, or grounding quality, making robust evaluation particularly challenging. As a result, human annotations remain the most reliable evaluation approach, despite its high cost and limited scalability.

% Techniques and implementation
Explanation and justification generation techniques are closely related to those used for classification and regression tasks. The most common techniques leverage LLMs' generative capabilities to produce both the final decision (class or score) and accompanying rationale. Prompting techniques, including zero-shot and few-shot, remain fundamental approaches for generating explanations. In \textbf{zero-shot prompting}, LLMs are often instructed to produce a veracity label followed by a justification, relying on their pre-trained knowledge and given evidence to provide the reasoning~\cite{li2023revisitfakenewsdataset,10.1145/3589335.3651521,althabiti2024takeed,setty2024endtoend,hsu-etal-2023-explanation,liu2024large,purbey-etal-2025-1,dmonte-etal-2025-gmu,kawamura-2025-team,pham-etal-2025-claimpkg,10.1145/3726302.3729931,10.1145/3717867.3717896,10.1145/3731120.3744581,10.1145/3654777.3676359, pan2023qacheck}. Many studies adopt this strategy, prompting the model to classify a claim and then explain the verdict, or to generate an explanation alone with the veracity implied through the explanation rather than stated explicitly~\cite{10.1007/978-3-031-47896-3_1, russo-etal-2023-benchmarking, 10.1145/3722570.3726897}. These prompts often specify an explicit role for the model~\cite{abburi-etal-2025-deloitte, russo-etal-2025-euroverdict, 10.1145/3726302.3729931, 10.1145/3731120.3744581}, such as instructing the model to act as an expert fact-checker or domain-specialist, and may also require the output to be concise, cite evidence, or follow a JSON schema~\cite{ma2023exfever, 10.1145/3731120.3744581}. \textbf{Few-shot prompting} introduces demonstration pairs where models observe examples of claims, evidence, and corresponding explanations, which helps enforce stylistic consistency and discourages overly verbose or speculative reasoning~\cite{wang2023explainable,pan-etal-2023-fact,zeng2024justilm,schlichtkrull2023averitec,duesterwald2025cornellnlp,10.1145/3626772.3657874}. Several studies further require explanations for each decomposed subclaim, encouraging the model to frame the justification as a synthesis of smaller, individually assessed components~\cite{wang2023explainable}.

Beyond standard prompting, advanced prompting strategies increasingly rely on explicit reasoning. In \textbf{CoT prompting}, the LLM is asked to produce intermediate reasoning steps before producing the explanation, while these reasoning steps typically involve identifying the claim's key assertions, aligning them with retrieved evidence, and evaluating any contradictions~\cite{wang2023explainable,pelrine2023towards,zhao-etal-2024-findver,dougrez-lewis-etal-2025-assessing}. Some works introduce \textbf{self-ask} formulations, where the model recursively generates sub-questions about the claim and answers them before producing the explanation~\cite{wang2023explainable}. Other studies used contrastive prompts to require the LLM to provide both supporting and refuting explanations~\cite{si-etal-2024-large}. \textbf{Deductive} and \textbf{abductive reasoning prompting} have also been explored, which explicitly guides the type of inference required~\cite{dougrez-lewis-etal-2025-assessing}. These methods aim to reduce hallucinations, improve faithfulness, and better align the explanation with the model's internal reasoning. Some studies also explored the order of generation, showing that producing the explanation before the label can lead to more accurate predictions by forcing the model to reason more explicitly~\cite{vergho2024comparing}.

\textbf{Fine-tuning} represents another important approach for explanation generation~\cite{bhatia2021automaticclaimreviewclimate,10.1007/978-3-031-47896-3_1,kanaani-etal-2024-triple}. However, the scarcity of datasets containing ground-truth explanations makes it difficult to use this technique. Here, models are commonly trained to jointly predict the veracity label or score and the associated justification, often on datasets where human annotators provide short rationales or fact-checking summaries~\cite{sahnan2025llmsautomatefactcheckingarticle}. Such methods can take the form of training on decomposed claims and their explanations, or supervising the generation of flawed, focused rationales that explicitly highlight incorrect assumptions, missing context, or logical fallacies in false claims~\cite{10.1145/3589335.3651521}. Other works fine-tune models to generate extractive or abstractive explanations derived from fact-checking articles~\cite{russo-etal-2023-benchmarking}, or to produce counterfactual explanations assessing user reactions to generated rationales~\cite{hsu-etal-2023-explanation}. Additional studies explored domain-specific training for specialized fact-checking settings~\cite{10.1145/3722570.3726897,10.1145/3701716.3715599,purbey-etal-2025-1,dmonte-etal-2025-gmu,kawamura-2025-team}. Fine-tuned models are generally more consistent than prompted ones, but their performance depends strongly on the availability and quality of annotated rationales.

In addition to prompting and fine-tuning, many approaches rely on \textbf{retrieval-augmented generation} to improve grounding. Several studies have generated explanations directly from retrieved evidence snippets, either as summaries or contextualized rationales~\cite{setty2024endtoend, si-etal-2024-large, abburi-etal-2025-deloitte}. Some studies compare long-context models with retrieval-augmented generation for explanation generation, showing that retrieval can improve grounding but may increase sensitivity to irrelevant context~\cite{zhao-etal-2024-findver}. Others design multi-step workflows that require explanations to cite specific evidence passages, sometimes enforced by JSON schema or output constraints~\cite{zarharan-etal-2024-tell,10.1145/3731120.3744581}. Retrieval- and context-augmented prompting also includes the generation of community-note–style explanations~\cite{10.1145/3696410.3714934}, reasoning over user comments~\cite{liu2024large}, and role-conditioned verdict explanations that adhere to specific rules or stylistic constraints~\cite{russo-etal-2025-euroverdict}. These methods aim to improve factual grounding and reduce hallucinations by tying the explanation closely to the retrieved or structured input context.

\paragraph{Other.}

Beyond the specific generation approaches discussed above, several studies use LLMs for more general forms of unstructured text generation that do not fit cleanly into categories such as claim, explanation, or summary generation. These methods rely on free-form generation to transform, restructure, or synthesize textual content within fact-checking pipelines.

This flexible use of LLMs allows models to introduce new phrasing, reorganize information, or generate intermediate representations, but it also increases the risk of semantic drift, hallucination, and loss of factual fidelity, which are particularly critical in the verification process. One important application of this flexibility is the generation of textual evidence that supports or contradicts a claim. In this setting, models produce coherent sentences or paragraphs that highlight relevant information from retrieved documents or internal knowledge, helping human fact-checkers assess judgments, reduce noise in multi-step pipelines, and provide focused context for downstream modules such as entailment or summarization. Ensuring that this generated content remains faithful to the sources is a key challenge, as models may omit important details, introduce unsupported information, or produce irrelevant content~\cite{ko-etal-2023-claimdiff,fukuoka2025ksu,ko-etal-2023-claimdiff}.

More broadly, unstructured text generation has been applied for a variety of auxiliary purposes. \textit{Claim extraction} frames the identification of check-worthy factual statements as a sequence-to-sequence generation task, directly generating claims from input text without span section~\cite{schuster-etal-2021-get}. \textit{Claim rewriting and decontextualization} reformulate ambiguous or context-dependent statements into self-contained claims suitable for verification~\cite{deng-etal-2024-document}. Other studies reformulate fact-checking subtasks as questions, prompting models to generate or select relevant textual snippets for claim detection or claim-object identification~\cite{kotitsas-etal-2024-leveraging}. Text generation also supports retrieval through \textit{hypothetical document generation}, where models produce anticipated answers that are then embedded to retrieve relevant evidence~\cite{gao-etal-2023-precise}, or translate claims into \textit{program-like representations} that decouple natural language generation from structured reasoning~\cite{pan-etal-2023-fact}. Finally, LLMs have been used to \textit{simulate user-generated content}, such as role-conditioned comments based on predefined demographic profiles, providing auxiliary signals for misinformation analysis rather than direct claim verification~\cite{10.1145/3627673.3679519}.

\subsection{Evaluation Using LLMs}
\label{sec:evaluation}

\begin{figure*}
\centering
\resizebox{\textwidth}{!}{
\begin{forest}
			forked edges,
			for tree={
				grow=east,
				reversed=true,
				anchor=base west,
				parent anchor=east,
				child anchor=west,
                node options={align=center},
                align = center,
				base=left,
				font=\small,
				rectangle,
				draw=hidden-draw,
				rounded corners,
				minimum width=4em,
				edge+={darkgray, line width=1pt},
				s sep=3pt,
				inner xsep=2pt,
				inner ysep=3pt,
				ver/.style={rotate=90, child anchor=north, parent anchor=south, anchor=center},
			},
			where level=1{text width=5.0em,font=\scriptsize}{},
			where level=2{text width=5.6em,font=\scriptsize}{},
			where level=3{text width=6.8em,font=\scriptsize}{},
			[
			Evaluation,text width=7.5em%, ver
            [
                Rubric-Based, text width=7.5em
                [
                    {\citet{sun2024trustllm},}
                    {\citet{10.1145/3589335.3651521},}
                    {\citet{kim2024llmsproducefaithfulexplanations},}
                    {\citet{10.1145/3699682.3728349},}
                    {\citet{sahnan2025llmsautomatefactcheckingarticle}}
        			, leaf, align = left, text width=34.4em
                ]
			]
            [
                Factuality-Based, text width=7.5em
                [
                    {\citet{hu-etal-2025-decomposition},}
                    {\citet{li-etal-2024-self},}
                    {\citet{metropolitansky-larson-2025-towards},}
                    {\citet{wang-etal-2024-factcheck}}
        			, leaf, align = left, text width=34.4em
                ]
			]
			]
		\end{forest}
}
\caption{Taxonomy of evaluation approaches using generative LLMs in fact-checking, grouped into LLM-as-a-judge and factuality evaluation, with representative studies for each category.}
\label{fig:evaluation-taxonomy}
\end{figure*}

Traditional evaluation of fact-checking systems relies on manually annotated datasets and automatic text-similarity metrics, such as ROUGE or BERTScore. While effective for short outputs, these metrics often fail to capture reasoning quality, factual accuracy, coherence, or alignment with evidence when models generate long-form explanations or full fact-checking articles. In response, recent research increasingly uses LLMs themselves as evaluators. In these approaches, the LLM serves not as a data generator or fact-checking agent; instead, it assesses the outputs of other systems along dimensions such as correctness, persuasiveness, coherence, or factuality~\cite{kim2024llmsproducefaithfulexplanations, 10.1145/3699682.3728349, li-etal-2024-self, wang-etal-2024-factcheck}. 

LLM-based evaluation offers several advantages. First, it enables scalable, fine-grained assessment guided by human-like criteria without the cost of full expert annotation~\cite{sahnan2025llmsautomatefactcheckingarticle}. Second, LLMs can interpret lengthy outputs, detect subtle reasoning errors, and assess alignment with evidence in ways the traditional string-based metrics often miss~\cite{kim2024llmsproducefaithfulexplanations}. Third, LLM evaluators can operate at different granularities, from sentence-level evaluation to document-level quality judgments, making them particularly useful for explanation-based fact-checking systems or generated articles~\cite{sahnan2025llmsautomatefactcheckingarticle}.

Nevertheless, using LLMs as evaluators raises challenges. LLMs may overestimate the quality of outputs from models with overlapping training data, exhibit bias, or produce inconsistent judgments depending on prompts, rubrics, or reasoning aids, such as error typologies. Moreover, LLM evaluators themselves may hallucinate, misinterpret evidence, or produce inconsistent verdicts across runs. To mitigate these issues, studies often validate LLM evaluation against human annotations~\cite{kim2024llmsproducefaithfulexplanations, sahnan2025llmsautomatefactcheckingarticle} or employ structured rubrics~\cite{sahnan2025llmsautomatefactcheckingarticle}, reasoning constraints~\cite{kim2024llmsproducefaithfulexplanations}, or multi-stage decomposition~\cite{li-etal-2024-self,wang-etal-2024-factcheck,metropolitansky-larson-2025-towards,hu-etal-2025-decomposition}. In this context, LLM-based evaluation is best understood as a complementary evaluation, highly scalable and sensitive to deep semantic properties, but requiring careful calibration, cross-validation, and transparency.

We categorize LLM-based evaluation into two complementary approaches: \textbf{rubric-based and qualitative evaluation} and \textbf{factuality-based evaluation}, as illustrated in Figure~\ref{fig:evaluation-taxonomy}.

% LLM-as-a-judge is primarily applied to evaluate generative tasks, including explanation correctness~\cite{kim2024llmsproducefaithfulexplanations,10.1145/3589335.3651521}, argument coherence and persuasiveness~\cite{10.1145/3699682.3728349}, and the quality of automatically generated fact-checking articles~\cite{sahnan2025llmsautomatefactcheckingarticle}. In contrast, factuality evaluation pipelines rely more heavily on modular decomposition, claim extraction, evidence retrieval, and verification~\cite{li-etal-2024-self,wang-etal-2024-factcheck,metropolitansky-larson-2025-towards,hu-etal-2025-decomposition}, where LLMs play the role of component-wise assessors. Across both LLM-based evaluation approaches, critical considerations include evaluator calibration, reproducibility, sensitivity to error types, and alignment with human judgments.

\paragraph{Rubric-Based and Qualitative Evaluation.}

Rubric-based evaluation uses LLMs to assess the overall quality of generated outputs according to structured criteria, without decomposing the content into individually verifiable claims~\cite{GU2026101253}. Evaluation may include reasoning coherence, clarity, persuasiveness, and alignment with evidence.

Structured rubrics define specific assessment criteria and scoring standards. For instance, \citet{kim2024llmsproducefaithfulexplanations} use the G-Eval framework~\cite{liu-etal-2023-g} to measure explanation faithfulness at both sentence and document levels, with error-topology prompts to guide judgment, reducing ambiguity and aligning LLM judgments with human evaluation. \citet{sahnan2025llmsautomatefactcheckingarticle} design rubrics covering relevance, comprehensibility, importance, and evidence presentation. The evaluator model applies these criteria on a Likert scale, enabling more nuanced distinctions across generated fact-checking articles than generic automatic metrics would allow. Such rubrics enable fine-grained distinction across outputs, identification of weaknesses, and support for ablation studies. %However, the results depend heavily on rubric design, prompt quality and model calibration. Therefore, such evaluation is commonly followed by human annotation of a subset of the data to assess inter-annotator agreement (IAA) between LLM-based evaluation and human annotations.

Correctness-oriented evaluation asks the LLM to assess how well outputs satisfy task requirements or cover the necessary reasoning steps. This includes evaluating the completeness, coherence, or persuasiveness of explanations and responses. For example, \citet{10.1145/3699682.3728349} evaluate how convincing a generated response to a misinformation claim is, explicitly leveraging factuality, while \citet{10.1145/3589335.3651521} assign continuous scores to justifications for completeness and reasoning quality, serving as a complementary metric to lexical or embedding-based baselines. On the other hand, \citet{sun2024trustllm} use the LLM to determine whether model responses successfully corrected misinformation injected into user prompts, demonstrating the applicability of LLM judges to adversarial or safety-related scenarios. These judgments complement traditional automatic metrics by capturing semantic dimensions of output quality.

Across rubric-based evaluations, LLMs judgments correlate well with human assessments when grounded in carefully designed prompts and rubrics~\cite{sahnan2025llmsautomatefactcheckingarticle}. Nonetheless, calibration remains crucial, as evaluations can be sensitive to prompt phrasing, model type, and ensemble configuration. Multi-judge approaches, cross-validation, or meta-evaluation techniques are commonly used to address these issues.

\paragraph{Factuality-Based Evaluation.}

Factuality-based evaluation focuses on systematically assessing the factual accuracy and verifiability of the content generated by an LLM, rather than generating predictions or labels. Unlike fact-checking prediction, where LLMs actively produce verdicts such as support, refute, or neutral for a claim, factuality evaluation examines the model's outputs to determine whether each statement is correct, verifiable, and grounded in evidence~\cite{wang-etal-2024-factuality}. This distinction is critical: factuality evaluation is applied post hoc to generated content to measure reliability, while fact-checking prediction constitutes an end-to-end system that generate the verdict itself.

Factuality evaluation typically relies on multi-stage pipelines, including claim extraction, evidence retrieval, and claim-level verification, to ensure that each component of the generated content can be rigorously evaluated. A crucial step is decomposing long-form outputs into atomic, verifiable units, which allows for precise assessment of each individual claim or statement~\cite{li-etal-2024-self, wang-etal-2024-factcheck}. This decomposition reduces ambiguity, facilitates evidence-grounded analysis, and enables fine-grained analysis of model errors. Systems such as \textit{Self-Checker}~\cite{li-etal-2024-self} extract claims, generate search queries, select evidence, and predict claim-level support/refute. Subsequent pipelines introduce de-contextualization, check-worthiness detection, stance classification, and correctness-preserving editing~\cite{wang-etal-2024-factcheck}, demonstrating the benefits of modular evaluation.

Recent work focuses on improving the accuracy and robustness of extraction and verification. For example, \citet{metropolitansky-larson-2025-towards} proposed a preliminary filtering step in which the model identifies verifiable sentences and rewrites ambiguous or referentially unclear segments. By ensuring that only unambiguous content enters the evaluation pipeline, the system reduces the propagation of errors. Similarly, \citet{hu-etal-2025-decomposition} introduced reflection-based refinement, where an additional LLM identifies decomposition errors, such as omitted claims or over-splitting, and suggests corrections before evaluation. These methods significantly enhance alignment between extracted claims and the content that requires verification.

Through the factuality-evaluation studies, several methodological patterns emerge. First, decomposition acknowledges that errors in generative models often arise from subtle or compound factual inaccuracies that cannot be captured at the document level~\cite{wang-etal-2024-factcheck, hu-etal-2025-decomposition}. Second, the integration with retrieval-augmented verification grounds the assessment in external evidence, mitigating hallucination risks. However, the crucial aspect of the factuality evaluation is the credibility of the sources used, as employing disinformation sources for factual evaluation introduces additional issues~\cite{vykopal2025assessingwebsearchcredibility}. Third, computing global factuality scores (e.g., FactScore~\cite{min-etal-2023-factscore}) by aggregating claim-level verdicts enables interpretable comparisons across systems and fine-grained profiling of weaknesses. At the same time, factuality-based evaluation faces limitations, such as decomposition can distort meaning, evidence retrieval may miss relevant information, and LLM-based stance detection can be sensitive to phrasing and prompt style. Furthermore, these pipelines are computationally intensive, often requiring multiple sequential passes through an LLM, making them costly to scale.

Overall, factuality-based evaluation represents a shift toward structured, evidence-grounded assessment of generative outputs. It complements rubric-based evaluation by focusing not on perceived quality or persuasiveness but on whether the output's factual claims are true, verifiable, and well-supported. As generative fact-checking systems become increasingly capable of producing long-form explanations and news-style articles, these fine-grained factuality pipelines will be essential for ensuring reliability, transparency, and accountability.

\section{AI Pipelines and Agentic Systems}

\begin{figure*}
\centering
\resizebox{\textwidth}{!}{
\begin{forest}
			forked edges,
			for tree={
				grow=east,
				reversed=true,
				anchor=base west,
				parent anchor=east,
				child anchor=west,
                node options={align=center},
                align = center,
				base=left,
				font=\small,
				rectangle,
				draw=hidden-draw,
				rounded corners,
				minimum width=4em,
				edge+={darkgray, line width=1pt},
				s sep=3pt,
				inner xsep=2pt,
				inner ysep=3pt,
				ver/.style={rotate=90, child anchor=north, parent anchor=south, anchor=center},
			},
			where level=1{text width=5.0em,font=\scriptsize}{},
			where level=2{text width=5.6em,font=\scriptsize}{},
			where level=3{text width=6.8em,font=\scriptsize}{},
			[
			AI Pipelines and\\Agentic Systems, ver
            [
                Retrieval-Augmented\\Generation, text width=7.5em
                [
                    {\citet{irnawan-etal-2025-claim},}
                    {\citet{JMLR:v24:23-0037},}
                    {\citet{liu-etal-2025-raemollm},}
                    {\citet{10.1145/3699682.3728349},}
                    {\citet{yue-etal-2024-retrieval},}
                    {\citet{zeng2024justilm},}\\
                    {\citet{zhang-gao-2024-reinforcement},}
                    {\citet{zhao-etal-2024-findver}}
        			, leaf, align = left, text width=34.4em
                ]
			]
            [
                Pipelines, text width=7.5em
                [
                    {\citet{kao-yen-2024-magic},}
                    {\citet{leippold2024automated},}
                    {\citet{li-etal-2025-imrrf},}
                    {\citet{liu-etal-2024-teller},}
                    {\citet{10.1145/3705754.3705757},}
                    {\citet{wang2023explainable},}\\
                    {\citet{zhang2023llmbased},}
                    {\citet{10.1145/3643562.3672613}}
        			, leaf, align = left, text width=34.4em
                ]
			]
            [
                Agentic Systems, text width=7.5em
                [
                    {\citet{kim2024llmsproducefaithfulexplanations},}
                    {\citet{10.1145/3726302.3730092},}
                    {\citet{10.1145/3696410.3714748},}
                    {\citet{sahnan2025llmsautomatefactcheckingarticle},}
                    {\citet{xie-etal-2025-fire},}
                    {\citet{zhao-etal-2024-pacar}}
        			, leaf, align = left, text width=34.4em
                ]
			]
			]
		\end{forest}
}
\caption{Taxonomy of the AI pipelines and agentic systems used for fact-checking, grouped based on used paradigm.}
\label{fig:pipeline-taxonomy}
\end{figure*}

Automated fact-checking with generative LLMs is evolving from simple, end-to-end prompting to more structured, flexible system designs. Early approaches typically used direct prompting of LLM, asking them to verify a claim in a single step. While straightforward, this often led to inconsistent results, unsupported statements, or hallucinations. To address these limitations, recent research has focused on AI pipelines and agentic systems, which organize LLMs into multi-stage frameworks that combine retrieval, reasoning, iterative refinement, and verification. These approaches enable more reliable, interpretable, and adaptive fact-checking, as each stage explicitly handles a distinct aspect of the verification process, such as gathering evidence, analyzing its relevance, or checking factual consistency~\cite{rahman2025hallucinationtruthreviewfactchecking}.

In the literature, these approaches can be categorized into two paradigms. \textit{Pipeline architectures} follow a predefined sequence of steps, where each component, such as claim decomposition, evidence retrieval, reasoning, and verification, performs a specific role within the workflow. In contrast, \textit{agentic systems} treat LLMs as autonomous or semi-autonomous agents capable of planning, collaboration, or using external tools to accomplish complex fact-checking tasks~\cite{zhao-etal-2024-pacar, li-etal-2025-imrrf, wang2023explainable}. Both paradigms rely on \textit{retrieval-augmented generation}~\cite{10.5555/3495724.3496517, zhang-gao-2024-reinforcement, zeng2024justilm}, which grounds LLM outputs in external evidence and mitigates hallucinations. The taxonomy of this section is shown in Figure~\ref{fig:pipeline-taxonomy}. Across these systems, a typical workflow emerges: claims are first decomposed, evidence is retrieved and refined, reasoning modules evaluate the information, and evaluators ensure the final output is accurate, coherent, and faithful. This structured approach enables more interpretable and reliable fact-checking compared to single-step prompting.

\subsection{Retrieval-Augmented Generation}

\textit{Retrieval-Augmented Generation (RAG)} has emerged as an essential paradigm for integrating generative LLMs into automated fact-checking~\cite{zeng2024justilm}. The core idea is to ground outputs in external evidence by retrieving relevant documents from structured sources and conditioning the generation on this retrieved context. In fact-checking, RAG is particularly beneficial because it directly addresses two main challenges of generative models: hallucination and lack of verifiable attribution.

\paragraph{General RAG Approaches for Fact-Checking.}

Most RAG-based fact-checking systems follow a retrieve-then-generate approach, although the degree of interaction between retrieval and generation varies. Early systems, such as \textit{Atlas}~\cite{JMLR:v24:23-0037}, tightly integrate dense retrieval with generation by jointly training the retriever and the language model. \textit{Atlas} encodes each retrieved document independently and concatenates the resulting representations, allowing the decoder to attend over multiple evidence sources. Crucially, fine-tuning the retriever is shown to be essential, as frozen retrievers lead to substantial performance degradation. 

Related work focused on fine-tuning the retriever without updating the generator. \textit{REPLUG}~\cite{zhang-gao-2024-reinforcement}, for example, connects dense retrieval with a frozen LLM by letting the model itself assess the relevance of retrieved documents. After retrieving a small candidate set using a dense retriever, each document is prepended separately to the input claim. The language model processes these augmented inputs in parallel, and token-level predictions are ensembled to produce the final output. The authors aimed to improve retrieval quality by fine-tuning the retriever while using LLMs' relevance signals for supervision.

\paragraph{RAG for Explainable Fact-Checking.}

RAG has been widely adopted for generating explanations and justifications, as well as for predicting veracity. \textit{JustiLM}~\cite{zeng2024justilm} retrieves evidence chunks based on the input claim and conditions the generator on this evidence to produce natural language justifications. To improve faithfulness, the authors introduce distillation objectives that align retrieved documents with full fact-checking articles available during training. \textit{FinDVer}~\cite{zhao-etal-2024-findver} follows a similar retrieval-based design for financial documents, highlighting the importance of evidence selection and ordering when verifying claims over long or structured content.

\paragraph{Multi-Stage RAG Systems.}

Beyond simple retrieval, some systems introduce multi-stage RAG pipelines to structure the verification process. \textit{RAFTS}~\cite{yue-etal-2024-retrieval} combines demonstration retrieval, document retrieval, and contrastive argument generation, providing the LLM with both supporting and refuting evidence before making a final prediction. In addition, \textit{RARG}~\cite{yue-etal-2024-evidence} extends this idea by incorporating reinforcement learning from human feedback, training the system to generate responses that are not only factually aligned with the retrieved evidence but also relevant and capable of explicitly addressing counter-evidence. Similarly, \citet{abdallah2025ngu_research} propose a four-stage pipeline for numeric claims by reformulating each claim into a question, retrieving relevant passages, and producing a concise answer that captures the numeric gist. These approaches show how multi-stage pipelines improve precision and evidence alignment for complex claims.

\paragraph{Query Reformulation and Personalization.}

Other approaches focused on improving retrieval quality by reformulating queries. \citet{irnawan-etal-2025-claim} use an LLM to generate and rank natural language queries derived from the claim before retrieving evidence for verification. Related strategies appear in knowledge-graph-based fact-checking~\cite{10.1145/3726302.3730142}, where multiple reformulated queries are used to retrieve and rank evidence from the web. RAG has also been extended to personalized fact-checking, where retrieved documents are filtered based on user trust preferences before being summarized and rephrased by an LLM~\cite{10.1145/3699682.3728349}.

\paragraph{Discussion.}

RAG offers several advantages for automated fact-checking by grounding generation in external evidence, reducing hallucinations, and increasing transparency through evidence citation. Moreover, modular retrieval components allow systems to be adapted with new information without the necessity to fine-tune the LLM. However, RAG also introduces challenges, such as retrieval errors that can propagate downstream, leading to confidently generated but incorrect conclusions grounded in irrelevant or incorrect evidence. In addition, chunking strategies, evidence ordering, and context window limitation can distort the original meaning of source documents, especially in long-form settings. Moreover, many systems assume the availability of high-quality, trustworthy corpora, and an assumption that may not hold in low-resource languages or emerging misinformation domains.

\subsection{Pipeline Architectures}

Pipeline architectures organize the fact-checking pipeline as a sequence of predefined steps, such as claim decomposition, evidence retrieval, and veracity prediction. Their main advantages are modularity, interpretability, and controllability, which allow researchers to inspect intermediate outputs, isolate errors, and integrate domain knowledge or external tools systematically.

\paragraph{Deterministic and Multi-Stage Pipelines.}

Early pipeline-based approaches emphasize the explicit decomposition of claims to simplify verification. Approaches such as \textit{FOLK}~\cite{wang2023explainable} and \textit{HiSS}~\cite{zhang2023llmbased} decompose complex claims into simpler sub-claims or questions that can be verified independently. \textit{FOLK} translates claims into first-order logic clauses, which are evaluated through targeted question-answering over web search results. \textit{HiSS} similarly decomposes claims and evaluates sub-claims through a confidence-driven mechanism, in which the model decides whether it can answer a question directly or requires evidence retrieval. These designs reduce hallucination and improve traceability, but they rely heavily on the quality of the initial decomposition, which can limit performance on implicit or nuanced claims. 

Several pipelines introduce iterative refinement to improve faithfulness. \textit{MAGIC}~\cite{kao-yen-2024-magic} incorporates self-refinement loops that verify alignment between generated arguments and retrieved evidence, discarding or revising unsupported reasoning. In contrast, \textit{TTFact}~\cite{10.1145/3705754.3705757} simplifies inference by leveraging task and template transformations, exploiting known limitations of LLMs in handling multi-label or Not Enough Information (NEI) cases. While such pipelines improve robustness, they increase complexity and require careful calibration to avoid propagation of errors.

\paragraph{Knowledge Augmented Pipelines.}

Beyond text retrieval, some pipelines integrate diverse knowledge sources. \textit{IMRRF}~\cite{li-etal-2025-imrrf} combines corpus-based retrieval, knowledge-graph reasoning, and the LLM's internal knowledge. It further applies summarization and redundancy filtering to reduce noise before verification. This design enhances coverage for multi-hop claims but raises concerns about information leakage from the LLM's parametric knowledge, making it difficult to attribute evidence sources and assess verification faithfulness. 

Hybrid pipelines further extend this paradigm by introducing human oversight. \textit{TriIntel}~\cite{10.1145/3643562.3672613} routes claims through deep neural networks (DNNs), LLMs, and human annotators based on uncertainty estimates. Such systems are particularly effective for addressing ambiguous or novel misinformation. However, they are costly and difficult to scale, highlighting the trade-off between accuracy and practical deployment.

\paragraph{Pipelines for Logical Reasoning.}

Some pipelines explicitly focus on improving logical reasoning. \textit{TELLER}~\cite{liu-etal-2024-teller} decomposes claims into structured yes/no predicates and aggregates the resulting truth values into a final veracity prediction. This produces interpretable decision rules and improves consistency, but it assumes that claims can be captured by predefined logical templates, which may limit coverage for open-domain claims or rhetorically framed misinformation that does not conform to explicit logical forms.

Overall, pipeline architectures are beneficial for transparency and error diagnosis, but tend to follow fixed execution flows. Errors introduced in early stages, such as decomposition or retrieval, often propagate downstream, motivating the development of more adaptive alternatives.

\subsection{Agentic Systems}

An agentic system approach for fact-checking involves the interaction of multiple LLM-based agents that plan, reason, collaborate, or debate. Unlike pipelines, which follow a fixed flow, agentic systems dynamically adapt execution through iterative revision and task specialization, making them well-suited for complex, multi-hop, or ambiguous claims.

\paragraph{Multi-Agent Collaboration.}

Several systems distribute the verification process across specialized agents. For example, \textit{LoCal}~\cite{10.1145/3696410.3714748} assigns different agents to decomposition, reasoning, logical evaluation, and counterfactual checking, iterating until sufficient confidence is reached. On the other hand, \textit{PACAR}~\cite{zhao-etal-2024-pacar} adopts a planner-executor paradigm, dynamically selecting specialized agents for tasks such as numerical reasoning, entity disambiguation, or general inference. Such specialization improves performance on structurally complex claims, but coordinating multiple agents introduces overhead and increases sensitivity to planning errors.

\paragraph{Debate- and Critique-Based Agents.}

Another class of agentic systems relies on adversarial or collaborative debate. \textit{TruEDebate}~\citep{10.1145/3726302.3730092} simulates structured debates between proponents and opponents, whose arguments are later synthesized into a final verdict. Similarly, \textit{MADR}~\cite{kim2024llmsproducefaithfulexplanations} employs multiple LLM agents as debaters to iteratively identify reasoning errors, debate discrepancies, and refine explanations. After the debate, a separate agent reconciles feedback from both debaters and triggers further iterations if disagreement remains, before refining the final explanation. These approaches encourage multi-perspective reasoning and expose hidden assumptions, but they increase inference cost and risk reinforcing errors when agents prematurely converge on flawed reasoning.

\paragraph{Planning and Reasoning.}

Planner-driven agentic designs blur the boundary between pipelines and agents by introducing conditional execution within otherwise structured workflows. \textit{PACAR}~\cite{zhao-etal-2024-pacar} and \textit{FIRE}~\cite{xie-etal-2025-fire} dynamically decide whether to retrieve evidence, continue reasoning, or terminate based on confidence estimates. This conditional execution reduces unnecessary retrieval and enables adaptive depth of reasoning. However, confidence estimation itself is prone to errors, and early or delayed stopping decisions can degrade verification accuracy.

\subsection{Unifying Patterns and Design Trade-Offs}

Across both pipelines and agentic systems, common patterns emerge. Claim decomposition is a universal concept that serves as the backbone of most systems, enabling fine-grained verification but making it sensitive to the quality of the decomposition. Retrieval and evidence refinement evolve from static search to interactive or reinforcement-optimized strategies~\cite{zhang-gao-2024-reinforcement}. Iterative reasoning and evaluation, including self-reflection, debate, alignment checking, and counterfactual reasoning, are increasingly used to improve faithfulness and robustness. 

From a system design perspective, pipelines and agentic systems embody distinct trade-offs in automated fact-checking. Pipeline-based approaches prioritize stability, interpretability, and predictable execution, which are often critical in high-stakes settings. In contrast, agentic systems emphasize adaptivity and flexible multi-step reasoning, enabling deeper exploration of complex or ambiguous claims at the cost of increased system complexity. These differences underscore that architectural choices in fact-checking systems are primarily driven by application requirements, resource constraints, and tolerance for uncertainty.

\section{Multilingual and Cross-Lingual Fact-Checking}

\begin{figure*}
\centering
\resizebox{\textwidth}{!}{
\begin{forest}
			forked edges,
			for tree={
				grow=east,
				reversed=true,
				anchor=base west,
				parent anchor=east,
				child anchor=west,
                node options={align=center},
                align = center,
				base=left,
				font=\small,
				rectangle,
				draw=hidden-draw,
				rounded corners,
				minimum width=4em,
				edge+={darkgray, line width=1pt},
				s sep=3pt,
				inner xsep=2pt,
				inner ysep=3pt,
				ver/.style={rotate=90, child anchor=north, parent anchor=south, anchor=center},
			},
			where level=1{text width=5.0em,font=\scriptsize}{},
			where level=2{text width=5.6em,font=\scriptsize}{},
			where level=3{text width=6.8em,font=\scriptsize}{},
			[
			Multilingual Approaches, ver
            [
                English-only\\Prompting, text width=7.5em
                [
                    {\citet{aarnes2024iaigroupcheckthat2024},}
                    {\citet{abdallah2025ngu_research},}
                    {\citet{aly-etal-2023-qa},}
                    {\citet{amatya2025factiverse},}
                    {\citet{anikina2025dfkinit2b},}
                    {\citet{chung2025translationllmbaseddatageneration},}\\
                    {\citet{eren2024turquaz},}
                    {\citet{golik2024dshacker},}
                    {\citet{10.1145/3583780.3614936},}
                    {\citet{huang2024fakegptfakenewsgeneration},}
                    {\citet{hyben2023bigger},}
                    {\citet{kim-etal-2023-covid},}\\
                    {\citet{10411561},}
                    {\citet{10.1145/3726302.3730092},}
                    {\citet{mittal-etal-2023-lost},}
                    {\citet{pelrine2023towards},}
                    {\citet{pramov2025ds},}
                    {\citet{quelle2023perils},}\\
                    {\citet{russo-etal-2025-euroverdict},}
                    {\citet{saeed2025mma},}
                    {\citet{sawinski2025openfact},}
                    {\citet{setty2024endtoend},}
                    {\citet{shcharbakova-etal-2025-scale},}
                    {\citet{strong-etal-2024-zero},}\\
                    {\citet{vergho2024comparing},}
                    {\citet{vineetha2025saivineetha},}
                    {\citet{vykopal2025largelanguagemodelsmultilingual},}
                    {\citet{zeng-etal-2024-ru22fact}}
        			, leaf, align = left, text width=34.4em
                ]
			]
            [
                Target-Language\\Prompting, text width=7.5em
                [
                    {\citet{almada2025akcit},}
                    {\citet{anikina2025dfkinit2b},}
                    {\citet{cao2023large},}
                    {\citet{cekinel-etal-2024-cross},}
                    {\citet{gao-etal-2023-precise},}
                    {\citet{le2025lis},}\\
                    {\citet{russo-etal-2025-euroverdict},}
                    {\citet{scaiella-etal-2024-leveraging}}
        			, leaf, align = left, text width=34.4em
                ]
			]
            [
                Translation-Based\\Approach, text width=7.5em
                [
                    {\citet{aarnes2024iaigroupcheckthat2024},}
                    {\citet{cekinel-etal-2024-cross}}
                    {\citet{modzelewski2023dshacker},}
                    {\citet{quelle2023perils},}
                    {\citet{tran2022ur},}\\
                    {\citet{vykopal2025largelanguagemodelsmultilingual}}
        			, leaf, align = left, text width=34.4em
                ]
			]
            [
                Multi-Lingual\\Fine-Tuning, text width=7.5em
                [
                    {\citet{amatya2025factiverse},}
                    {\citet{anikina2025dfkinit2b},}
                    {\citet{beltran2025umuteam},}
                    {\citet{du2022nus},}
                    {\citet{10.1145/3643491.3660290},}
                    {\citet{russo-etal-2025-euroverdict},}\\
                    {\citet{saeed2025mma},}
                    {\citet{sawinski2025openfact},}
                    {\citet{schutz2022ait_fhstp}}
                    {\citet{shcharbakova-etal-2025-scale}}
        			, leaf, align = left, text width=34.4em
                ]
			]
            [
                Per-Language\\Fine-Tuning, text width=7.5em
                [
                    {\citet{anikina2025dfkinit2b},}
                    {\citet{beltran2025umuteam},}
                    {\citet{cekinel-etal-2024-cross},}
                    {\citet{hashmi2025investigators},}
                    {\citet{10.1145/3627673.3679519},}
                    {\citet{saeed2025mma},}\\
                    {\citet{scaiella-etal-2024-leveraging},}
                    {\citet{vineetha2025saivineetha}}
        			, leaf, align = left, text width=34.4em
                ]
			]
            [
                Cross-Lingual\\Transfer, text width=7.5em
                [
                    {\citet{agrestia2022polimi},}
                    {\citet{aly-etal-2023-qa},}
                    {\citet{hashmi2025investigators},}
                    {\citet{vykopal-etal-2025-soft}}
        			, leaf, align = left, text width=34.4em
                ]
			]
			]
		\end{forest}
}
\caption{Taxonomy of multilingual and cross-lingual approaches using LLMs. Approaches are grouped according to how linguistic diversity is handled, ranging from English-centric prompting and translation-based pipelines to multilingual fine-tuning and explicit cross-lingual transfer, with corresponding studies for each category.}
\label{fig:multilinguality-taxonomy}
\end{figure*}

\subsection{Language Distribution of Studies}

\begin{figure*}
    \centering
    \includegraphics[width=1\linewidth]{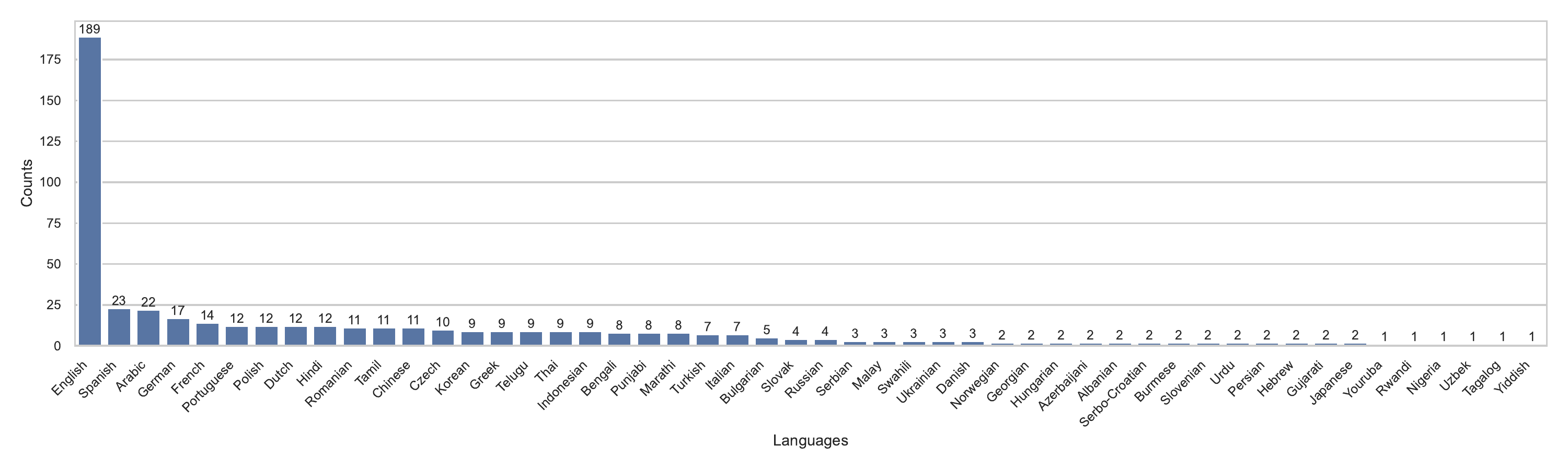}
    \caption{Top 50 most common languages in fact-checking papers. The bar represents the total number of papers that evaluated the language.}
    \label{fig:language_distribution}
\end{figure*}

\begin{figure*}
    \centering
    \includegraphics[width=1\linewidth]{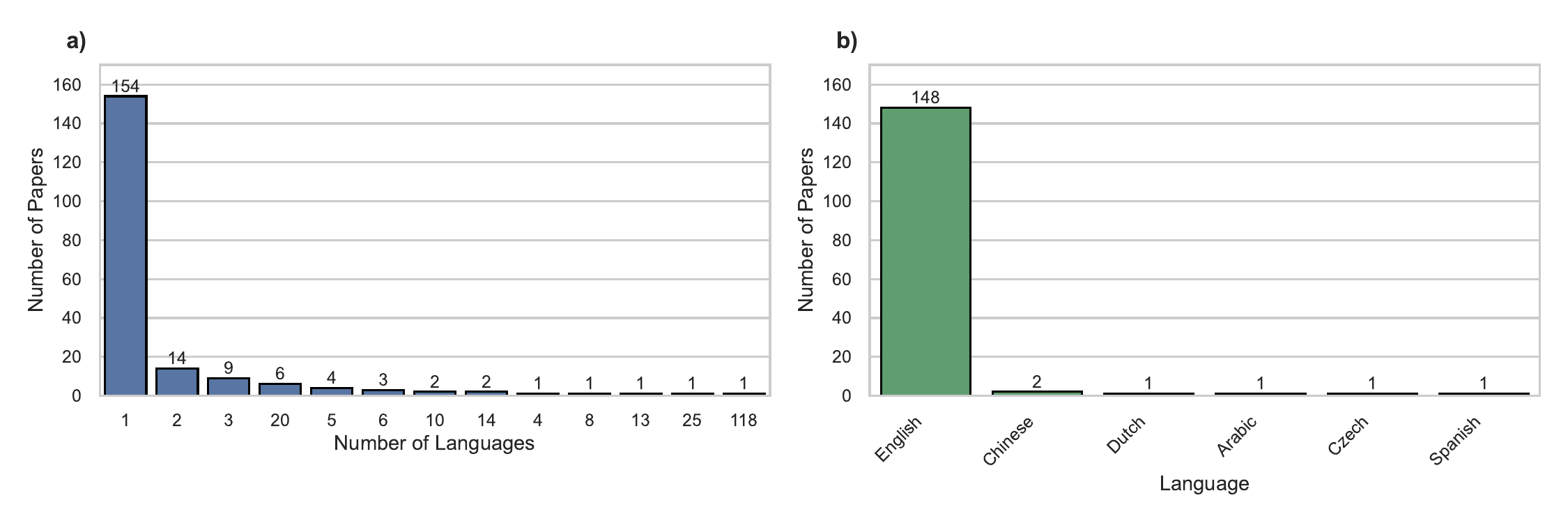}
    \caption{Distribution of languages in fact-checking papers. \textbf{(a) Number of languages studied per paper}. Each bar represents the number of papers that evaluated a given number of languages. \textbf{(b) Languages that appeared in papers evaluating only a single language.} Each bar represents the number of papers in which that language was the sole language studied.}
    \label{fig:count_distribution}
\end{figure*}

Despite the global nature of false information, research on LLM-based fact-checking remains disproportionately centered on English (see Figure~\ref{fig:language_distribution}). Out of the reviewed papers, 154 focused on a single language (see Figure~\ref{fig:count_distribution} a)), and the vast majority used only English datasets ($148\times$, see Figure~\ref{fig:count_distribution} b)). Only a few addressed other languages, such as Chinese~\cite{zhang2024need, 10.1145/3705754.3705757}, Dutch~\cite{weering2024fc_rug}, Spanish~\cite{acosta2025ucom_unam_pln}, Arabic~\cite{althabiti2024takeed}, or Czech~\cite{10.1145/3589335.3651914}, typically reflecting the availability of resources rather than a systematic multilingual design.

In contrast, 45 papers considered at least two languages, and 15 extended their experiments to more than ten (see Figure~\ref{fig:count_distribution} a)). The greatest multilingual efforts cover 118 languages~\cite{setty2024endtoend}, demonstrating both the ambition and scalability of synthetic datasets creation empowered by machine translation and LLM prompting. The list of all approaches for multilingual fact-checking is shown in Figure~\ref{fig:multilinguality-taxonomy}.

The prevalence of English-centric studies reflects persistent structural disparities in dataset availability, annotation resources, and model evaluation benchmarks. However, the growing body of multilingual work signals a shift toward more inclusive fact-checking systems that are resilient across typologically diverse languages and scripts. This methodological expansion also highlights that generative LLMs can serve as a unifying layer for multilingual processing, even where high-quality training data are scarce.

\subsection{Prompt Language vs Target-Language Strategies}

A central consideration in multilingual fact-checking is the distinction between \textbf{multilingual} and \textbf{cross-lingual} approaches. Multilingual systems are explicitly trained or fine-tuned on data from multiple languages, enabling them to process inputs in any of the covered languages without additional translation. In contrast, cross-lingual approaches aim to transfer knowledge from a high-resource language, typically English, to one or more low-resource target languages. These methods leverage shared representations, translation mechanisms, or prompting strategies to generalize fact-checking capabilities, even when labeled data in the target language is scarce. 

Many studies adopt English-only prompts, treating LLMs as inherently multilingual models capable of understanding non-English input~\cite{mittal-etal-2023-lost, strong-etal-2024-zero, zeng-etal-2024-ru22fact}. This approach benefits from the fact that instruction-tuned models tend to be optimized for English, often resulting in stronger task adherence, better reasoning chains, and more stable outputs. It also simplifies pipeline development by avoiding language-specific prompt engineering and translating instructions to specific languages.

However, English-centric prompting also has disadvantages. It may unintentionally encode linguistic bias, thereby amplifying performance disparities between high-resource European languages and those of low-resource languages. Moreover, models may misinterpret idiomatic or culturally specific content in the claim if they cannot robustly ground it in English semantic space.

Translation-based approaches balance these two extremes. Some studies translate input claims into English to ground reasoning in a well-supported language~\cite{quelle2023perils, tran2022ur}, while others instruct the LLM to internally translate during inference~\cite{vykopal2025largelanguagemodelsmultilingual} (e.g., cross-lingual though prompting~\cite{huang-etal-2023-languages}). These strategies provide the advantage of more reliable reasoning, but the disadvantage lies in the risk of semantic drift introduced by machine translation, especially problematic in misinformation, where subtle wording changes can flip veracity labels.

\subsection{Multilingual Fine-Tuning and Cross-Lingual Transfer}

Beyond prompting, \textbf{multilingual fine-tuning} is a core technique used to achieve generalization across languages. Jointly fine-tuning a single model on data from multiple languages allows shared representations to emerge, often improving performance for low-resource languages through knowledge transfer~\cite{du2022nus, russo-etal-2025-euroverdict}. The key advantage of this approach is its scalability, which enables one model to handle multiple languages without requiring per-language tuning. However, this same property can become a drawback when dominant high-resource languages dominate weaker ones, leading to suboptimal adaptation or potentially negative transfer.

\paragraph{Language-Specific Fine-Tuning.}

Language-specific fine-tuning via separate models or adapters~\cite{beltran2025umuteam, scaiella-etal-2024-leveraging} enhances specialization and tends to yield strong monolingual performance. This approach allows models to capture language-specific patterns and structural nuances that may be missed by multilingual fine-tuning. However, the disadvantage of language-specific fine-tuning is that it does not scale well to many languages and fails to leverage cross-lingual transfer, especially when training data are scarce. As a result, maintaining separate models for each language can be resource-intensive and may limit generalization to low-resource languages where annotated datasets are limited or unavailable.

\paragraph{Cross-Lingual Transfer.}

Cross-lingual transfer techniques attempt to unify the benefits of both paradigms. Soft language prompts, adapter-based architectures, and hybrid task-language prompting strategies~\cite{vykopal-etal-2025-soft, agrestia2022polimi} highlight how LLMs can be guided to share linguistic knowledge across high- and low-resource languages. Their advantage lies in flexibility, where task-specific or language-specific components can be mixed to optimize particular linguistic conditions. Yet, these techniques depend heavily on representations from the pre-trained LLM: transfer is consistently stronger between languages that share scripts or typological features, whereas transfer across distant languages (e.g., from English to Telugu) remains challenging.

\paragraph{Synthetic Multilingual Training Data.}

Synthetic multilingual training data, often created through LLM-based translation, paraphrases, or machine translation~\cite{setty2024endtoend, amatya2025factiverse}, offer a practical way to expand coverage. The advantage is rapid dataset scaling across dozens or hundreds of languages. The disadvantage is variable translation quality, which may introduce noise or propagate hallucination when used for training.

Zero-shot cross-lingual transfer remains promising because it requires no target-language labels~\cite{hashmi2025investigators, aly-etal-2023-qa}. While LLMs show surprising generalization ability in this setting, performance gaps persist, especially for typologically distant languages. Few-shot prompting using multilingual demonstrations can alleviate these gaps, but this depends on carefully selecting demonstrations and maintaining consistent instruction templates.

\section{Challenges and Future Directions}
 
Despite their rapid adoption in automated fact-checking pipelines, generative LLMs introduce challenges that differ fundamentally from those of traditional supervised or retrieval-based systems. These challenges arise from the generative nature of LLMs and their increasing use as autonomous or semi-autonomous decision-makers within complex fact-checking pipelines. This section focuses on the key challenges in using LLMs for fact-checking and outlines promising directions for future research.

\subsection{Hallucinations and Factual Faithfulness}

\paragraph{Challenges.}

One of the most critical challenges in applying LLMs to fact-checking is their tendency to hallucinate plausible but incorrect information. Because LLMs generate the answers autoregressively, they may produce statements that are not supported by the provided evidence or grounded in external sources~\cite{akhtar-etal-2023-multimodal, xie-etal-2025-fire, zhang2023llmbased, chen2023felm, guan2023language}. These hallucinations may be \textit{intrinsic}, when the model misinterprets evidence present in the context, or \textit{extrinsic}, when it fabricates unsupported facts or references~\cite{kim2024llmsproducefaithfulexplanations}. A particularly problematic case in fact-checking is \textit{verification hallucination}, where a model predicts the correct veracity label but supports it with incorrect reasoning or hallucinated evidence, making the output appear sound despite being unfaithful~\cite{wang2023explainable}. Empirical studies suggest that roughly 80~\% of zero-shot explanations contain hallucinated details, often fluent and persuasive enough that human reviewers can accept them uncritically~\cite{kim2024llmsproducefaithfulexplanations}. Numerical claims add another level of difficulty as LLMs frequently misinterpret quantities, dates, or arithmetic relationships even when correct values are present in the retrieved context~\cite{abdallah2025ngu_research}.

\paragraph{Future Directions.}

Retrieval-augmented generation and chain-of-though strategies that encourage models to justify predictions using retrieved documents represent steps towards grounding model reasoning in evidence~\cite{10.1145/3726302.3729931, dhuliawala-etal-2024-chain, 10.1145/3726302.3730142}. However, hallucinated reasoning remains common even when evidence is provided, indicating that surface-level access to documents is insufficient. More rigorous approaches, such as constrained decoding, explicit evidence attribution, or strategies in which models iteratively validate each reasoning step before committing to a verdict, remain underexplored and represent a promising avenue for future research. Developing models that can explicitly express uncertainty when evidence is insufficient also remains an important open problem.

\subsection{Multilinguality and Cross-Lingual Fact-Checking}

\paragraph{Challenges.}

LLM-based fact-checking systems remain strongly biased toward high-resource languages, creating a substantial performance gap with low-resource languages~\cite {vykopal2025largelanguagemodelsmultilingual, shcharbakova-etal-2025-scale, mirza2024globalliar}. Translating non-English claims into English before verification leads to improved performance~\cite{vykopal2025largelanguagemodelsmultilingual, quelle2023perils}, indicating that multilingual reasoning capabilities remain limited~\cite{huang-etal-2023-languages}. However, it introduces new risks, such as subtle linguistic cues, hedging, or culturally specific references may be altered or lost, affecting claim interpretation~\cite{quelle2023perils}. Cross-lingual reasoning is further complicated by script differences, morphological complexity, and culturally grounded narratives~\cite{chung2025translationllmbaseddatageneration, mirza2024globalliar, pelrine2023towards}. Models often produce incoherent or malformed outputs for Arabic, Telugu, or Burmese inputs, and may switch language during intermediate reasoning steps~\cite{anikina2025dfkinit2b, cao2023large, vykopal2025largelanguagemodelsmultilingual}. Beyond linguistic form, geopolitical biases further skew performance, as models trained predominantly on Western corpora apply implicit cultural assumptions, which undermine locally grounded fact-checking~\cite{mirza2024globalliar}. Evaluation is limited by the scarcity of native multilingual datasets, as many benchmarks rely on translated content that may not reflect real-world misinformation accurately.

\paragraph{Future Directions.}

Future work should prioritize reasoning directly in the target language, reducing reliance on translation-based pipelines. This includes improved cross-lingual transfer techniques~\cite{vykopal-etal-2025-soft}, multilingual prompting strategies that leverage high-resource reasoning to support low-resource outputs~\cite{vykopal2025largelanguagemodelsmultilingual}. While such approaches improve modularity, they rely heavily on translation-based approaches and may overlook cultural or contextual nuances in local claims. Future research should therefore focus on multilingual reasoning mechanisms that operate directly in the target language, including methods for modeling cross-lingual semantic alignment and culturally grounded interpenetration. Additionally, generating high-quality synthetic fact-checking data for low-resource languages using LLM represents a further step toward multilingual reliability~\cite{chung2025translationllmbaseddatageneration}.

\subsection{Agentic and Multi-Agent Systems}

\paragraph{Challenges.}

As fact-checking grows in complexity, recent work increasingly employs agentic architectures that decompose verification into multiple steps~\cite{li-etal-2024-self}. While these designs improve modularity, they introduce challenges related to coordination and error propagation~\cite{zhao-etal-2024-pacar, 10.1145/3696410.3714748}. Claim decomposition is particularly fragile, as over-decomposition may remove essential context while under-decomposition may omit important factual dependencies~\cite{hu-etal-2025-decomposition}. Errors introduced at early stages often propagate through subsequent components, as most current systems lack a mechanism for revising earlier decisions. Multi-agent debate framework partially addresses these issues but introduces coordination overhead and can generate consensus on incorrect conclusions when agents share similar parametric biases~\cite {kim2024llmsproducefaithfulexplanations,leippold2024automated}.

\paragraph{Future Directions.}

Iterative self-reflection mechanisms, where agents critique and revise intermediate outputs before producing the final verdict, show promise in improving reasoning reliability~\cite{leippold2024automated, 10.1145/3696410.3714748, sawinski2025openfact, li-etal-2024-self}, though current implementations remain difficult to control. Future research should focus on developing structured communication among agents,  dynamic task allocation that redistributes sub-tasks when an agent's confidence falls below a threshold, and methods for detecting and correcting cascading errors across pipeline stages.

\subsection{Prompt Sensitivity and Reasoning Stability}

\paragraph{Challenges.}

LLM-based fact-checking systems are highly sensitive to prompt formulation. Small changes in wording, prompt structure, or example selection can lead to substantially different predictions for the same claim~\cite{mamta-cocarascu-2025-facteval, 10.1145/3696410.3714569, guan2023language, schlichtkrull2023averitec}. This instability undermines reliability and complicates deployment in real-world settings. Chain-of-thought prompting has been widely adopted to improve multi-step reasoning, yet the generated reasoning traces themselves are not always reliable and may reinforce incorrect interpretations or misleading correlations~\cite{guan2023language, 10.1145/3696410.3714748}. High temperature settings further destabilize outputs, exposing the same claim to inconsistent verdicts across repeated queries, which is untenable in high-stakes deployment contexts.

\paragraph{Future Directions.}

Consistency-based approaches, where models are trained or prompted to produce stable verdicts across paraphrases, negations, and formulations of a claim under uncertainty, offer a principled response to this fragility~\cite{zeng-gao-2023-prompt}. Self-consistency and multi-path reasoning strategies, which aggregate multiple independent reasoning traces, may further reduce reliance on any single fragile prompt formulation. More broadly, fact-checking benchmarks should explicitly measure performance variance under prompt perturbation, rather than reporting results on a single fixed prompt, to expose instability that current evaluation protocols tend to conceal.

\subsection{Integration of External Knowledge and Temporal Reasoning}

\paragraph{Challenges.}

LLM integration with retrieval systems is complicated by both structural and temporal factors that jointly undermine faithful reasoning. At the structural level, large context windows do not guarantee effective evidence utilization: the \textit{lost-in-the-middle} phenomenon~\cite{liu-etal-2024-lost} shows that models frequently ignore relevant evidence unless it appears at salient positions within the prompt~\cite{chen-etal-2024-complex}. These problems are compounded by the static nature of LLM training, where models carry fixed parametric knowledge up to a cutoff date and may confidently apply outdated facts to contemporary claims without signaling temporal mismatch~\cite{mirza2024globalliar, li2023revisitfakenewsdataset, sun2024trustllm}. Critically, even when up-to-date evidence is retrieved and injected into the prompt, LLMs can exhibit \textit{memorization bias}, systematically prioritizing parametric knowledge over retrieved evidence, leading to reasoning conflicts in which recent accurate sources are selectively discarded~\cite{sun2024trustllm}.

\paragraph{Future Directions.}

Closing the gap between retrieval and reasoning requires tighter integration at multiple levels. Evidence-aware reranking strategies score retrieved documents by verification utility rather than semantic similarity, improving the relevance of what enters the context window. Reinforcement learning approaches that use LLM-generated feedback as a reward signal offer a path to retriever optimization without access to model internals~\cite{zhang-gao-2024-reinforcement}. On the temporal side, future systems should incorporate explicit temporal grounding mechanisms that require models to identify the time scope of both a claim and its evidence before reasoning about veracity. Dynamic retrieval-augmented generation frameworks should be designed to prioritize recency-aware evidence selection and actively reconcile conflicts between retrieved information and parametric memory, enabling models to represent temporal uncertainty and distinguish between outdated, disputed, and current facts as a component of the verification process.

\subsection{Evaluation of LLM-Based Fact-Checking Systems}

\paragraph{Challenges.}

Evaluating LLM-based fact-checking systems poses challenges that substantially differ from those of earlier extractive or classification-based approaches. LLMs generate free-form verdicts, explanations, and reasoning traces, making standard automatic metrics insufficient for capturing factual correctness and logical validity~\cite{sahnan2025llmsautomatefactcheckingarticle}. Text similarity measures may reward fluent but incorrect explanations, while ignoring reasoning errors or unsupported claims~\cite{mujahid2025stresstestingfactualconsistency}. Moreover, models may produce correct veracity labels through flawed reasoning, making it difficult to assess the reliability of generated explanations~\cite{wang2023explainable}. Current evaluation practices often collapse these dimensions into a single score, obscuring failure modes that are specific to generative reasoning. 

\paragraph{Future Directions.}

The increasing adoption of LLM-as-a-judge paradigms partially addressed scalability concerns but introduces new risks, including evaluator bias, self-agreement effects, and an overemphasis on persuasive language rather than evidence-grounded justification~\cite{10.5555/3666122.3668142}. Future research should focus on evaluation frameworks that separately assess verdict correctness, evidence usage, and reasoning faithfulness. Designing benchmarks that capture robustness across paraphrases, counterfactual claims, and multilingual contexts will also be important for evaluating next-generation fact-checking systems.

\section{Conclusion}

The rapid advancement of generative LLMs has sparked considerable interest in their potential applications within the fact-checking domain. In this survey, we provide a comprehensive and systematic overview of how generative LLMs are reshaping automated fact-checking. Unlike prior surveys that focus either on traditional fact-checking pipelines or on factuality and hallucinations, we structured the literature around three distinct yet interconnected roles of generative LLMs: (1) \textit{synthetic data generation}, (2) \textit{LLM-based prediction for downstream fact-checking tasks}, and (3) \textit{LLM-based evaluation}. This perspective highlights the diverse ways in which generative models contribute to fact-checking workflows, extending far beyond their role as standalone predictors.

Across core fact-checking tasks, including claim detection, previously fact-checked claim retrieval, evidence retrieval, and fact verification, generative LLMs have enabled a shift from narrowly defined classification problems toward more flexible, reasoning-oriented formulations. Models are increasingly used to normalize claims, decompose complex statements, generate questions, and produce faithful explanations or full fact-checking articles. These developments demonstrate how generative capabilities enable systems to more closely align with real-world fact-checking workflows, while also introducing new dependencies on prompt design, retrieval quality, and evidence grounding.

The survey further highlights the role of data in shaping LLM-based fact-checking systems. Generative models are increasingly used to augment existing datasets or to construct entirely synthetic resources that address data scarcity, domain imbalance, and limited language coverage. While such approaches have accelerated experimentation and enabled multilingual and low-resource research, they also blur the boundary between training and evaluation data, raising concerns about bias propagation and the development of benchmarks that distinguish between synthetic and real-world claims.

From a methodological perspective, the literature reveals a broad range of task-level paradigms for formulating fact-checking tasks when using generative LLMs. Traditional classification remains prevalent, particularly for fact verification and fake news detection, but is more often complemented by ranking-based formulation for evidence retrieval and previously fact-checked claim retrieval. At the same time, generative paradigms, such as claim normalization, explanation generation, question generation, or full fact-checking article generation, are gaining prominence, reflecting a shift toward outputs that are more informative and natural for human fact-checkers. Regression and scoring-based formulations further extend this space by enabling graded assessments of credibility or factuality. This diversity of paradigms underscores that the impact of generative LLMs extends beyond improved accuracy to enable alternative task formulations that better capture the complexity and interpretive nature of fact-checking.

Evaluation has emerged as a crucial and unresolved challenge. Traditional automatic metrics remain insufficient for assessing factual correctness, reasoning quality, and faithfulness of explanations in generative outputs. In response, LLM-based evaluation frameworks, framed as LLM-as-a-judge, have gained traction as scalable alternatives to human annotation. Our overview indicates that while these approaches can correlate with human judgments in certain settings, they are highly sensitive to the design of the prompt, the choice of model, and the evaluation criteria. As a result, LLM-based evaluation should be applied with caution and complemented by human oversight, especially when used to assess systems that themselves rely on generative models.

Finally, the survey reveals a persistent imbalance in language and domain coverage. Despite the promise of generative LLMs for cross-lingual and multilingual fact-checking, most empirical studies remain focused on English and a limited set of domains, such as politics and health. Expanding fact-checking research to underrepresented languages, scripts, and cultural contexts remains essential for building systems that are both globally applicable and socially responsible.

In conclusion, generative LLMs are redefining automated fact-checking by shaping how tasks are defined, data is generated, predictions are made, and systems are evaluated. Their integration offers significant opportunities to scale and enhance fact-checking workflows, but it also introduces new challenges related to hallucination, bias, and overconfidence in automated judgments. Addressing these challenges will require advances in evidence-grounded generation, transparent evaluation practices, multilingual benchmarking, and human-centered system design. By consolidating existing work through a unified taxonomy and identifying key limitations, this survey aims to support the development of more reliable, interpretable, and inclusive fact-checking systems in the area of generative AI.

\begin{acks}
The authors acknowledge using ChatGPT and Grammarly to check grammar, improve readability, language, and the flow of the text. After using these tools, the authors reviewed and edited the content as needed to ensure accuracy. Beyond the English improvements, we used the Qwen2.5 72B to filter the retrieved papers, followed by manual screening.

This research was partially supported by \textit{DisAI - Improving scientific excellence and creativity in combating disinformation with artificial intelligence and language technologies}, a project funded by the European Union under the \grantsponsor{disai-grant}{Horizon Europe}{https://doi.org/10.3030/101079164}, GA No.~\grantnum[https://doi.org/10.3030/101079164]{disai-grant}{101079164}; by the \textit{MIMEDIS}, a project funded by the \grantsponsor{mimedis}{Slovak Research and Development Agency}{} under GA No.~\grantnum{mimedis}{APVV-21-0114}; by the \textit{lorAI - Low Resource Artificial Intelligence}, a project funded by the \grantsponsor{lorai-grant}{European Union}{https://doi.org/10.3030/101136646} under the GA No.~\grantnum[https://doi.org/10.3030/101136646]{lorai-grant}{101136646}; by the \grantsponsor{sensai}{European Union NextGenerationEU}{} through the Recovery and Resilience Plan for Slovakia under the project No.~\grantnum{sensai}{09I01-03-V04-00100}; and by the \grantsponsor{trails-grant}{German Federal Ministry of Research, Technology and Space (BMFTR)}{} as part of the project TRAILS (\grantnum{trails-grant}{01IW24005}).
\end{acks}

% disai - \href{https://doi.org/10.3030/101079164}
% lorai - \href{https://doi.org/10.3030/101136646}

\bibliographystyle{ACM-Reference-Format}
\bibliography{sample-base}

% \appendix

% \input{content/appendix}

\end{document}